\documentclass[10pt, conference, compsocconf]{IEEEtran}

\usepackage{graphicx}
\usepackage{amsthm}

\usepackage{amssymb,latexsym,multicol,caption,algorithm}     % Standard packages
\usepackage{algorithm}  
\usepackage{algorithmic}
\usepackage{graphicx}
\usepackage{epsfig} 
\usepackage{epstopdf}
\usepackage{color}
\usepackage{multirow}
\usepackage{subfigure}
\usepackage{placeins}
\usepackage{bm}
\usepackage{graphics}
\usepackage{footnote}
\usepackage{multicol}
\usepackage{multirow}
\usepackage{url}
\usepackage{indentfirst}
\usepackage{longtable}
\usepackage{array}
\usepackage{mdwmath}
\usepackage{mdwtab}
\usepackage{rotating}
\usepackage{color}
\usepackage{morefloats}
\usepackage{slashbox}
\usepackage{rotating}
\usepackage{mcite}
\usepackage{eqnarray}
\usepackage{cite}
\usepackage[square, comma, sort&compress, numbers]{natbib}
\usepackage{float}
\usepackage{makecell}
\usepackage{amsmath}

\usepackage{pifont}

\newcommand{\algref}[1]{Algorithm \ref{#1}}

\renewcommand{\eqref}[1]{(\ref{#1})}

\theoremstyle{plain}

\theoremstyle{definition}

\theoremstyle{remark}

\ifCLASSINFOpdf
\else
\fi
\begin{document}
%
% paper title
% can use linebreaks \\ within to get better formatting as desired
\title{A Fast Factorization-based Approach to Robust PCA}

% author names and affiliations
% use a multiple column layout for up to two different
% affiliations

\author{\IEEEauthorblockN{Chong Peng, Zhao Kang, and Qiang Cheng}
\IEEEauthorblockA{ Department of Computer Science, Southern Illinois University,Carbondale, IL 62901 USA\\
Email: \{pchong,zhao.kang,qcheng\}@siu.edu}
}

\maketitle

\begin{abstract}
Robust principal component analysis (RPCA) has been widely used for recovering low-rank matrices in many data mining and machine learning problems. It separates a data matrix into a low-rank part and a sparse part. The convex approach has been well studied in the literature. However, state-of-the-art algorithms for the convex approach usually have relatively high complexity due to the need of solving (partial) singular value decompositions of large matrices. A non-convex approach, AltProj, has also been proposed with lighter complexity and better scalability. Given the true rank $r$ of the underlying low rank matrix, AltProj has a complexity of $O(r^2dn)$, where $d\times n$ is the size of data matrix. In this paper, we propose a novel factorization-based model of RPCA, which has a complexity of $O(kdn)$, where $k$ is an upper bound of the true rank. Our method does not need the precise value of the true rank. From extensive experiments, we observe that AltProj can work only when $r$ is precisely known in advance; however, when the needed rank parameter $r$ is specified to a value different from the true rank, AltProj cannot fully separate the two parts while our method succeeds. Even when both work, our method is about 4 times faster than AltProj. Our method can be used as a light-weight, scalable tool for RPCA in the absence of the precise value of the true rank. 

\end{abstract}


% For peer review papers, you can put extra information on the cover
% page as needed:
% \ifCLASSOPTIONpeerreview
% \begin{center} \bfseries EDICS Category: 3-BBND \end{center}
% \fi
%
% For peerreview papers, this IEEEtran command inserts a page break and
% creates the second title. It will be ignored for other modes.
\IEEEpeerreviewmaketitle

\section{Introduction}
\label{sec_intto}
Principal component analysis (PCA) is a fundamental technique of exploratory data analysis and has been widely used in many data mining tasks. Given a data matrix $X\in\mathcal{R}^{d\times n}$, the classic PCA seeks the best rank-$k$ approximation with complexity $O(kdn)$. It is well known that PCA is sensitive to outliers. To combat this drawback, in the past decade, a number of approaches to robust PCA (RPCA) have been proposed, including alternating minimization \cite{ke2005robust}, random sampling techniques \cite{maronna2006robust,de2003framework}, multivariate trimming \cite{gnanadesikan1972robust}, and others \cite{xu1995robust,croux2000principal}. More recently, a new type of RPCA method has emerged and become popular \cite{wright2009robust,candes2011robust}. It assumes that $X$ can be separated into two parts, i.e.,  a low-rank $L$ and a sparse $S$, by solving the following problem:
\begin{equation}
\label{eq_rpca}
\min_{L,S} \text{rank}(L) + \lambda \|S\|_0,	\quad s.t.\quad X = L+S,
\end{equation}
where $\|\cdot\|_{0}$ is the $\ell_0$ (pseudo) norm which counts the number of nonzero elements of a matrix, and $\lambda > 0$ is a balancing parameter. Because minimizing the rank and the $\ell_{0}$ norm in (\ref{eq_rpca}) is generally NP-hard, in practice, (\ref{eq_rpca}) is often relaxed to the following convex optimization problem \cite{candes2011robust}:
\begin{equation}
\label{eq_rpca_l1}
\min_{L,S} \|L\|_* + \lambda \|S\|_1,	\quad s.t.\quad X = L+S,
\end{equation}
where $\|L\|_{*}=\sum_{i=1}^{\min{(d,n)}}\sigma_{i}(L)$ is the nuclear norm of $L$ with $\sigma_{i}(L)$ denoting the $i$-th largest singular value of $L$, and $\|S\|_{1}=\sum_{ij}|S_{ij}|$ is the $\ell_1$ norm. Theoretically, under some mild conditions, (\ref{eq_rpca_l1}) can exactly separate $L$ with the true rank $r$ from $S$. A number of algorithms have been developed to solve (\ref{eq_rpca_l1}), including singular value thresholding (SVT) \cite{cai2010singular}, accelerated proximal gradient (APG) \cite{toh2010accelerated}, and two versions of augmented Lagrange multipliers (ALM) based approaches \cite{lin2010augmented}: exact ALM and inexact ALM (IALM). Among these algorithms, the ALM based are the state-of-the-art ones for solving (\ref{eq_rpca_l1}), which need to compute SVDs of $d\times n$ matrices per iteration. To improve efficiency, another ALM based algorithm adopts PROPACK package \cite{ding2011bayesian}, which solves only partial SVDs instead of full SVDs. However, this is still computationally costly when $d$ and $n$ are both large. Despite the elegant theory of the convex RPCA in (\ref{eq_rpca_l1}), it has two major drawbacks: 1) When the underlying matrix has no incoherence guarantee \cite{candes2011robust}, or the data get grossly corrupted, the results may be far from the underlying true ones; 2) The nuclear norm may lead to biased estimation of the rank \cite{kang2015robustpca}. To combat these drawbacks, \cite{leow2013background} fixes the rank of $L$ as a hard constraint, while \cite{kang2015robustpca} uses a non-convex rank approximation to more accurately approximate the rank of $L$, which needs to solve full SVDs. \cite{leow2013background,netrapalli2014non} need only to solve partial SVDs, which significantly reduces the complexity compared to computation of full SVD; for example, AltProj has a complexity of $O(r^2dn)$ \cite{netrapalli2014non}. However, if $r$ is not known a priori, \cite{leow2013background,netrapalli2014non} may fail to recover $L$ correctly.

To further reduce the complexity, enhance the scalability and alleviate the dependence on the knowledge of $r$, in this paper, we propose a factorization-based model for RPCA, where $L$ can be decomposed as $UCV^T$ with $U\in\mathcal{R}^{d\times k}$, $V\in\mathcal{R}^{n\times k}$, $C\in\mathcal{R}^{k\times k}$, and $k\ll \min{(d,n)}$. This model relaxes the requirement on a priori knowledge of the rank of $L$, assuming only that it is upper bounded by $k$. Scalable algorithms are developed to optimize our models efficiently. After acceptance of this paper, it came to our attention that \cite{liu2012active} also proposed a factorization-based approach for such a problem. However, our model is distinct from \cite{liu2012active}. Our model has two variants, one of which uses an explicit rank constraint by a matrix factorization in the case that the ground truth rank is known, while the other uses non-convex rank approximation in the case that the ground truth rank is unknown. Both variants of our model differ starkly from \cite{liu2012active}, which only considers the second case and simply uses the nuclear norm.

We summarize the contributions of this paper as follows: We propose a factorization-based model for RPCA, allowing the recovery of the low-rank component with or without a priori knowledge of its true rank. In the absence of true rank, a non-convex rank approximation is adopted, which can approximate the rank of a matrix more accurately than the nuclear norm. Efficient ALM-type optimization algorithms are developed with scalability in both $d$ and $n$. This is contrasted to AltProj whose cost is $O(r^2dn)$. This difference is important when $d$ and $n$ are large. Empirically, extensive experiments testify to the effectiveness of our model and algorithms both quantitatively and qualitatively in various applications.

\section{Related Work}
\label{sec_related}
The convex RPCA (\ref{eq_rpca_l1}) has been thoroughly studied \cite{cai2010singular}. To exploit the example-wise sparsity of the sparse component, $\ell_{2,1}$ norm has been adopted \cite{xu2010robust,mccoy2011two}:
\begin{equation}
\label{eq_rpca_l21}
\min_{L,S} \|L\|_* + \lambda \|S\|_{2,1},	\quad s.t.\quad X = L+S,
\end{equation}
where $\|\cdot\|_{2,1}$ is defined to be the sum of $\ell_2$ norms of column vectors of a matrix. When a matrix has large singular values, the nuclear norm may be far from an accurate approximation of the rank. Non-convex rank approximations have been considered in a number of applications, such as subspace clustering \cite{peng2015subspace,peng2016feature}. A non-convex rank approximation has also been used in RPCA \cite{kang2015robustpca}, which replaces the nuclear norm in (\ref{eq_rpca_l21}) by a non-convex rank approximation $\|L\|_{\gamma}=\sum_i \frac{(1+\gamma)\sigma_i(L)}{\gamma+\sigma_i(L)}$, with $\gamma > 0$, and $\sigma_i(L)$ being the $i$-th largest singular value of $L$. The above approaches usually need to solve SVDs. When the matrix involved is large, the computation of SVD, in general, is intensive. To reduce the complexity of RPCA, several approaches have been attempted. For example, AlgProj \cite{netrapalli2014non} uses non-convex alternating minimization techniques in RPCA and admits $O(r^2dn)$ complexity. \cite{liu2012active} uses a factorization approach:
\begin{equation}
\label{eq_rpca_fact}
\min_{U,V,S} \|V\|_{*} + \lambda \|S\|_{1},	\quad s.t. X = UV^T+S, U^TU = I,
\end{equation}
which solves SVDs of thin matrices and hence admits scalability, where $I$ is identity matrix with proper size.

\section{Fast Factorization-based RPCA}
\label{sec_proposed}
In this section, we formulate the \underline{F}ast \underline{F}actorization-based R\underline{P}CA (FFP). We model the data as $X=UCV^T+S$, where $S$ is a sparse component and $L=UCV^T$ is a low-rank approximation of $X$ with $C\in\mathcal{R}^{k\times k}$ being the core matrix, $U\in\mathcal{R}^{d\times k}$, $V\in\mathcal{R}^{n\times k}$ satisfying $U^TU=V^TV=I$ with $I$ being an identity matrix of a proper size. It is seen that the factorization provides a natural upper bound for the rank of the low-rank component of $X$. The upper bound, which is $k$, can be used to relax the stringent requirement on the knowledge of the true rank by AltProj algorithm. To capture the sparse structure of $S$, we adopt $\ell_{1}$ norm to obtain sparsity. Thus, we consider the following objective function:
\begin{equation}
\label{eq_obj_fpca}
\begin{aligned}
& \min_{S,C,U^TU=I,V^TV=I}\|S\|_{1}	\quad s.t. \quad X = UCV^T+S. 
%& s.t. \quad L = UV^T+S, \quad V^TV = I.
\end{aligned}
\end{equation}
For many applications, the true rank of $L$, which is $r$, is known. In the presence of this prior knowledge, we let $k=r$ in \ref{eq_obj_fpca}. However, when this information on $r$ is not present, \ref{eq_obj_fpca} may lack the desired capability of recovering $L$ with a (unknown) rank of $r$ while an arbitrary $k$ used. In this situation, we propose to marry the advantages of the convex and fixed-rank RPCA approaches, yielding the following optimization problem:
\begin{equation}
\label{eq_obj_ufpca_uv}
\begin{aligned}
& \min_{S,C,U,V}\|S\|_{1} + \lambda\|UCV^T\|_{*}	   \\ %s.t.  X \!=\! UCV^T \!+\! S,	\\
& s.t. \quad X = UCV^T+S, \quad U^TU = I, \quad V^TV = I,
\end{aligned}
\end{equation}
where $\lambda >0$ is a balancing parameter. Even in the case that knowledge of the precise value of $r$ is not available, a proper $k\ge r$ can still be chosen because, in the worst case, we may let $k = \min(d, n)$. Oftentimes, with domain information on the application at hand, an upper bound $k\ll \min{(d,n)}$ can be obtained. It has been shown that the nuclear norm can not approximate the true rank well if there are dominant singular values in a matrix, and non-convex rank approximations may help improve learning performance \cite{kang2015robustpca}. Here, we adopt a log-determinant rank approximation \cite{peng2016feature}, $\| Y \|_{ld} = \log \det (I + (Y^TY)^{\frac{1}{2}} )$, to obtain the following RPCA model:
\begin{equation}
\label{eq_obj_ufpca_uv_ld}
\begin{aligned}
& \min_{S,C,U,V}\|S\|_{1} \!+\! \lambda\|UCV^T\|_{ld}	  \\ % s.t.  X \!=\! UCV^T \!+\! S,	\\
& s.t. \quad X = UCV^T+S, \quad U^TU = I, \quad V^TV = I,
\end{aligned}
\end{equation}
Due to the fact that 
\begin{equation}
\begin{aligned}
	&	\|UCV^T\|_{ld}	=	\log \det (I + (VC^TU^T UCV^T)^{\frac{1}{2}} ) \\
=	&	\log \det (I + (C^TC)^{\frac{1}{2}} ) = \|C\|_{ld},
\end{aligned}
\end{equation}
model \ref{eq_obj_ufpca_uv_ld} can be reduced to the following model:

\begin{equation}
\label{eq_obj_ufpca}
\begin{aligned}
& \min_{S,C,U,V}\|S\|_{1} + \lambda\|C\|_{ld} \\	% \quad s.t. \quad X = UCV^T+S,		\\
& s.t. \quad X = UCV^T+S, \quad U^TU = I, \quad V^TV = I.
\end{aligned}
\end{equation}
We name models (\ref{eq_obj_fpca}) and (\ref{eq_obj_ufpca}) \underline{F}ixed Rank \underline{FFP} (F-FFP) and \underline{U}nfixed Rank FFP (U-FFP), respectively.

\section{Optimization}
\label{sec_optimization}
The augmented Lagrange functions of (\ref{eq_obj_fpca}) and (\ref{eq_obj_ufpca}) are
\begin{equation}
\label{eq_alm_ffp}
\min_{S,C,U^TU=V^TV=I} \|S\|_{1} + \frac{\rho}{2}\| X-UCV^T-S+\Theta / \rho \|_F^2,
\end{equation}
and
\begin{equation}
\begin{aligned}
\!& \min_{S,C,U,V} \|S\|_{1} \!+\! \lambda \|C\|_{ld} \!+\! \frac{\rho}{2}\| X \!-\! UCV^T \!-\! S \!+\! \frac{1}{\rho}\Theta \|_F^2	\\
& s.t.\quad U^TU = I, V^TV = I,
\end{aligned}
\end{equation}
respectively. The derivations for optimization are similar to those in \cite{peng2016feature}. We summarize the optimization in \algref{alg_ufr}. Here, we define $\mathcal{P}(\cdot)$ and $\mathcal{Q}(\cdot)$ to be the left and right singular vectors of a matrix from thin SVD, and define the operator $\mathcal{D}_{ \tau }(D) = \mathcal{P}(D) \text{diag}\{\sigma_i^*\} (\mathcal{Q}(D))^T$, with
\begin{equation}
\sigma_{i}^* = 
\begin{cases}
\xi,&\mbox{ if $f_i(\xi) \le f_i(0)$ and $ (1 + \sigma_{i}(D))^2 > 4\tau$, }	\\
0, 	&\mbox{ otherwise, }
\end{cases}
\end{equation}
%with $f_i(\sigma_i^*) = \frac{1}{2}(\sigma_i^*-\sigma_i(D))^2 + \tau \log (1+\sigma_i^*)$, and $\xi = \frac{\sigma_i(D)-1}{2} + \sqrt{\frac{(1+\sigma_i(D))^2}{2} - \tau }$.
% 
with $f_i(x) = \frac{1}{2}(x-\sigma_i(D))^2 + \tau \log (1+x)$, and $\xi = \frac{\sigma_i(D)-1}{2} + \sqrt{\frac{(1+\sigma_i(D))^2}{2} - \tau }$. 
%
%
%respectively. We summarize the optimization in \algref{alg_ufr}. Here, we define $\mathcal{P}(\cdot)$ and $\mathcal{Q}(\cdot)$ to be the left and right singular vectors of a matrix from thin SVD, and define the operator $\mathcal{D}_{ \tau }(D) = \mathcal{P}(D) \text{diag}\{\sigma_i(C)\} (\mathcal{Q}(D))^T$, with
%\begin{equation}
%\sigma_{i}(C) = 
%\begin{cases}
%\xi,&\mbox{ if $f_i(\xi) \le f_i(0)$ and $ (1 + \sigma_{i}(D))^2 > 4\tau $ },	\\
%0, 	&\mbox{ otherwise },
%\end{cases}
%\end{equation}
%with $f_i(\sigma_i(C)) = \frac{1}{2}(\sigma_i(C)-\sigma_i(D))^2 + \tau \log (1+\sigma_i(C))$, and $\xi = \frac{\sigma_i(D)-1}{2} + \sqrt{\frac{(1+\sigma_i(D))^2}{2} - \tau }$. 

\begin{algorithm}[h]
\algsetup{linenosize=\scriptsize } \scriptsize
\caption{ \small F-FFP for Solving \ref{eq_obj_fpca} (and, U-FFP for Solving \ref{eq_obj_ufpca}) } 
%\hrule  
\vspace{1mm}
\begin{algorithmic}[1] 
\STATE \textbf{Input}: $X$, $k$, $\lambda$, $\rho$, $\kappa>1$, $t_{max}$
\STATE \textbf{Initialize:} $S$, $U$, $V$, $\Theta$, $\rho$, and $t=0$. 
\REPEAT
\STATE $S_{ij} = (|[ X-UCV^T+\Theta / \rho ]_{ij} - 1/\rho|)\text{sgn}([ X-UCV^T+\Theta / \rho ]_{ij})$
\STATE $V = \mathcal{P}((X-S+\Theta/\rho)^TUC)(\mathcal{Q}((X-S+\Theta/\rho)^TUC))^T$
\STATE $U = \mathcal{P}((X-S+\Theta/\rho)VC^T)(\mathcal{Q}((X-S+\Theta/\rho)VC^T))^T$
\STATE $(\text{For F-FFP}) C = U^T(X-S+\Theta / \rho)V$
\STATE $(\text{For U-FFP}) C = \mathcal{D}_{ \lambda / \rho }{(U^T(X-S+\Theta / \rho)V)}$
\STATE $\Theta = \Theta + \rho(X-UCV^T-S)$, $\rho = \rho \kappa$
\UNTIL $t\geq t_{max}$ or convergence
\STATE \textbf{Output}: $S$, $U$, $V$, $C$
\vspace{1mm}
%\hrule   
\end{algorithmic}
\label{alg_ufr}
\end{algorithm}

\subsection{Complexity Analysis}
\label{sec_complexity}
Given that $k\ll \min{(d,n)}$, both F-FFP and U-FFP have complexity $O(ndk)$. % It is noted that when performing \algref{alg_ufr} for F-FFP, the multiplication $UC$, rather than $U$ and $C$ individually, can be calculated as an intermediate variable for efficiency, in a form similar to \ref{eq_rpca_fact}.

\section{Experiments}
\label{sec_experiments}
To evaluate the proposed model and algorithms, we consider three important applications: foreground-background separation, shadow removal from face images, and anomaly detection. We compare our algorithms with the state-of-the-art methods, including IALM\footnote{\scriptsize \url{http://perception.csl.illinois.edu/matrix-rank/sample_code.html\#RPCA}.} \cite{candes2011robust} and AltProj\footnote{\scriptsize \url{http://www.personal.psu.edu/nsa10/codes.html}. }, both of which make use of the PROPACK package \cite{ding2011bayesian} to solve SVDs for efficiency. All experiments in this section are conducted using Matlab on a dual-core Intel Xeon E3-1240 V2 3.40 GHz Linux Server with 8 GB memory. For purpose of reproductivity, we provide our codes on the website\footnote{\url{https://www.researchgate.net/publication/308174615_codes_icdm2016}}. 

\subsection{Foreground-background separation}
\label{sec_exp_background}
Foreground-background separation is to detect moving objects or interesting activities in a scene, and remove background(s) from a video sequence. For this task, we use 15 datasets, as listed in the first column of Table \ref{tab_known}, among which the first 11 contain a single background while the rest 4 have 2 backgrounds\footnote{\scriptsize Datasets used in this subsection can be found at:\\
\url{http://perception.i2r.a-star.edu.sg/bk_model/bk_index.html }\\ \url{http://limu.ait.kyushu-u.ac.jp/dataset/en/ }\\ \url{http://wordpress-jodoin.dmi.usherb.ca/dataset2012/ }\\ \url{http://research.microsoft.com/en-us/um/people/jckrumm/wallflower/testimages.htm }.}. In this case, we have $r=1$ and $2$ for the first 11 and last 4 datasets, respectively. For each dataset, the data matrix is constructed by treating all vectorized frames as columns\footnote{\scriptsize For computational ease, down-sampling is performed on Camera Parameter, Highway, Office, Shopping Mall, Pedestrian, and Time of Day data sets.}. In the following, we test F-FFP and U-FFP in two cases according to whether $r$ is known.

\subsubsection{Case 1 ($r$ is known)}
We set $k=r$ for F-FFP and AltProj. We terminate all methods after 200 iterations or when $\frac{\|X-L-S\|_F}{\|X\|_F}\leq 10^{-3}$ is reached. For IALM, we fix $\rho=0.0001$ and $\kappa=1.5$ for all these data sets for fast convergence and good visual quality. The balancing parameter is chosen as the theoretical one \cite{candes2011robust}. For Altproj, the default parameters are used. For F-FFP, we use the same $\rho$ and $\kappa$ as IALM. Without specification, the parameter settings remain the same throughout this paper.

The numerical results are reported in Table \ref{tab_known}. It is observed that IALM separates $S$ more sparsely but fails to recover $L$ with low rank, while F-FFP and AltProj recover $L$ properly. F-FFP generates more sparse $S$ than AltProj for most of the datasets. All these methods have competitive fitting error. However, it is possible for F-FFP to obtain more accurate fitting if the same iterations or time is provided as IALM or AltProj. Furthermore, F-FFP needs the least amount of time on all these datasets.

\begin{table}[h]
\huge
\centering
\caption{ Results on Different Datasets with $r$ Known }
\resizebox{1.0\columnwidth}{!}{
\begin{tabular}{|c||c|| c |c |c |c | c| c| }
\hline		
Data 	
& Method	& Rank($L$) & ${\|S\|_0}/{(d n)}$ & $\frac{\|X-L-S\|_F}{\|X\|_F}$	& \# of Iter. 	& \# of SVDs	& Time	\\ \hline
	
\multirow{3}{3cm}{ Highway } 	
& AltProj	& 1			& 0.9331		& 2.96e-4		& 37			& 38			& 49.65		\\	\cline{2-8}
& IALM		& 539		& 0.8175		& 6.02e-4		& 12			& 13			& 269.10	\\	\cline{2-8}
& F-FFP		& 1			& 0.8854 		& 5.74e-4		& 24			& 24			& 14.83		\\	\hline\hline

\multirow{3}{3cm}{ Office } 	
& AltProj	& 1			& 0.8018		& 9.40e-4		& 51			& 52			& 84.43		\\	\cline{2-8}
& IALM		& 374		& 0.7582		& 9.46e-4		& 11			& 12			& 230.53	\\	\cline{2-8}
& F-FFP		& 1			& 0.8761 		& 5.33e-4		& 24			& 24			& 19.92		\\	\hline\hline

\multirow{3}{3cm}{ PETS2006 } 	
& AltProj	& 1			& 0.8590		& 5.20e-4		& 35			& 36			& 44.64		\\	\cline{2-8}
& IALM		& 293		& 0.8649		& 5.63e-4		& 12			& 13			& 144.26	\\	\cline{2-8}
& F-FFP		& 1			& 0.8675		& 5.61e-4		& 24			& 24			& 14.33		\\	\hline\hline

\multirow{3}{3cm}{ Shopping Mall } 	
& AltProj	& 1			& 0.9853		& 3.91e-5		& 45			& 46			& 45.35		\\	\cline{2-8}
& IALM		& 328		& 0.8158		& 9.37e-4		& 11			& 12			& 123.99	\\	\cline{2-8}
& F-FFP		& 1			& 0.9122		& 7.70e-4		& 23			& 23			& 11.65		\\	\hline\hline

\multirow{3}{3cm}{ Pedestrian} 	
& AltProj	& 1			& 0.5869		& 9.32e-4		& 41			& 42			& 37.90		\\	\cline{2-8}
& IALM		& 35		& 0.8910		& 5.69e-4		& 11			& 12			& 36.18		\\	\cline{2-8}
& F-FFP		& 1			& 0.6023 		& 9.98e-4		& 22			& 22			& 10.53		\\	\hline\hline

\multirow{3}{3cm}{ Bootstrap } 	
& AltProj	& 1			& 0.9747		& 1.17e-4		& 44			& 45			& 107.15	\\	\cline{2-8}
& IALM		& 1146		& 0.8095		& 6.27e-4		& 12			& 13			& 1182.92	\\	\cline{2-8}
& F-FFP		& 1			& 0.9288		& 7.71e-4		& 23			& 23			& 25.38		\\	\hline\hline

\multirow{3}{3cm}{Water Surface} 	
& AltProj	& 1			& 0.8890		& 3.97e-4		& 47			& 48			& 27.27		\\	\cline{2-8}
& IALM		& 224		& 0.7861		& 5.32e-4		& 12			& 13			& 51.00		\\	\cline{2-8}
& F-FFP		& 1			& 0.8355		& 9.91e-4		& 23			& 23			& 5.68		\\	\hline\hline

\multirow{3}{3cm}{ Campus } 	
& AltProj	& 1			& 0.9790		& 9.50e-5		& 41			& 42			& 54.1		\\	\cline{2-8}
& IALM		& 488		& 0.8136		& 9.30e-4		& 11			& 12			& 242.59	\\	\cline{2-8}
& F-FFP		& 1			& 0.9378 		& 6.26e-4		& 23			& 23			& 12.85		\\	\hline\hline

\multirow{3}{3cm}{ Curtain } 	
& AltProj	& 1			& 0.8280		& 7.46e-4		& 40			& 41			& 102.41	\\	\cline{2-8}
& IALM		& 834		& 0.7398		& 6.84e-4		& 12			& 13			& 747.36	\\	\cline{2-8}
& F-FFP		& 1			& 0.8680 		& 6.28e-4		& 24			& 24			& 27.51 	\\	\hline\hline

\multirow{3}{3cm}{ Fountain } 	
& AltProj	& 1			& 0.9113		& 2.91e-4		& 50			& 51			& 23.90		\\	\cline{2-8}
& IALM		& 102		& 0.8272		& 4.91e-4		& 12			& 13			& 25.62		\\	\cline{2-8}
& F-FFP		& 1			& 0.8854		& 4.89e-4		& 24			& 24			& 5.00		\\	\hline\hline

\multirow{3}{3cm}{ Escalator Airport } 	
& AltProj	& 1			& 0.9152		& 2.29e-4		& 40			& 41			& 110.75	\\	\cline{2-8}
& IALM		& 957		& 0.7744		& 7.76e-4		& 11			& 12			& 1,040.91	\\	\cline{2-8}
& F-FFP		& 1			& 0.8877		& 5.45e-4		& 23			& 23			& 30.78		\\ 	\hline	\hline

\multirow{3}{3cm}{Lobby} 	
& AltProj	& 2			& 0.9243		& 1.88e-4		& 39			& 41			& 47.32		\\	\cline{2-8}
& IALM		& 223		& 0.8346		& 6.19e-4		& 12			& 13			& 152.54	\\	\cline{2-8}
& F-FFP		& 2			& 0.8524		& 6.42e-4		& 24			& 24			& 15.20		\\	\hline\hline

\multirow{3}{3cm}{Light Switch-2} 	
& AltProj	& 2			& 0.9050		& 2.24e-4		& 47			& 49			& 87.35	\\	\cline{2-8}
& IALM		& 591		& 0.7921		& 7.93e-4		& 12			& 13			& 613.98	\\	\cline{2-8}
& F-FFP		& 2			& 0.8323		& 7.54e-4		& 24			& 24			& 24.12		\\	\hline\hline

\multirow{3}{3cm}{Camera Parameter} 	
& AltProj	& 2			& 0.8806		& 5.34e-4		& 47			& 49			& 84.99	\\	\cline{2-8}
& IALM		& 607		& 0.7750		& 6.86e-4		& 12			& 13			& 433.47	\\	\cline{2-8}
& F-FFP		& 2			& 0.8684		& 6.16e-4		& 24			& 24			& 22.25		\\	\hline\hline

\multirow{3}{3cm}{Time Of Day} 	
& AltProj	& 2			& 0.8646		& 4.72e-4		& 44			& 46			& 61.63		\\	\cline{2-8}
& IALM		& 351		& 0.6990		& 6.12e-4		& 13			& 14			& 265.87	\\	\cline{2-8}
& F-FFP		& 2			& 0.8441		& 6.81e-4		& 25			& 25			& 18.49		\\	\hline
%---------------------
\end{tabular}
}
\scriptsize 
\begin{flushleft}
For IALM and AltProj, (partial) SVDs are for $d\times n$ matrices. For F-FFP, SVDs are for $n\times k$ matrices, which are computationally far less expensive than those required by IALM and AltProj.
\end{flushleft}
\label{tab_known}
\end{table}

\subsubsection{Case 2 ($r$ is unknown)}
$k$ is specified as a tight upper bound of $r$ based on domain knowledge on the video. In this test, we set $k=5$ on all datasets and compare U-FFP with AltProj and IALM\footnote{\scriptsize Here the results of IALM are not shown since they are the same as Case 1.}. For U-FFP, $\lambda$ is chosen from $1e\{6,7,8,9\}$. We show the numerical results in Table \ref{tab_unknown}. It is observed that U-FFP is still able to recover $L$ with the true rank, whereas AltProj fails in this case. Besides, the time cost of U-FFP increases slightly by less than 1 second for most of the datasets while the time cost of AltProj increases by about 10-20 seconds.

\begin{figure}[h]
\centering
\resizebox{1.0\columnwidth}{!}{
\begin{tabular}{c c || c c c  }
\includegraphics[width=0.2\columnwidth]{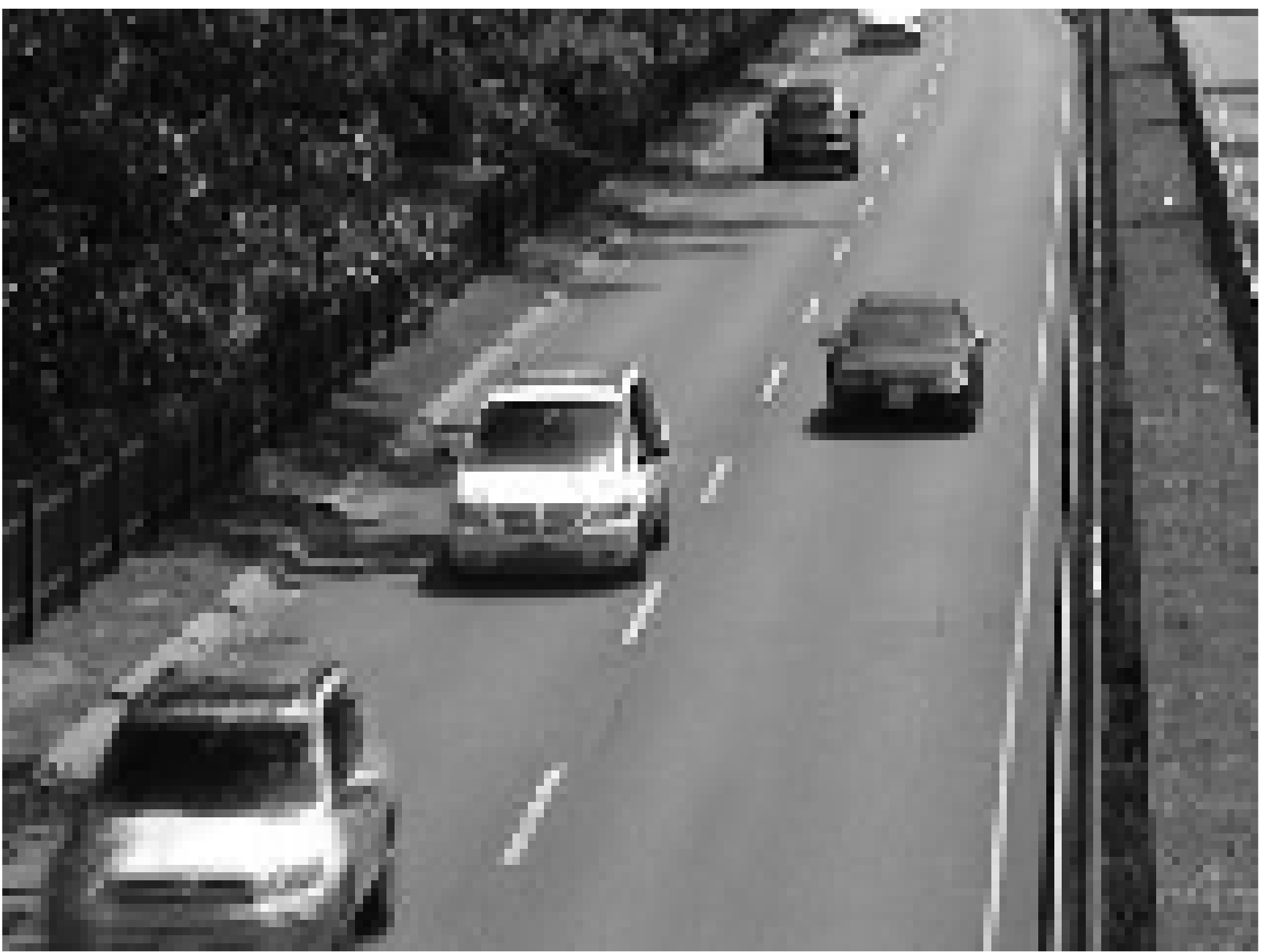}	&
&
&
\includegraphics[width=0.2\columnwidth]{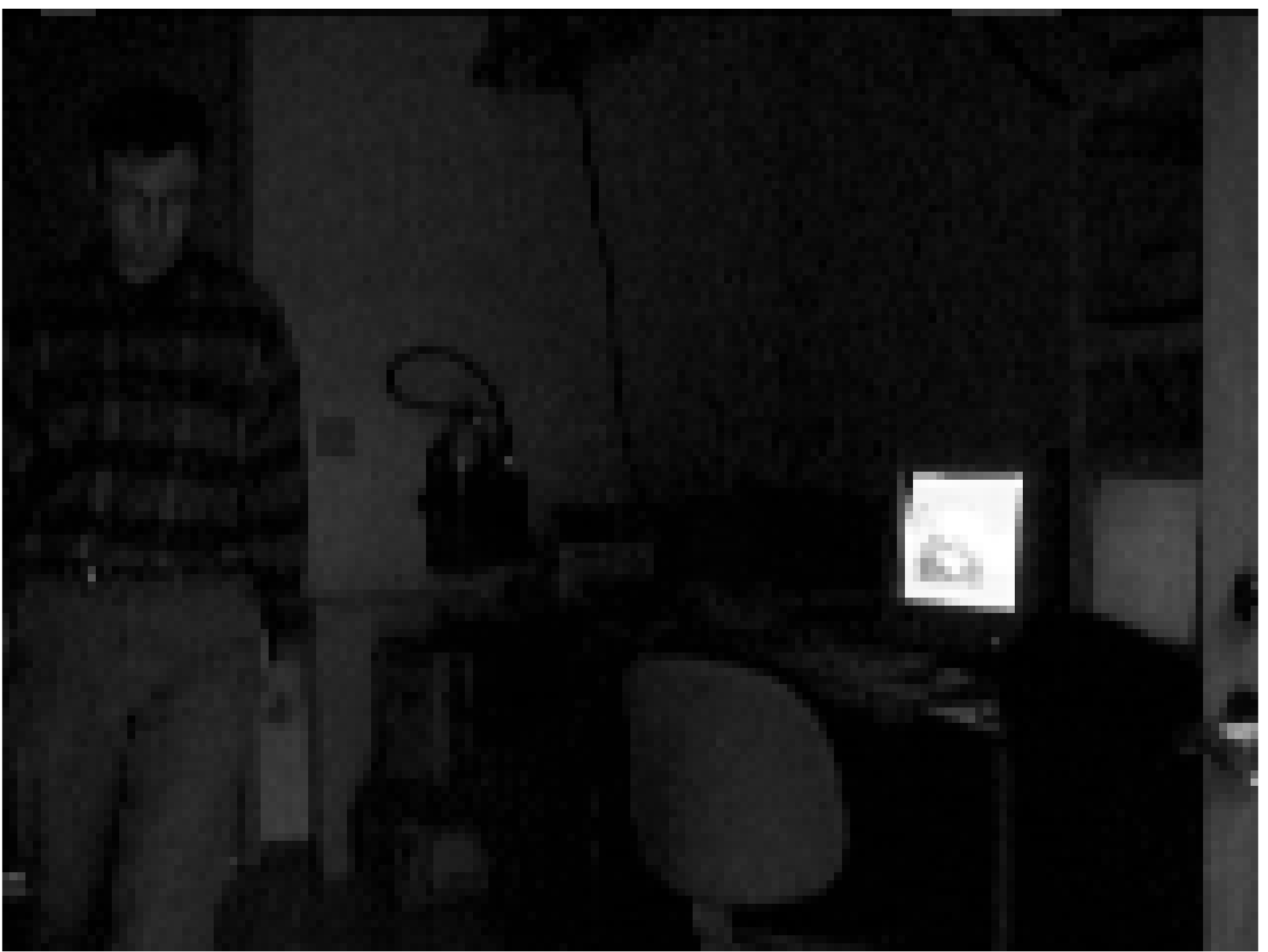}	&
\includegraphics[width=0.2\columnwidth]{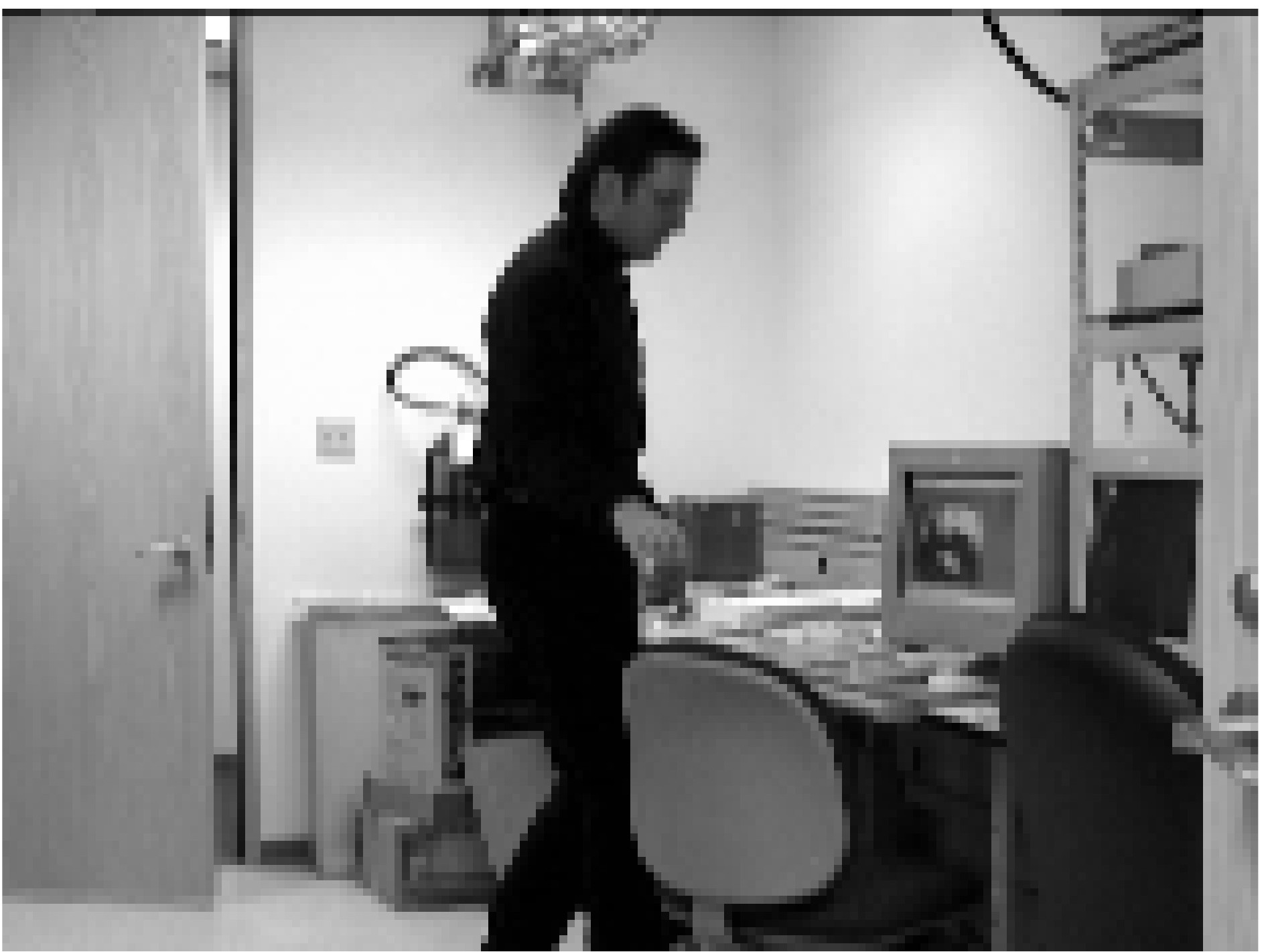}			\\
 \footnotesize{(1)}  	& \multicolumn{2}{c}{}  	& \footnotesize{(2)}  	& \footnotesize{(3)} 
\end{tabular}
}
\caption{ \footnotesize (1) is a frame from Highway and (2)-(3) are two frames from Light Switch-2. } 
\label{fig_sep_frame}
\end{figure}

\begin{figure}[h]
\centering
\scriptsize
{\includegraphics[width=0.18\columnwidth]{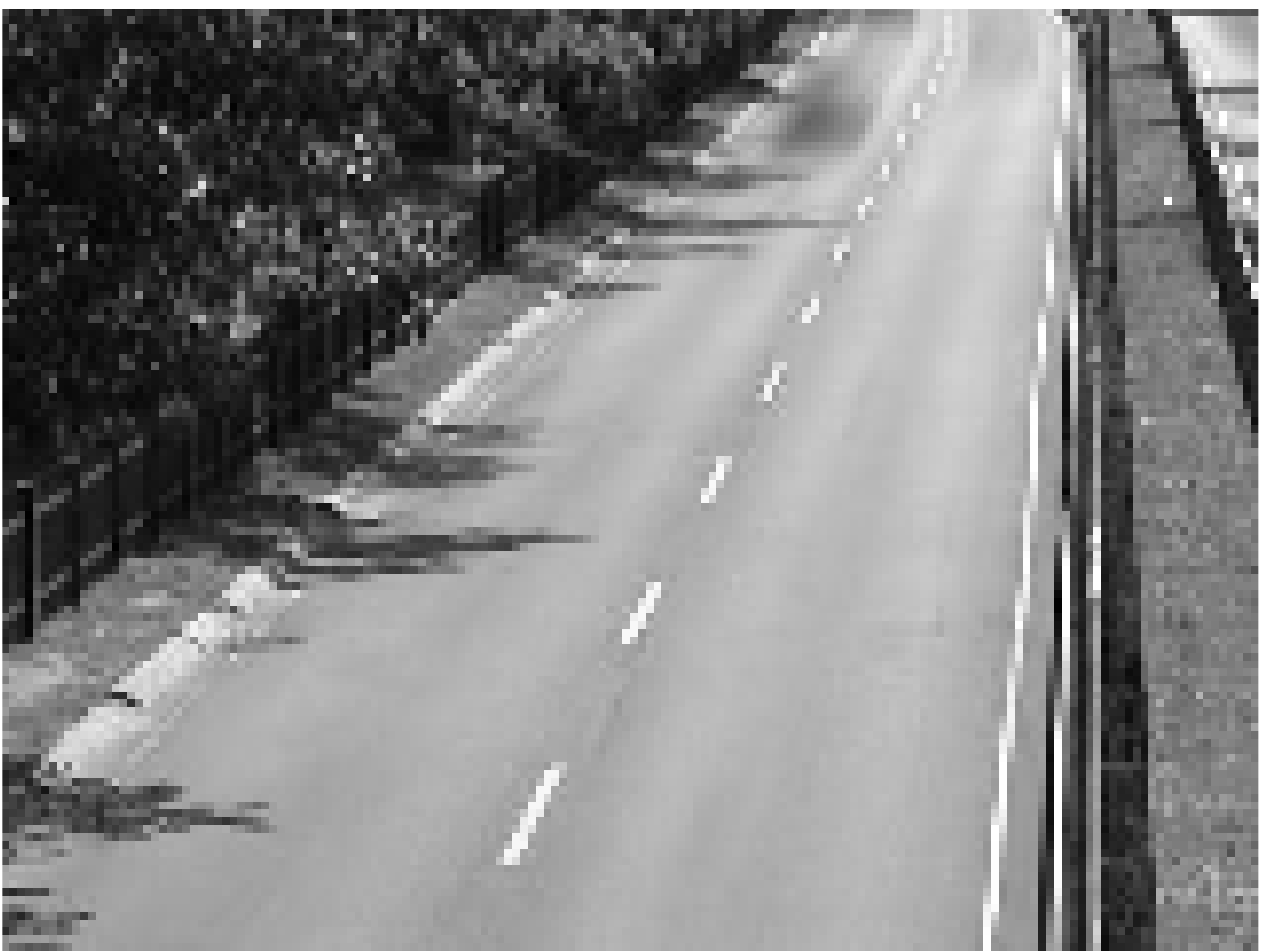} } \hspace{-1mm}
{\includegraphics[width=0.18\columnwidth]{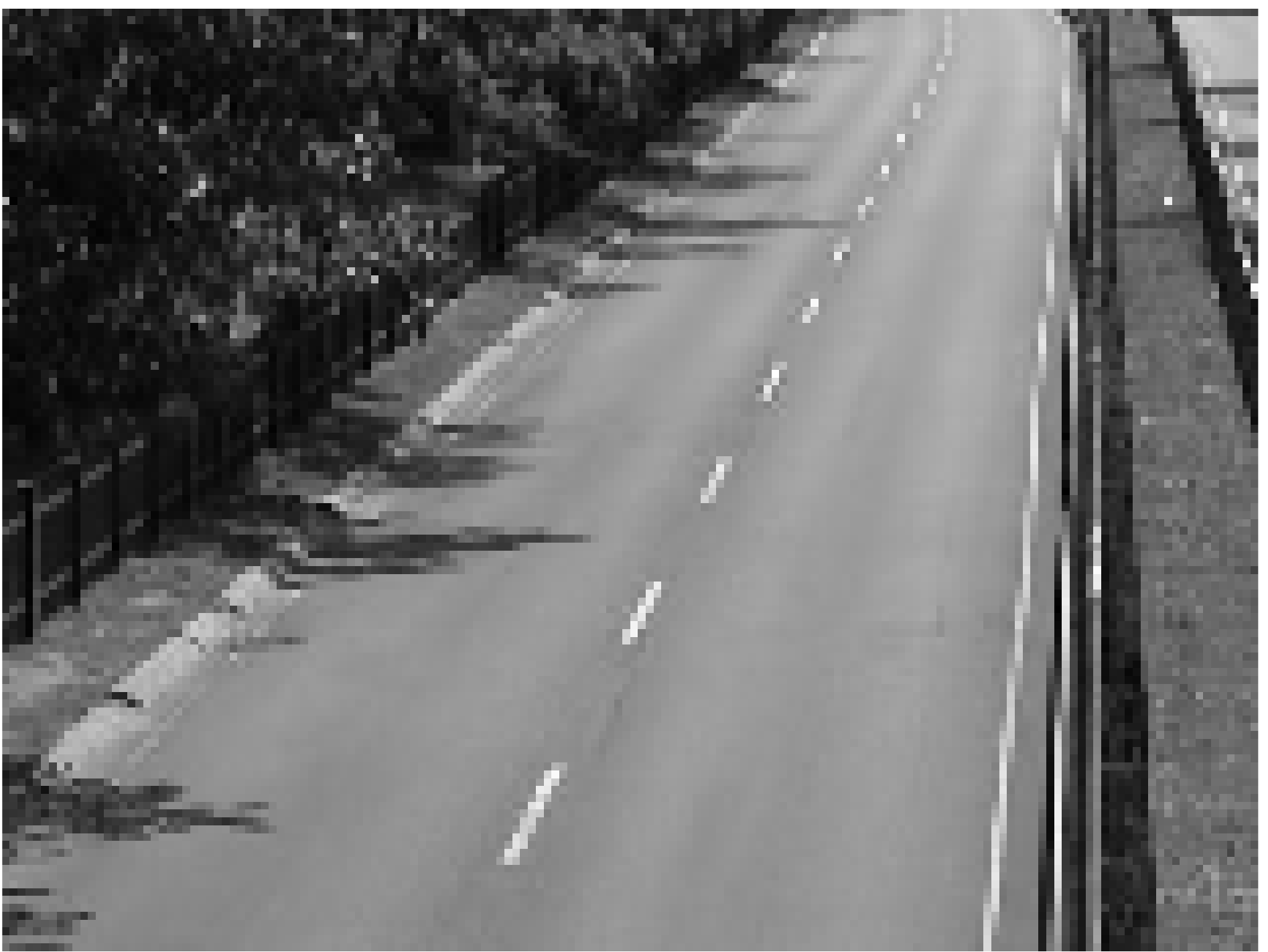} } \hspace{-1mm}
{\includegraphics[width=0.18\columnwidth]{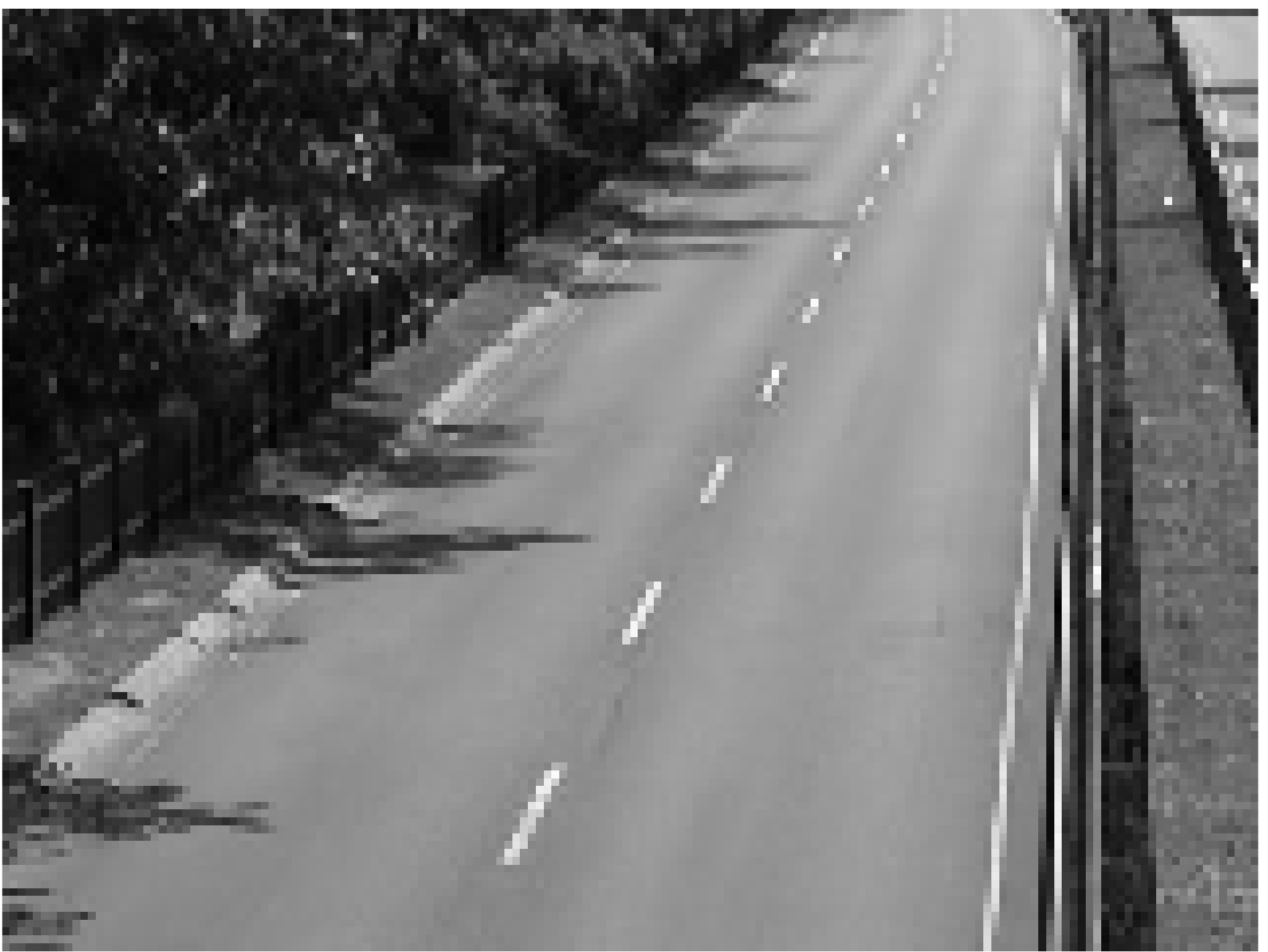} }		\hspace{-1mm}
{\includegraphics[width=0.18\columnwidth]{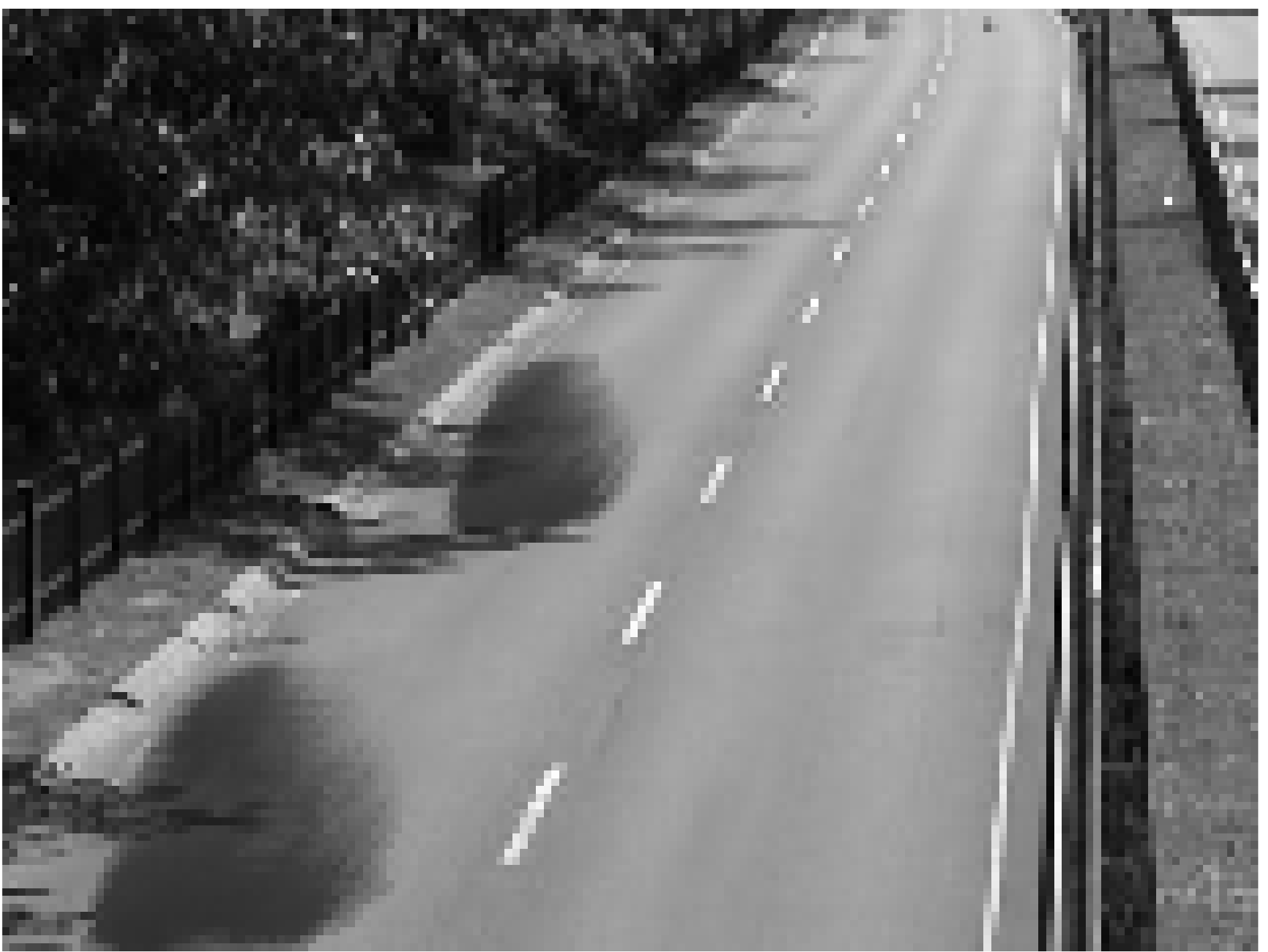} }	\hspace{-1mm}
{\includegraphics[width=0.18\columnwidth]{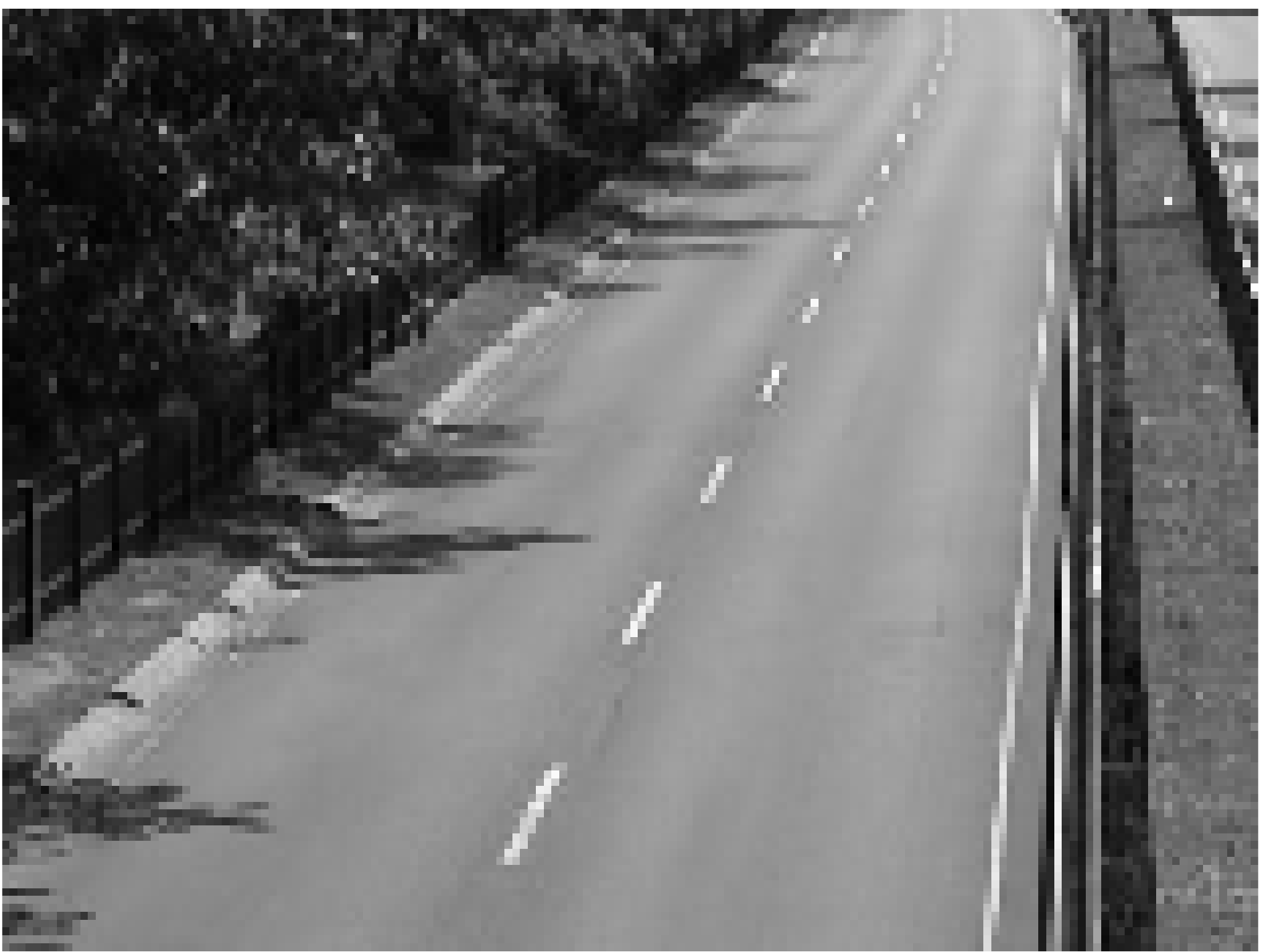} }			\\
	
{\includegraphics[width=0.18\columnwidth]{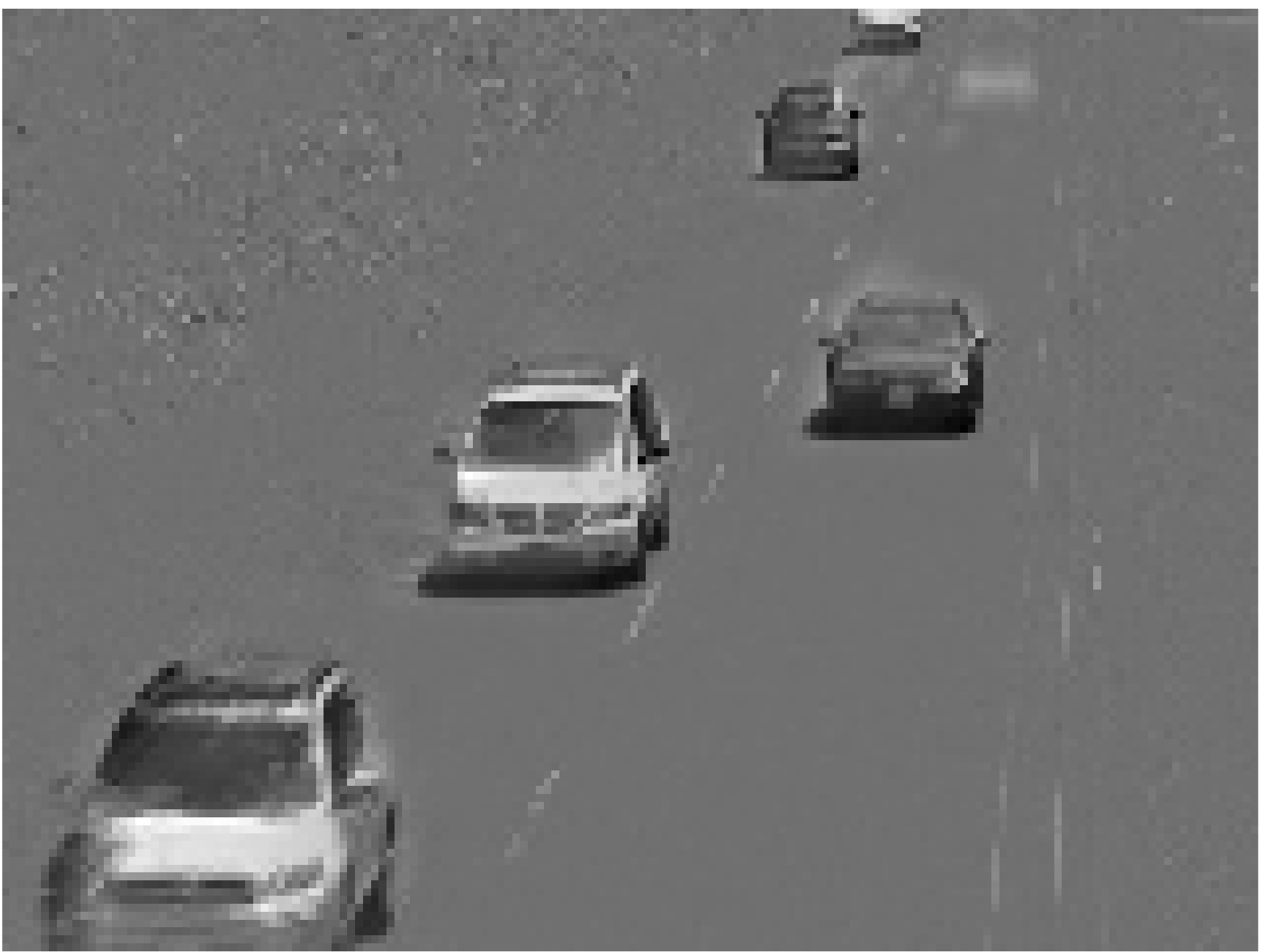} }	\hspace{-1mm}
{\includegraphics[width=0.18\columnwidth]{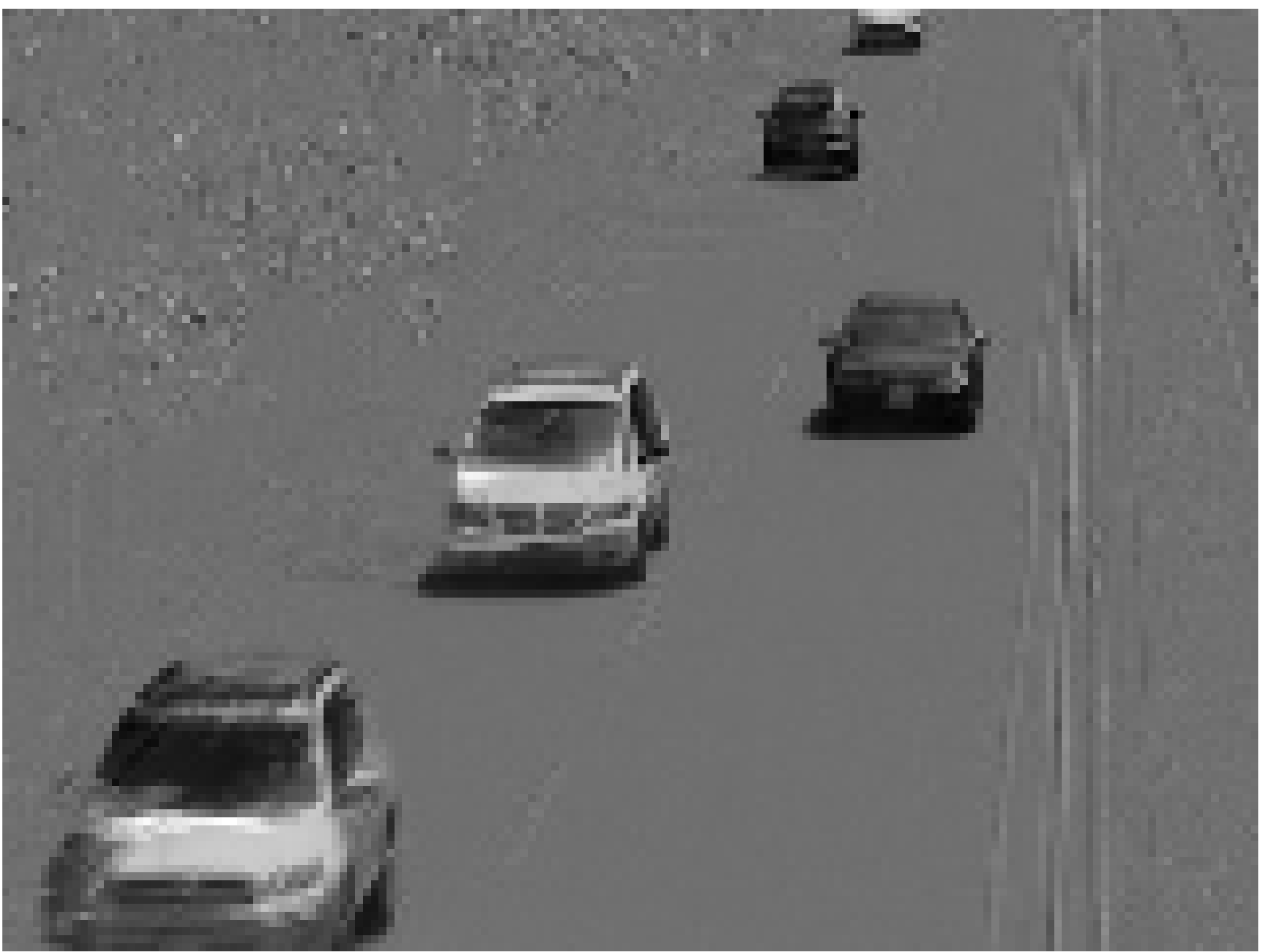} }	\hspace{-1mm}
{\includegraphics[width=0.18\columnwidth]{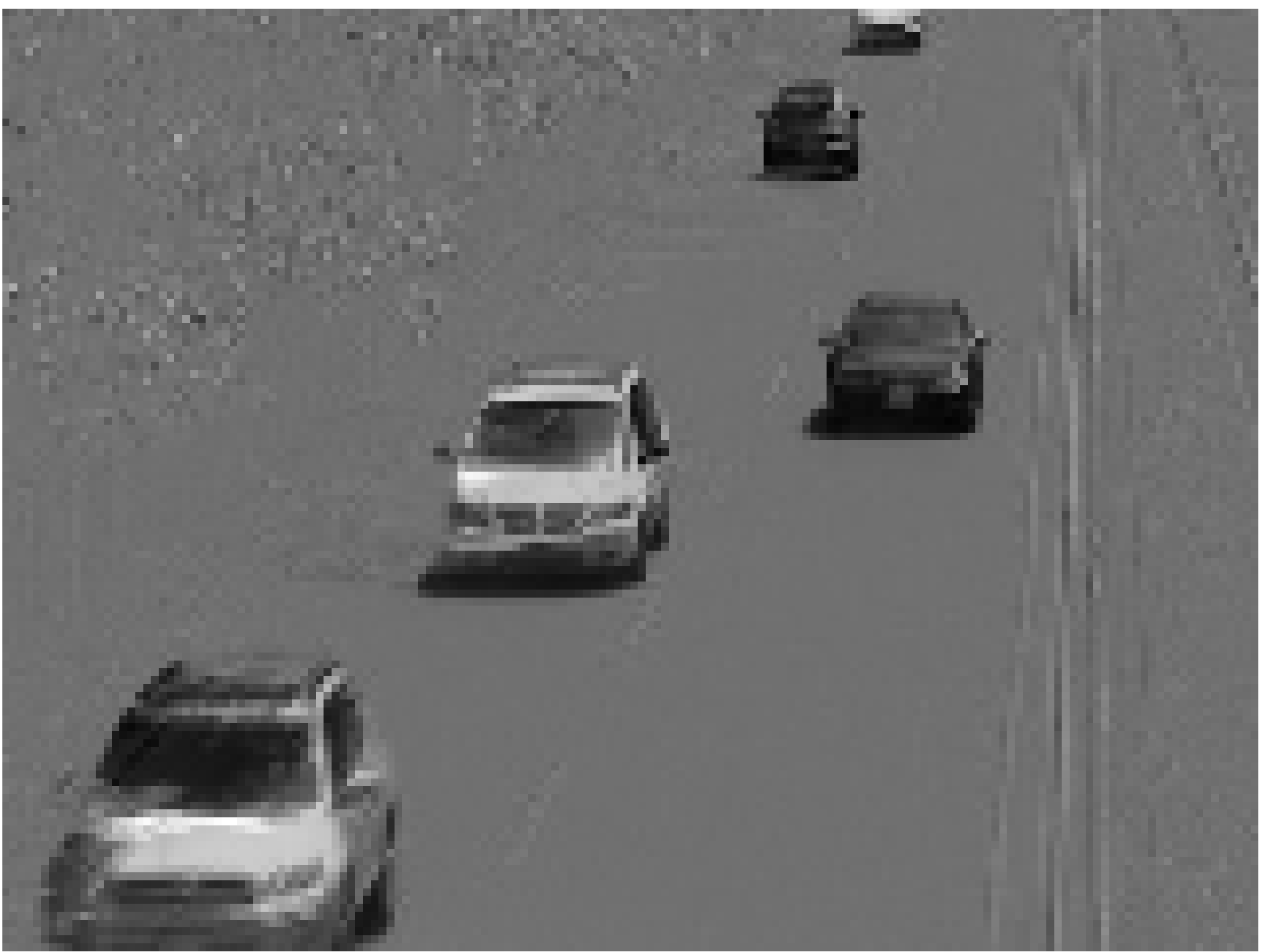} }		\hspace{-1mm}
{\includegraphics[width=0.18\columnwidth]{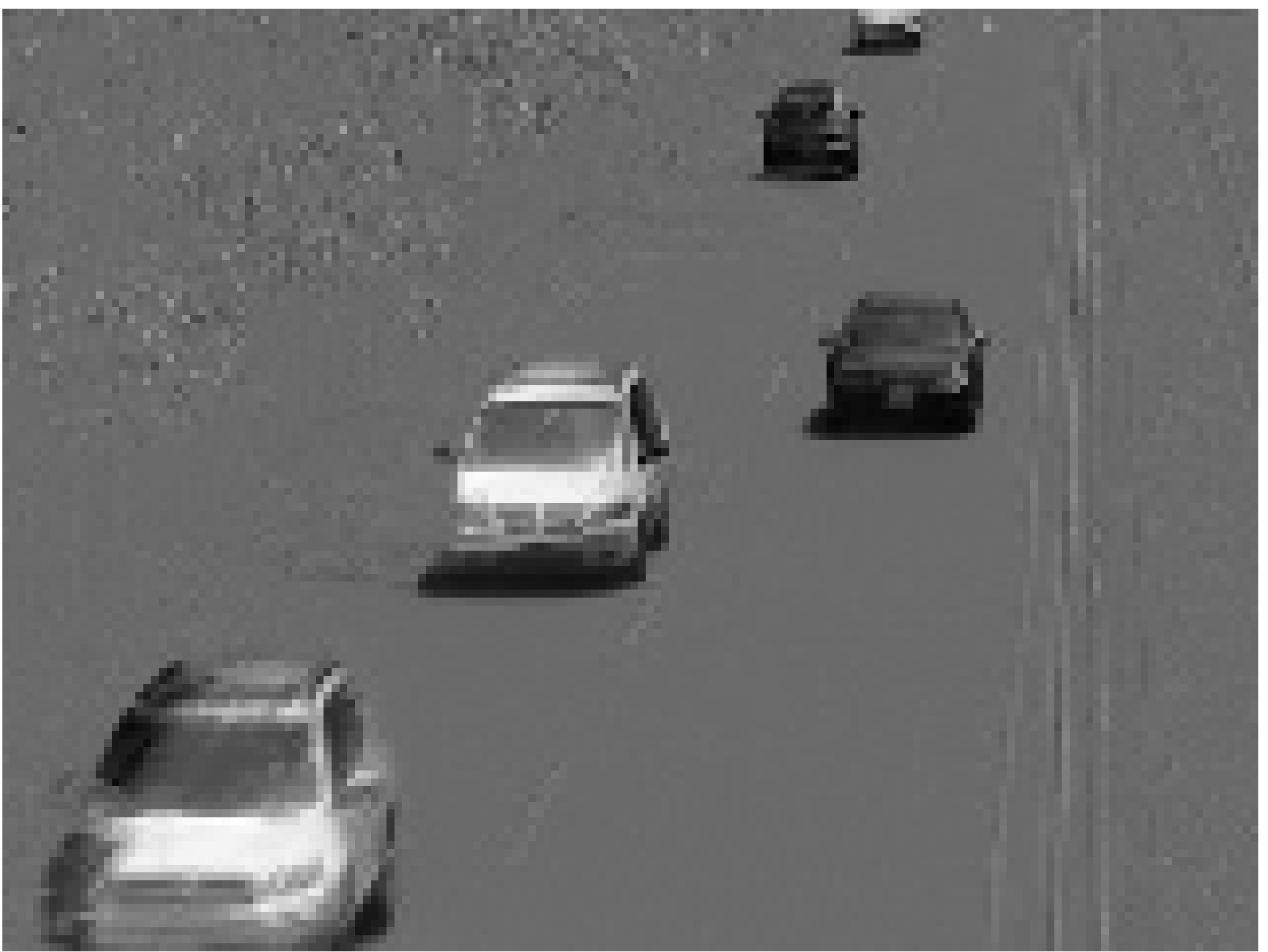} }	\hspace{-1mm}
{\includegraphics[width=0.18\columnwidth]{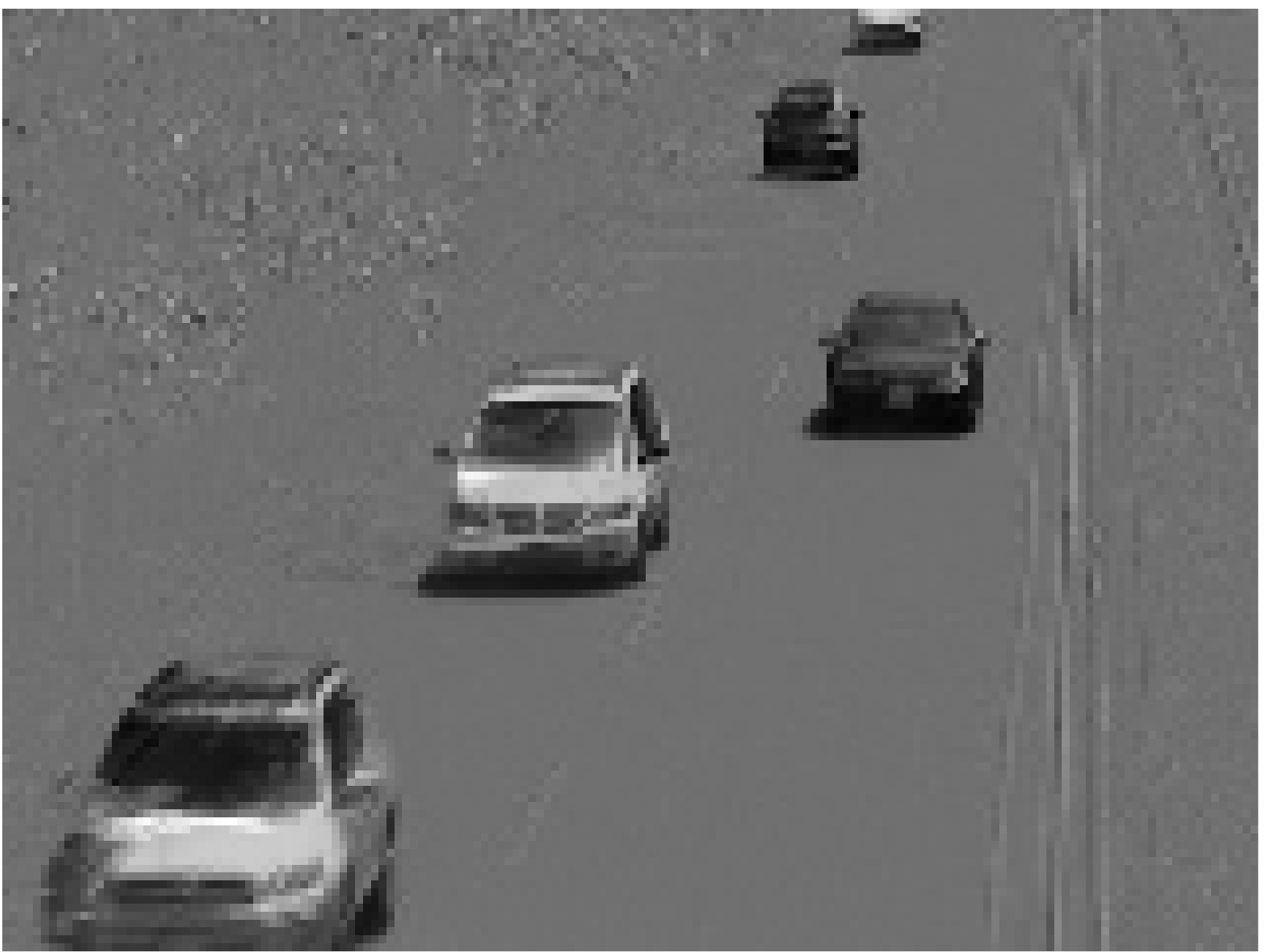} }			\\

\vspace{2mm}

{\includegraphics[width=0.18\columnwidth]{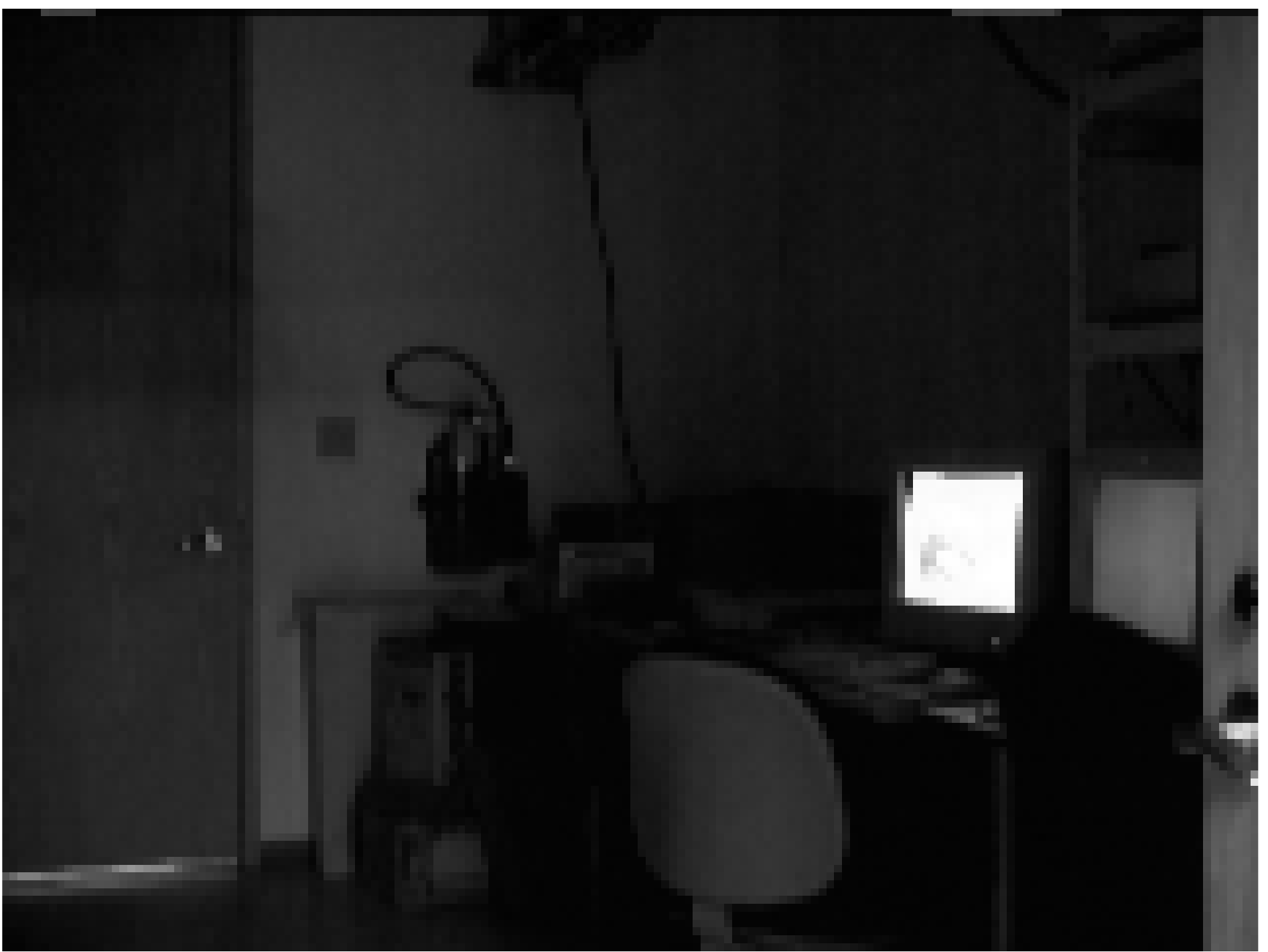} }	\hspace{-1mm}
{\includegraphics[width=0.18\columnwidth]{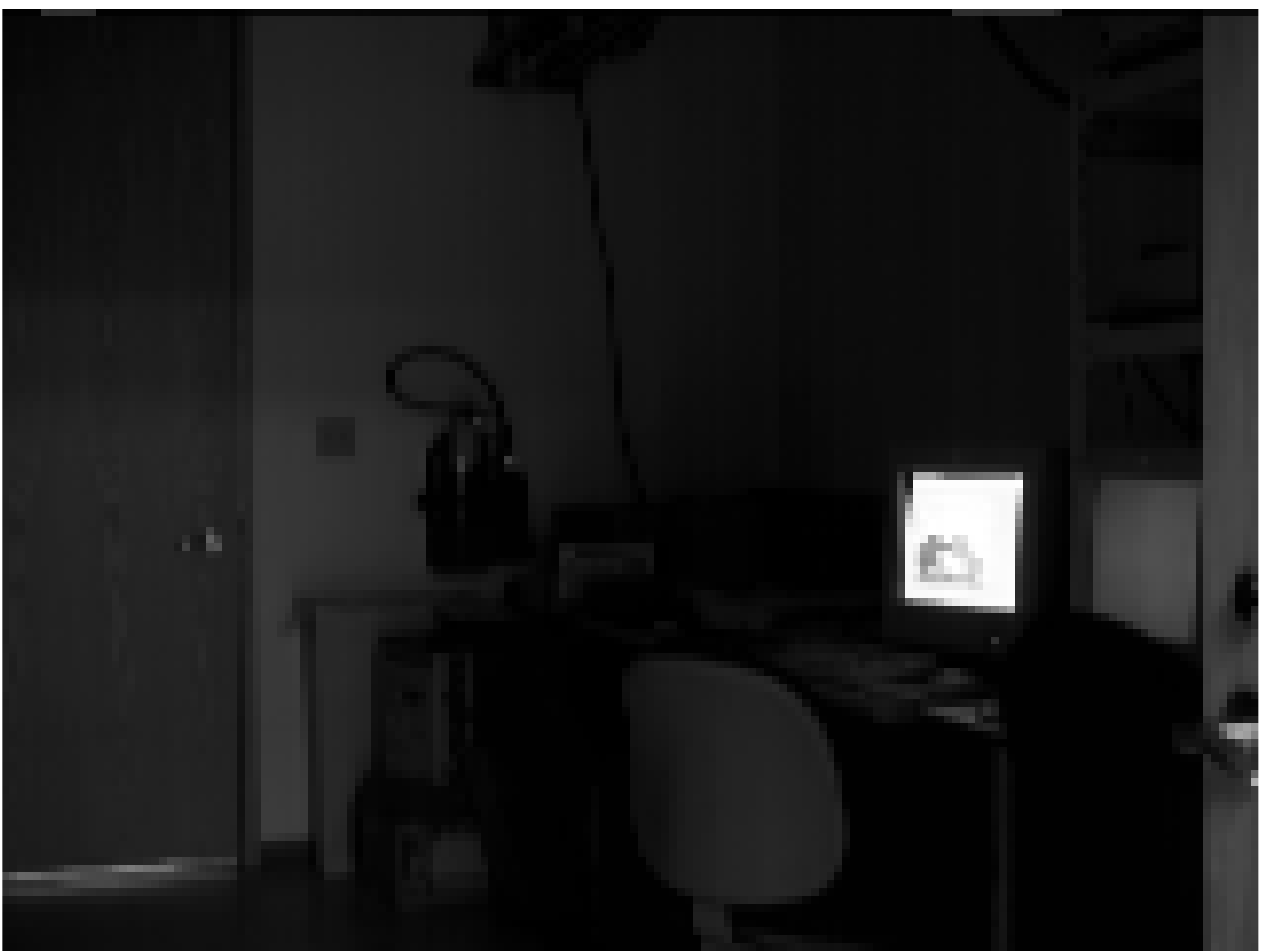} }	\hspace{-1mm}
{\includegraphics[width=0.18\columnwidth]{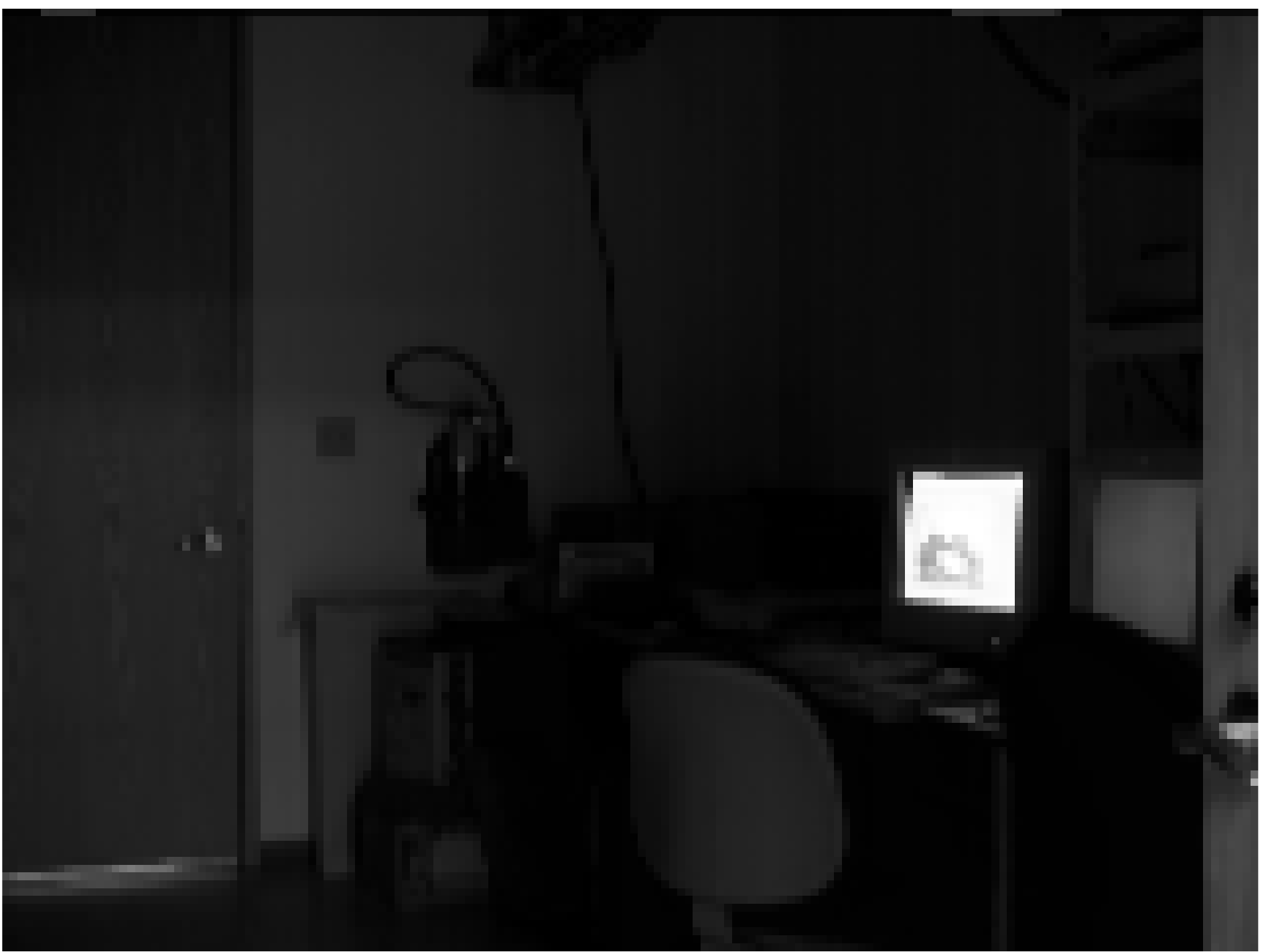} }		\hspace{-1mm}
{\includegraphics[width=0.18\columnwidth]{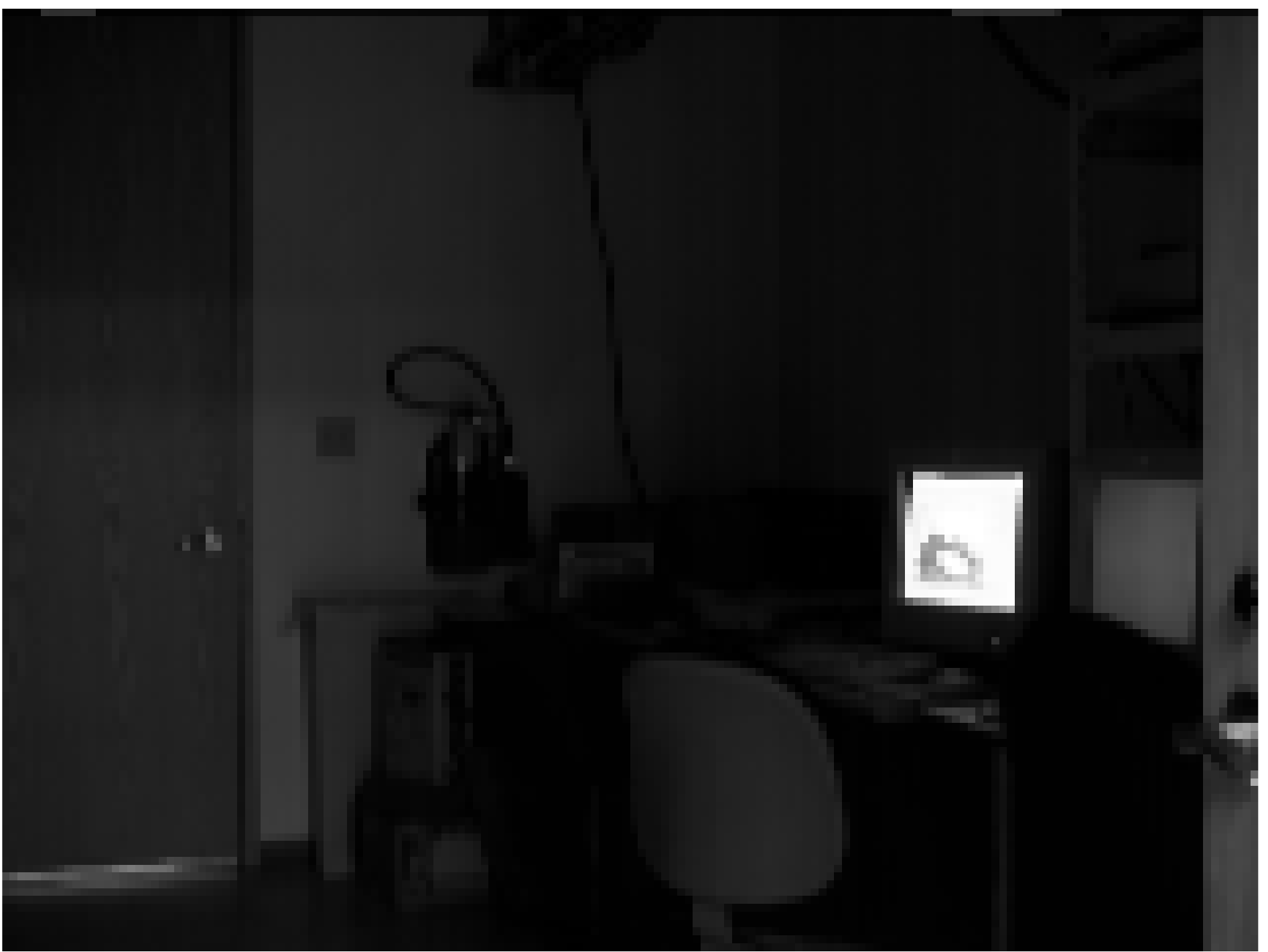} }	\hspace{-1mm}
{\includegraphics[width=0.18\columnwidth]{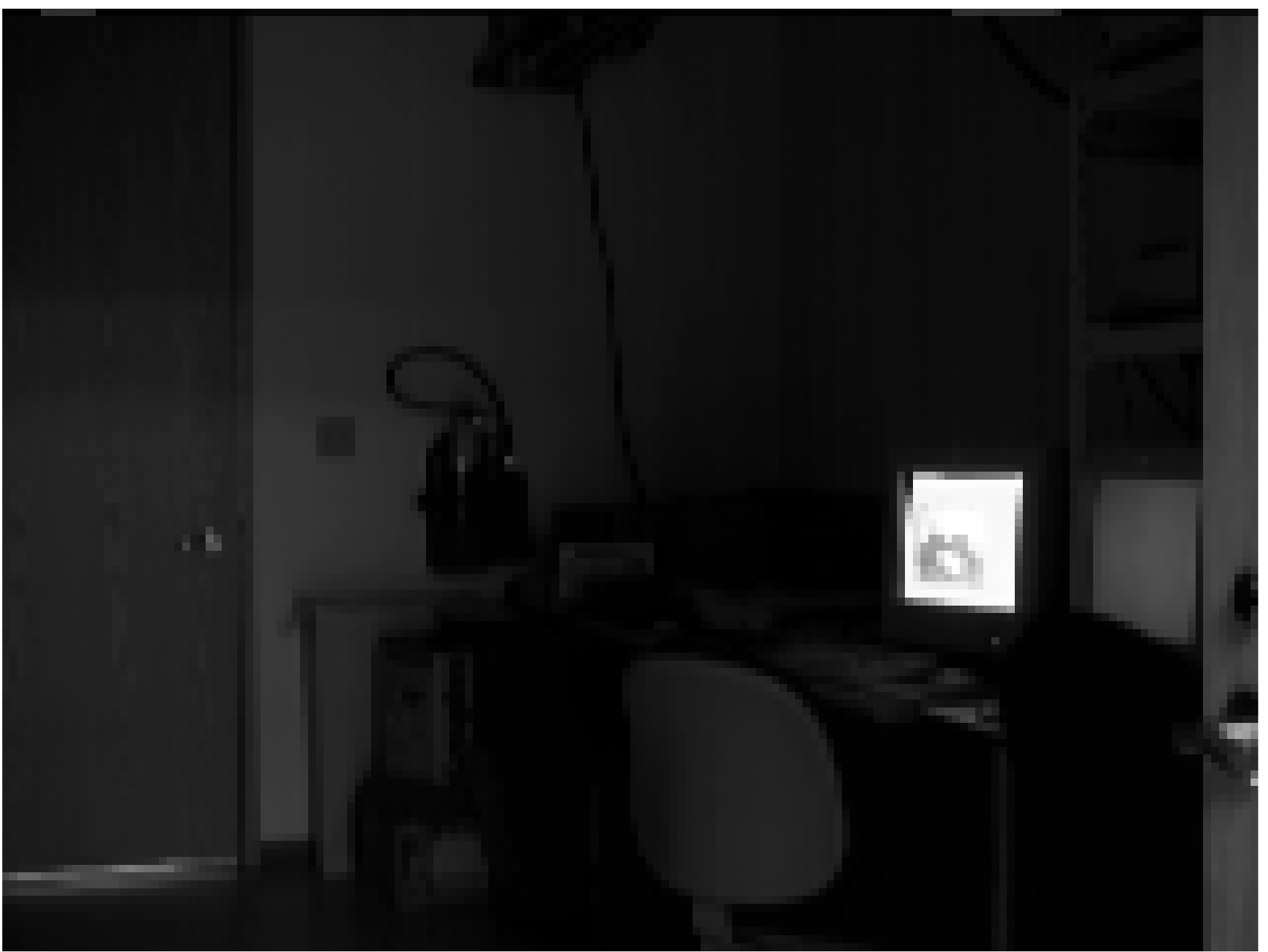} }			\\

{\includegraphics[width=0.18\columnwidth]{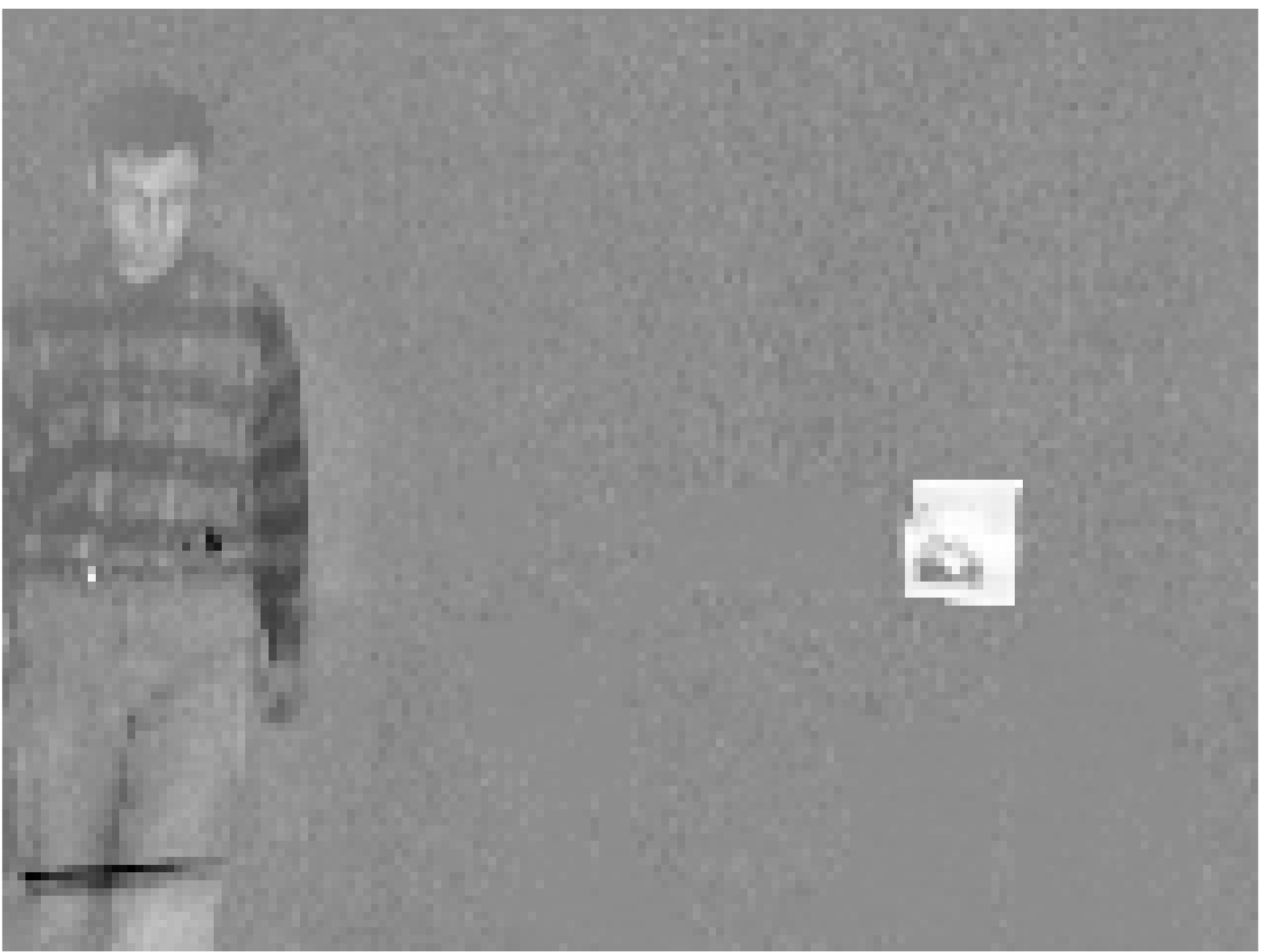} }	\hspace{-1mm}
{\includegraphics[width=0.18\columnwidth]{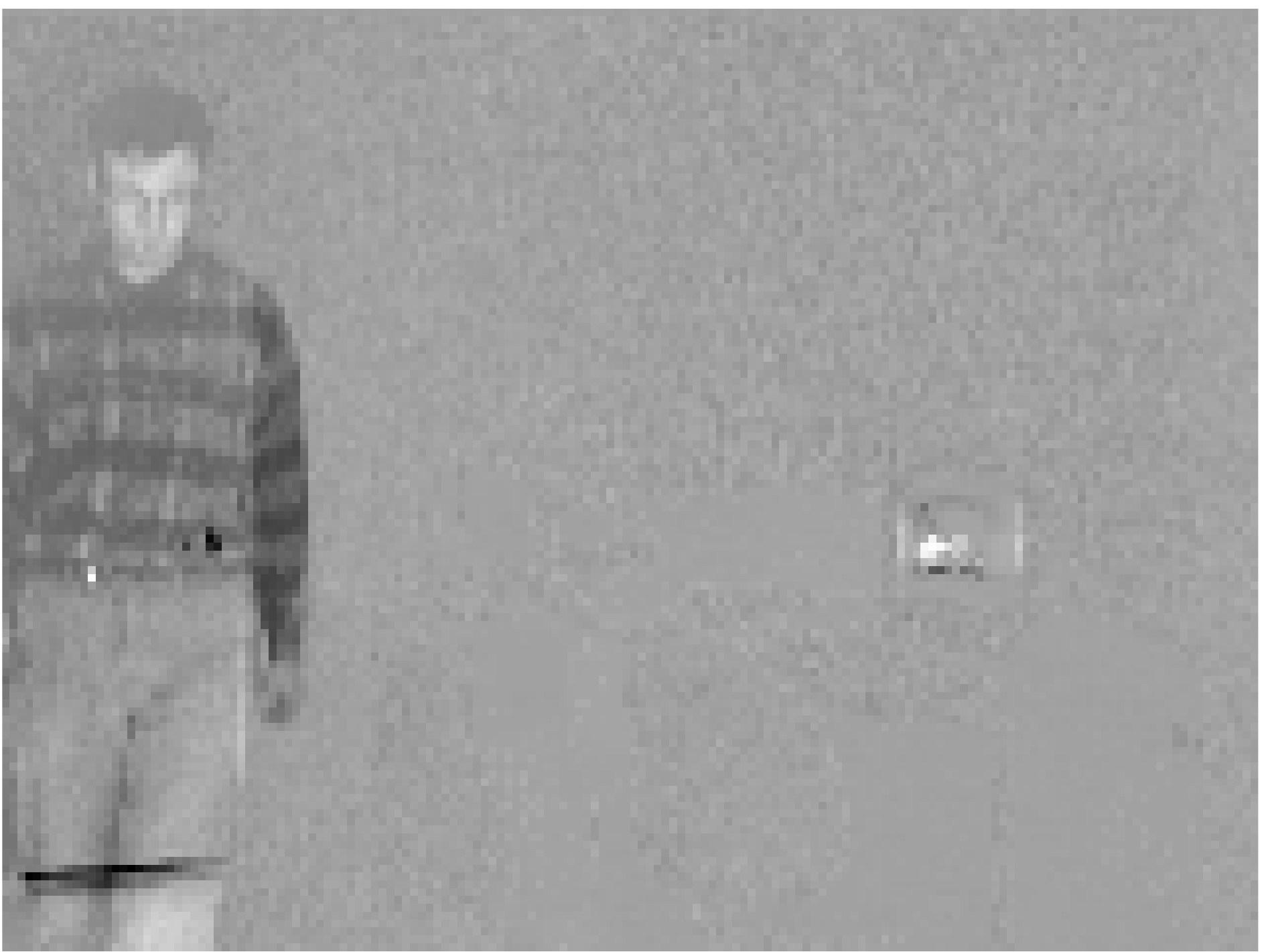} }		\hspace{-1mm}
{\includegraphics[width=0.18\columnwidth]{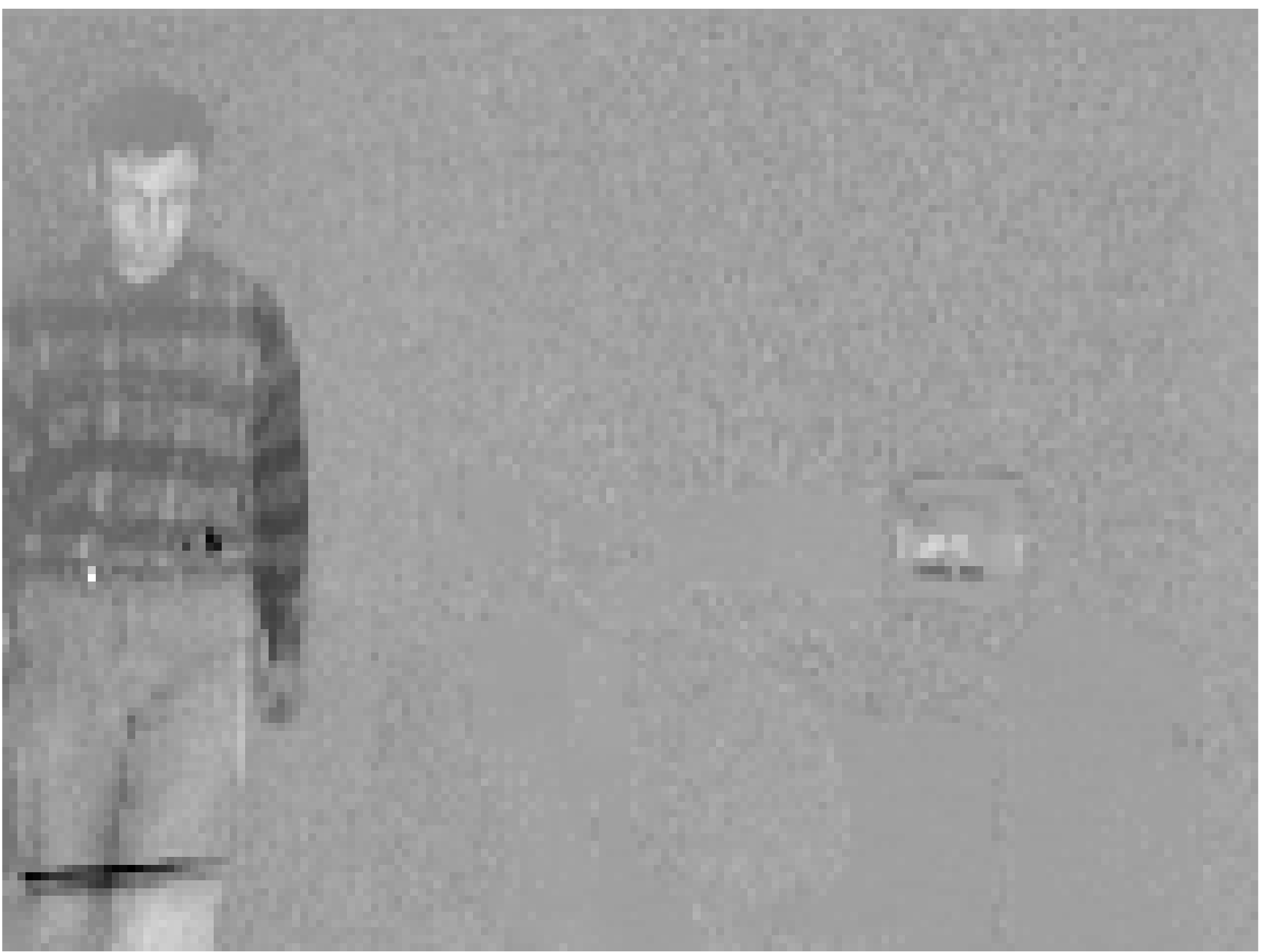} }	\hspace{-1mm}
{\includegraphics[width=0.18\columnwidth]{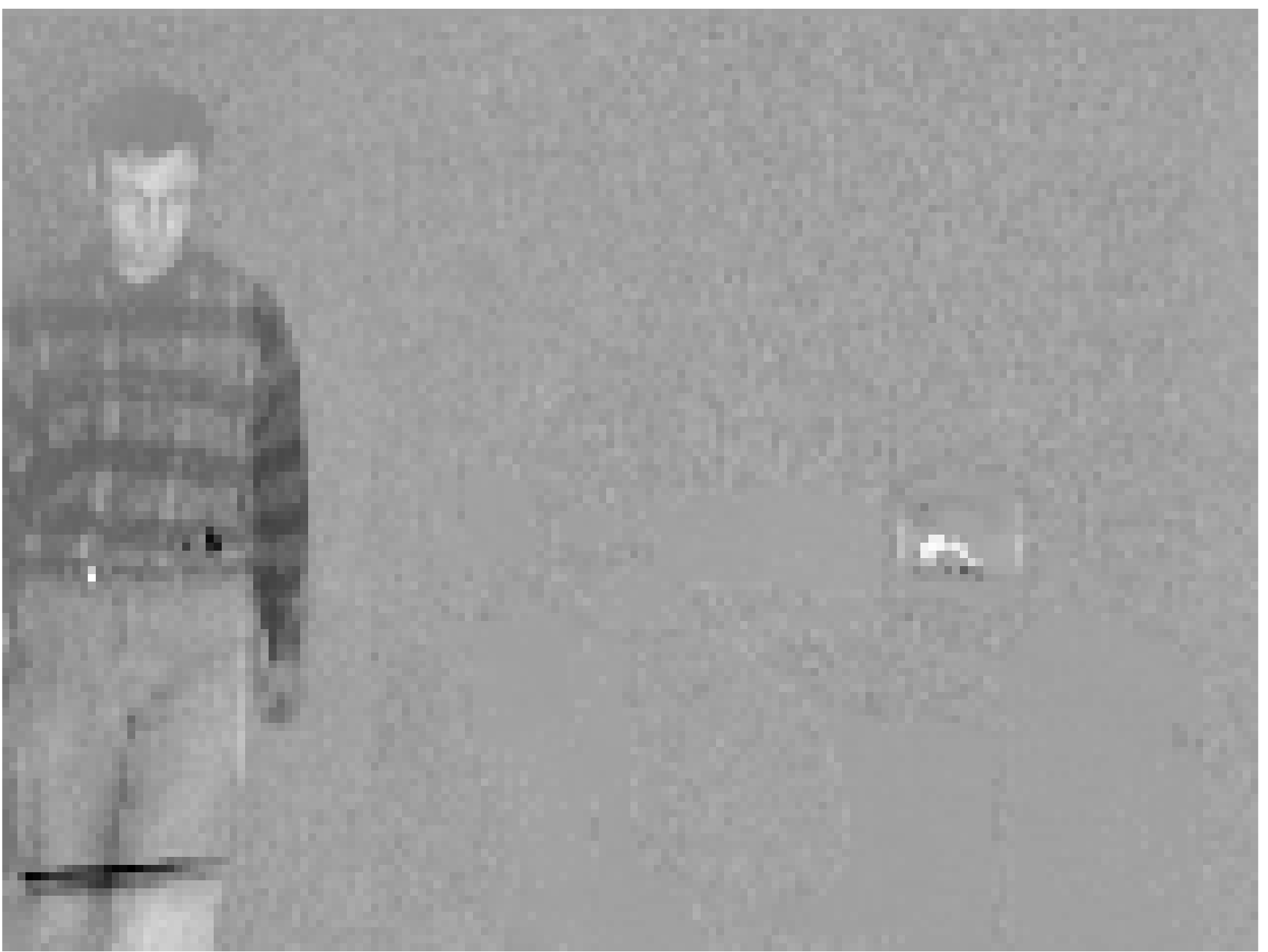} }	\hspace{-1mm}
{\includegraphics[width=0.18\columnwidth]{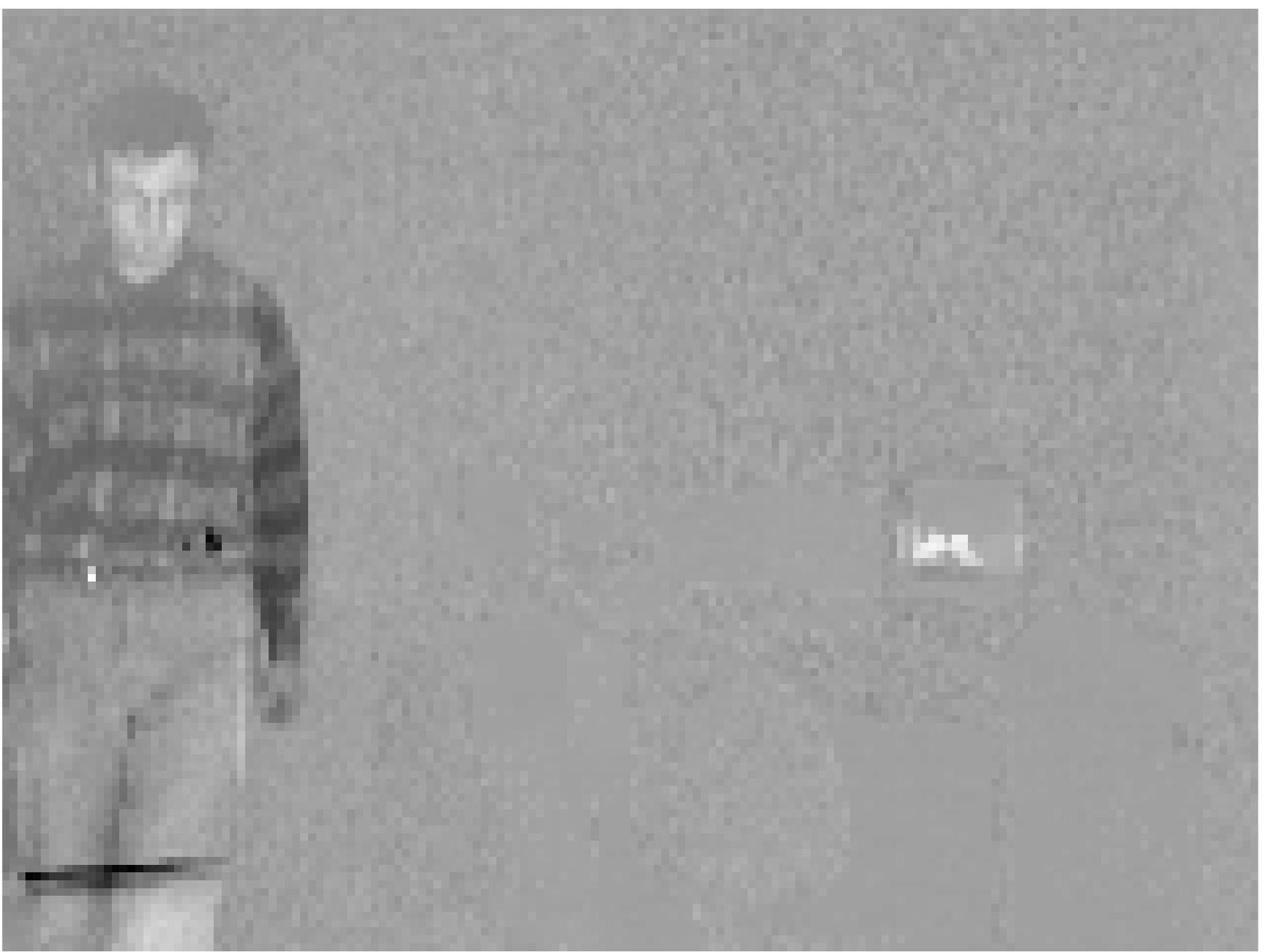} }			\\

\vspace{2mm}

{\includegraphics[width=0.18\columnwidth]{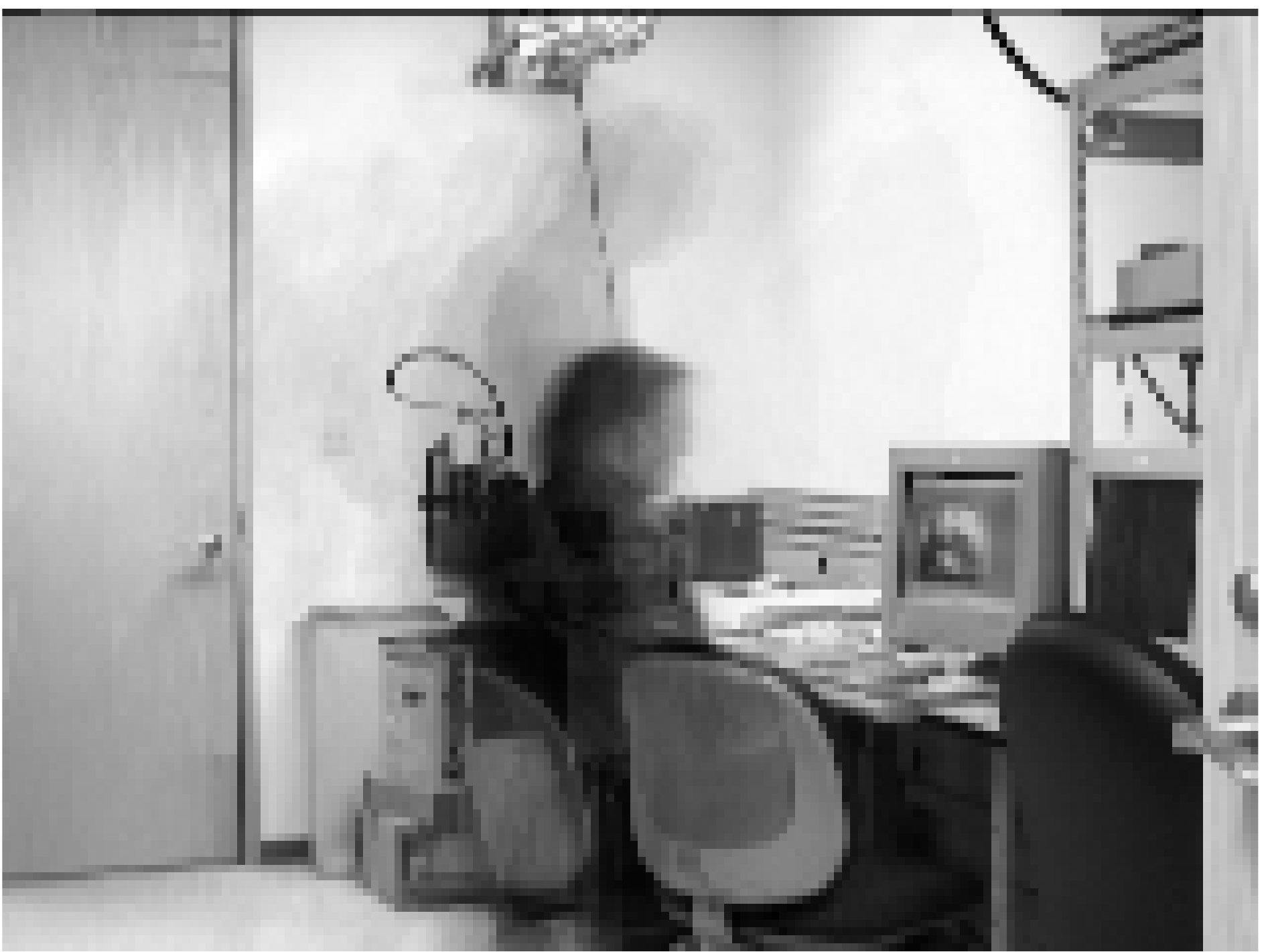} }	\hspace{-1mm}
{\includegraphics[width=0.18\columnwidth]{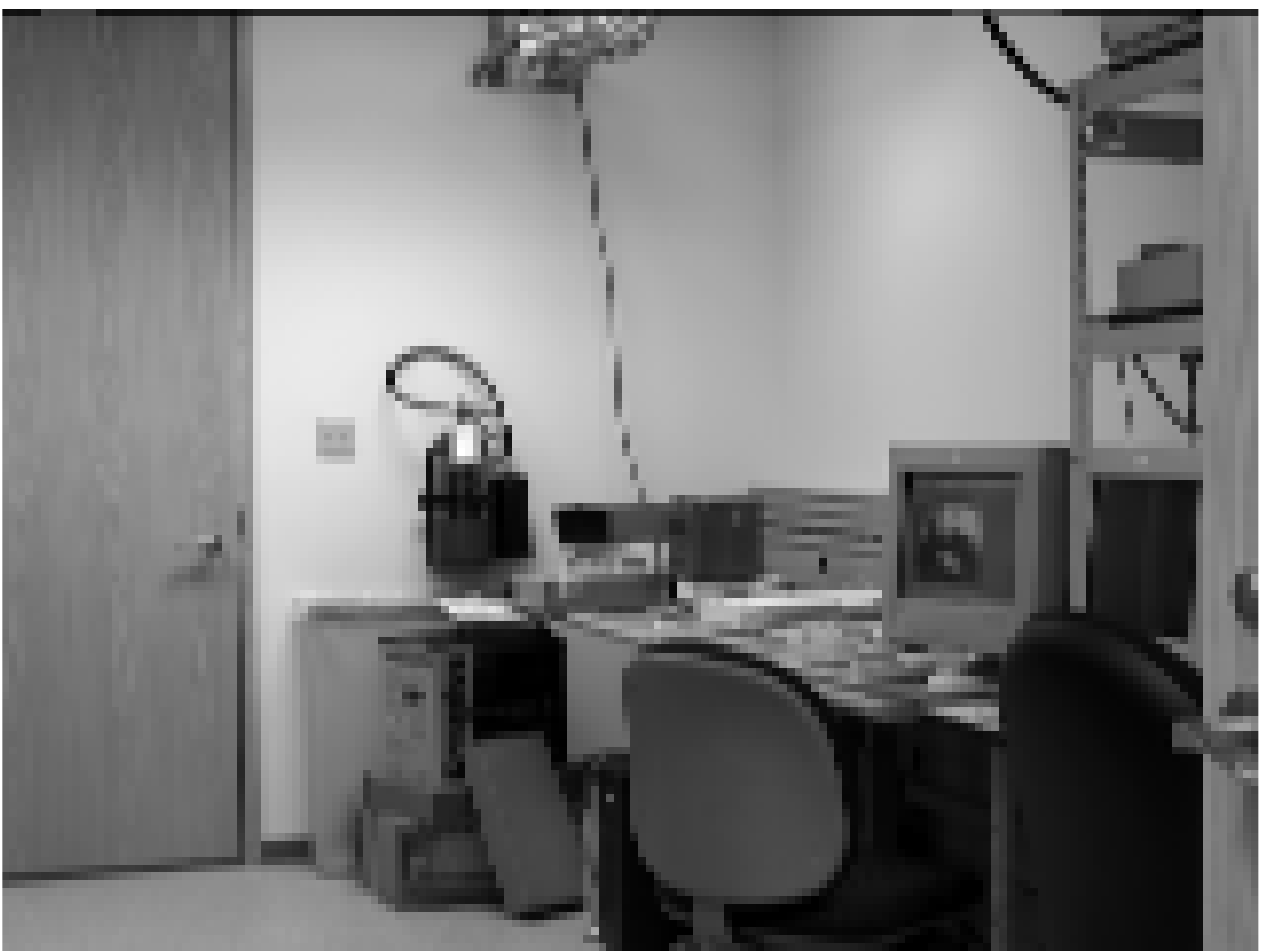} }	\hspace{-1mm}
{\includegraphics[width=0.18\columnwidth]{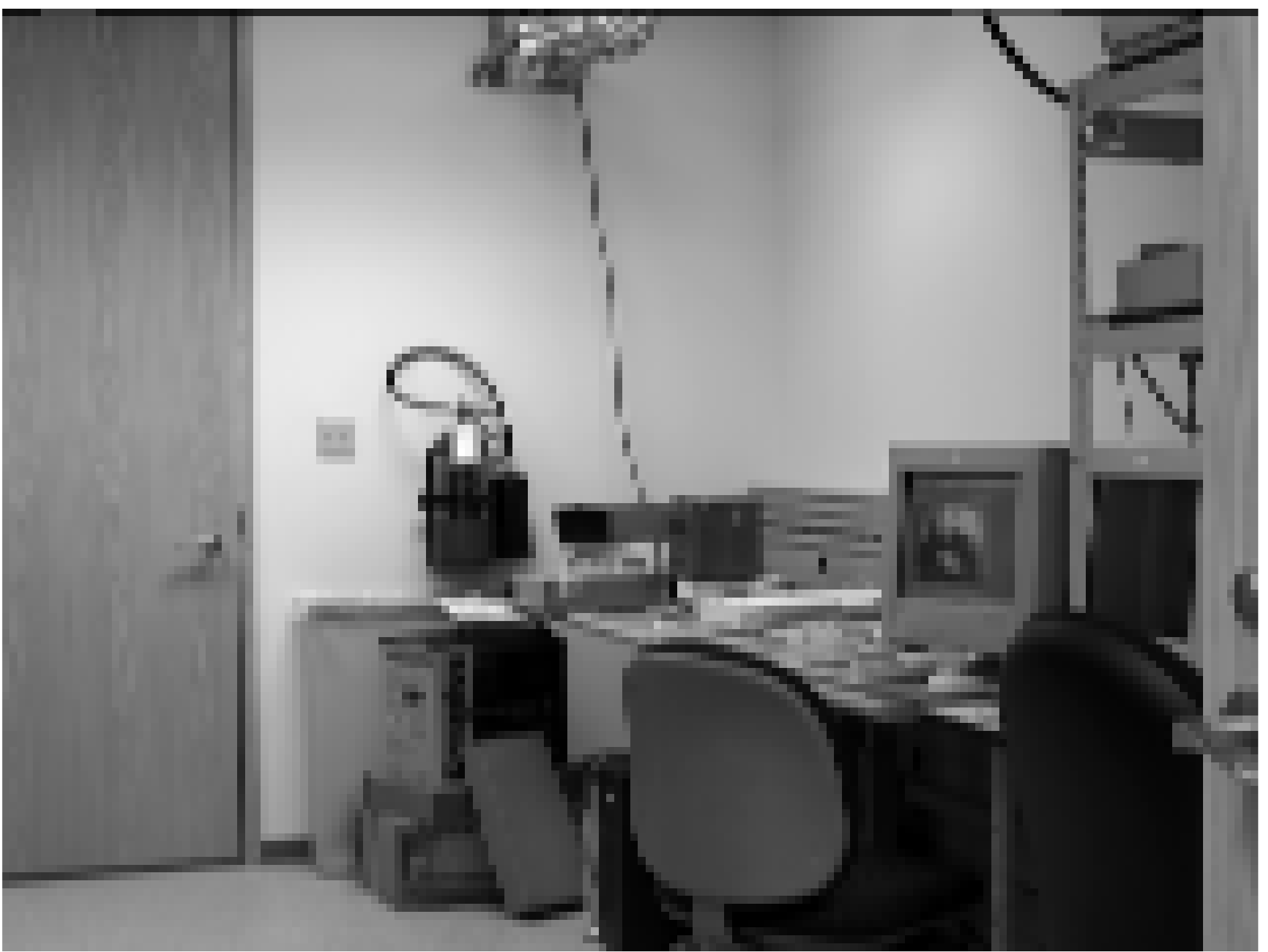} }	\hspace{-1mm}
{\includegraphics[width=0.18\columnwidth]{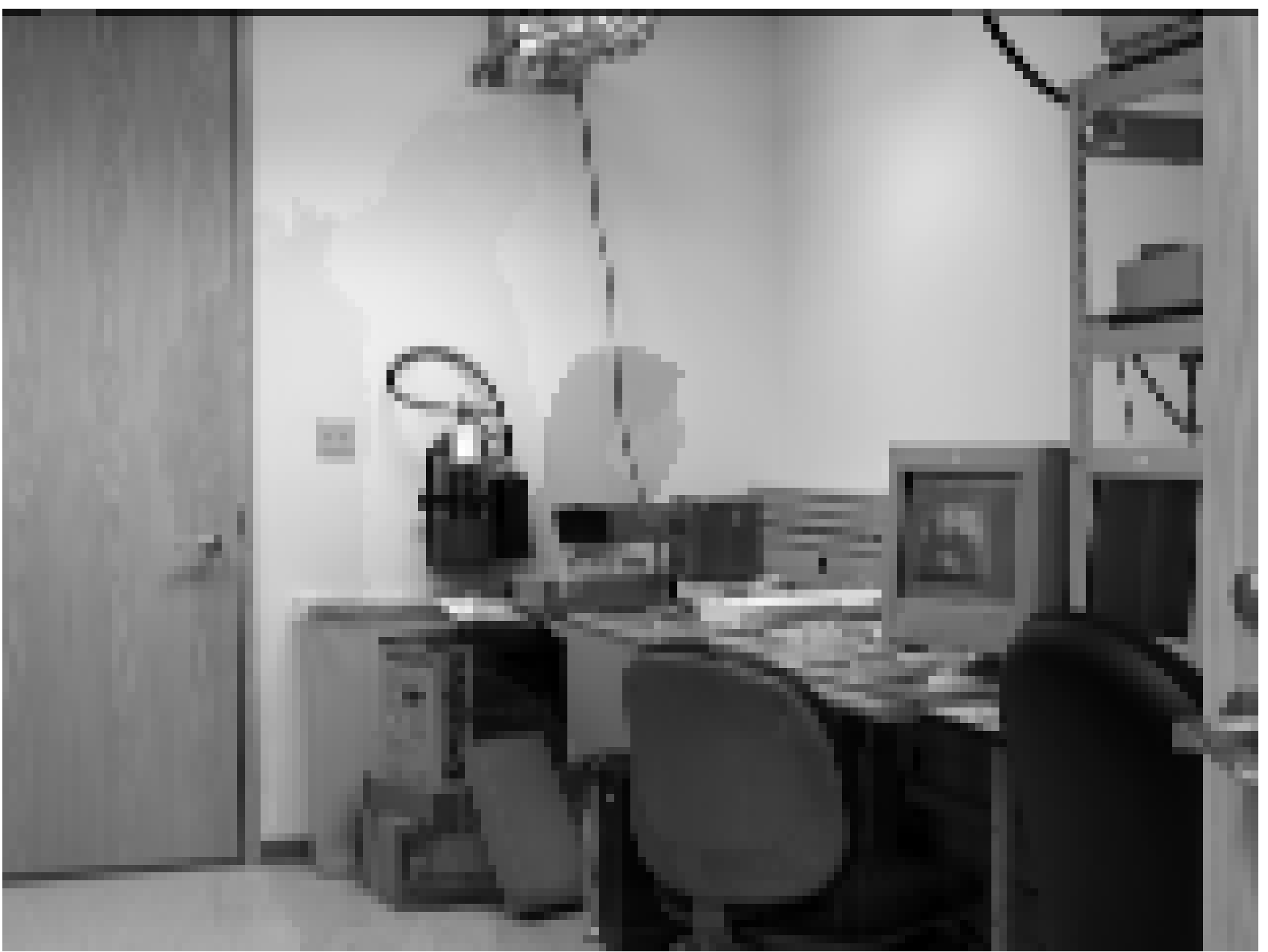} }		\hspace{-1mm}
{\includegraphics[width=0.18\columnwidth]{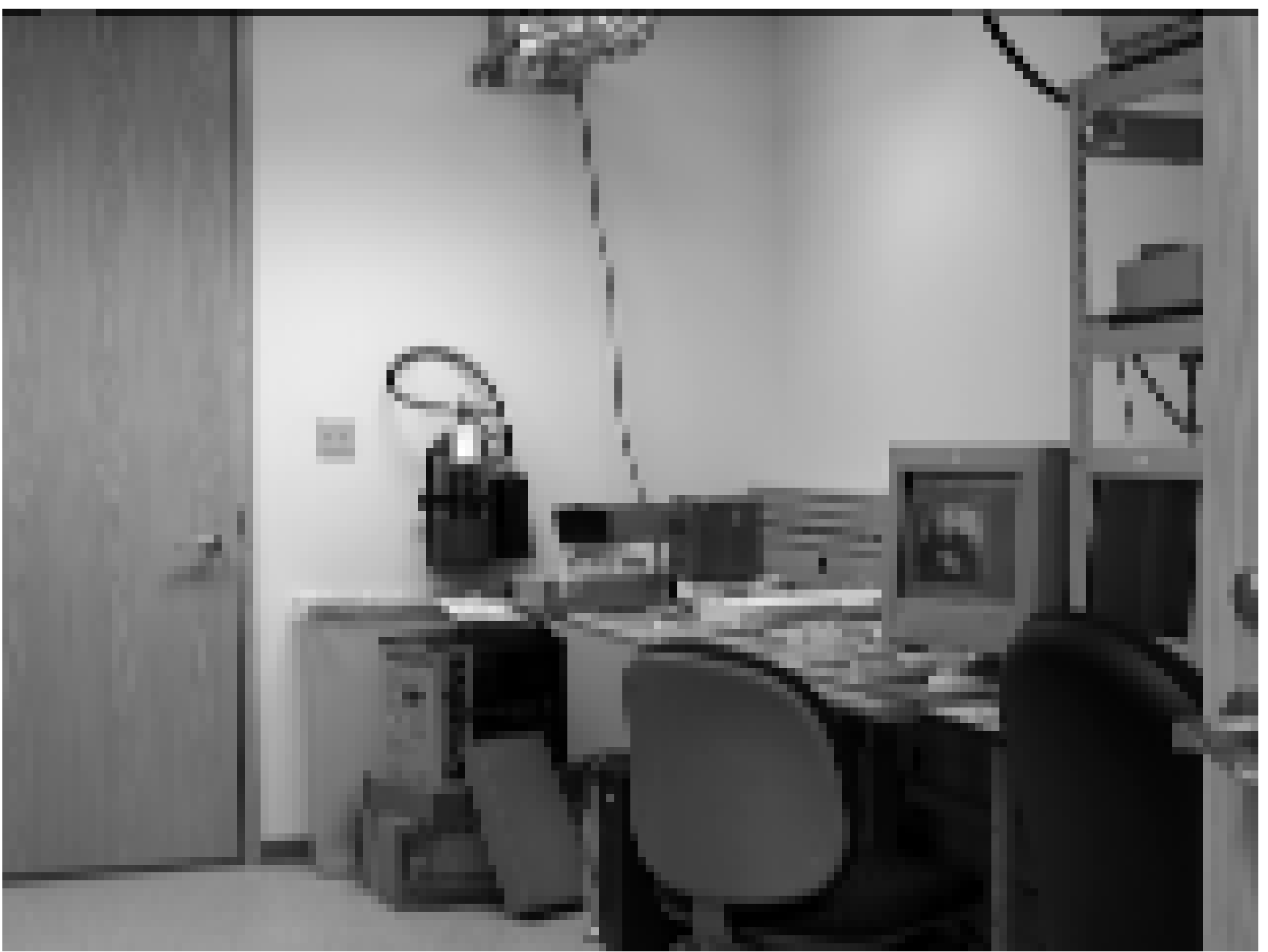} }			\\	
	
{\includegraphics[width=0.18\columnwidth]{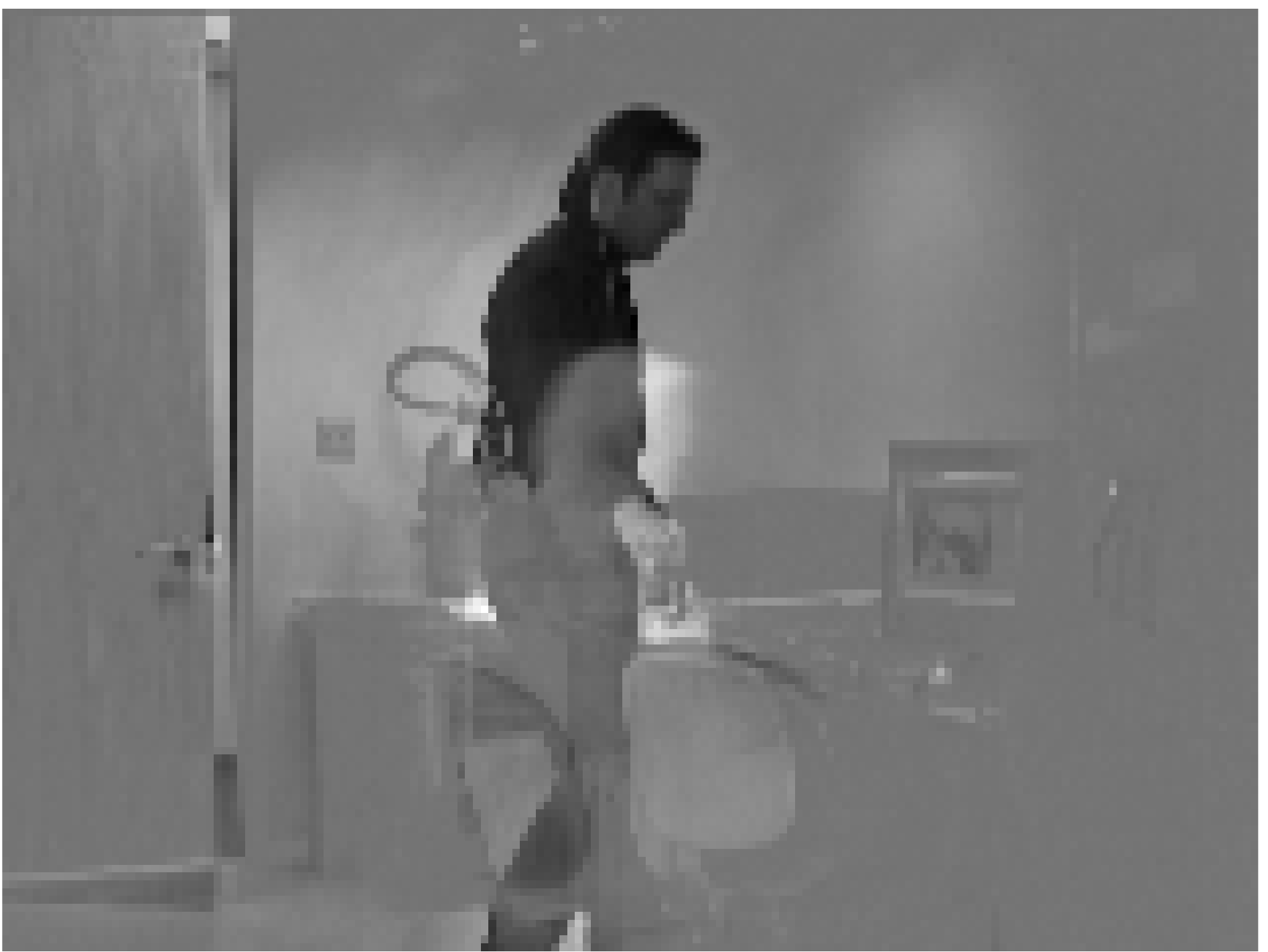} }	\hspace{-1mm}
{\includegraphics[width=0.18\columnwidth]{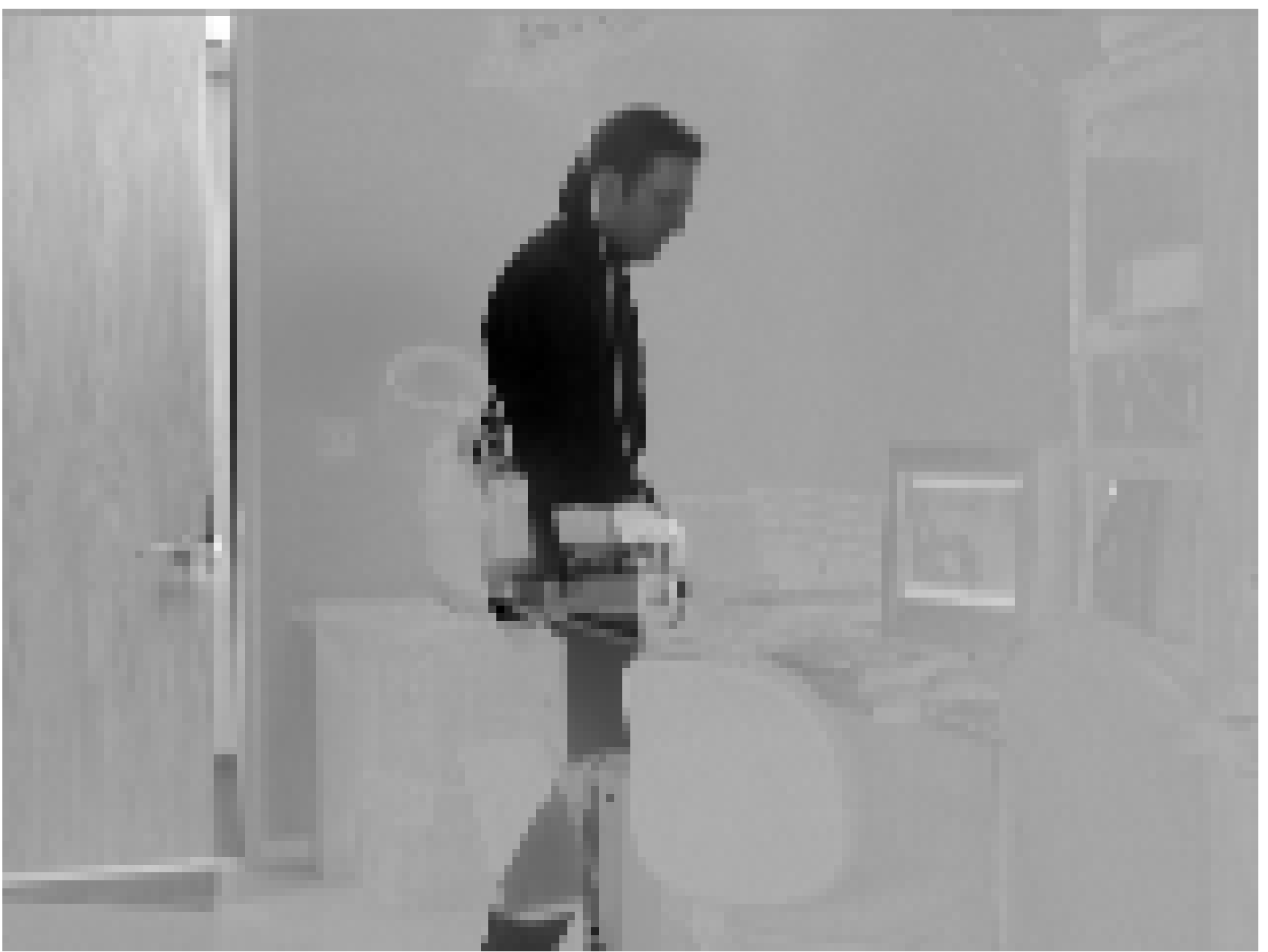} }	\hspace{-1mm}
{\includegraphics[width=0.18\columnwidth]{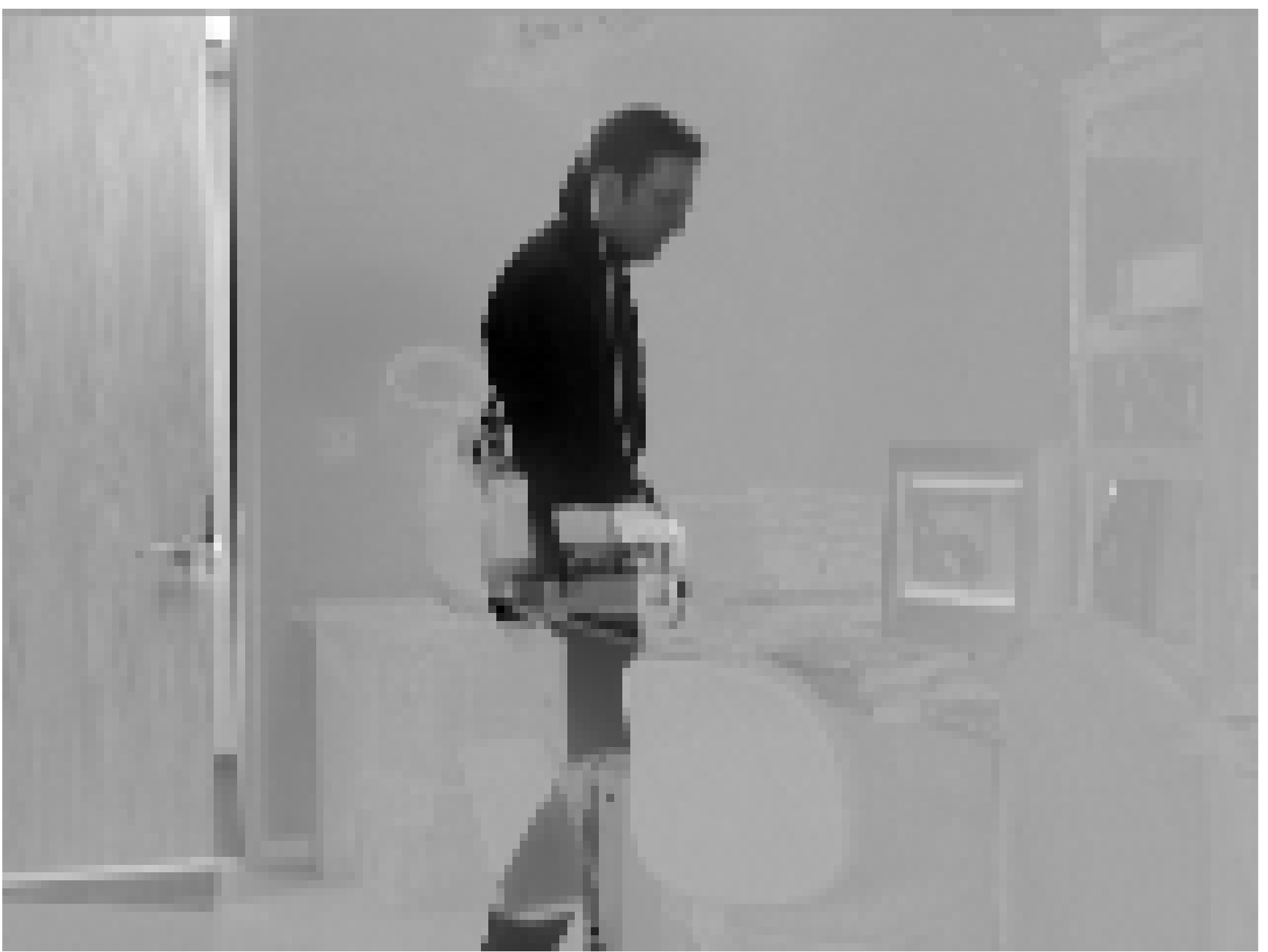} }	\hspace{-1mm}
{\includegraphics[width=0.18\columnwidth]{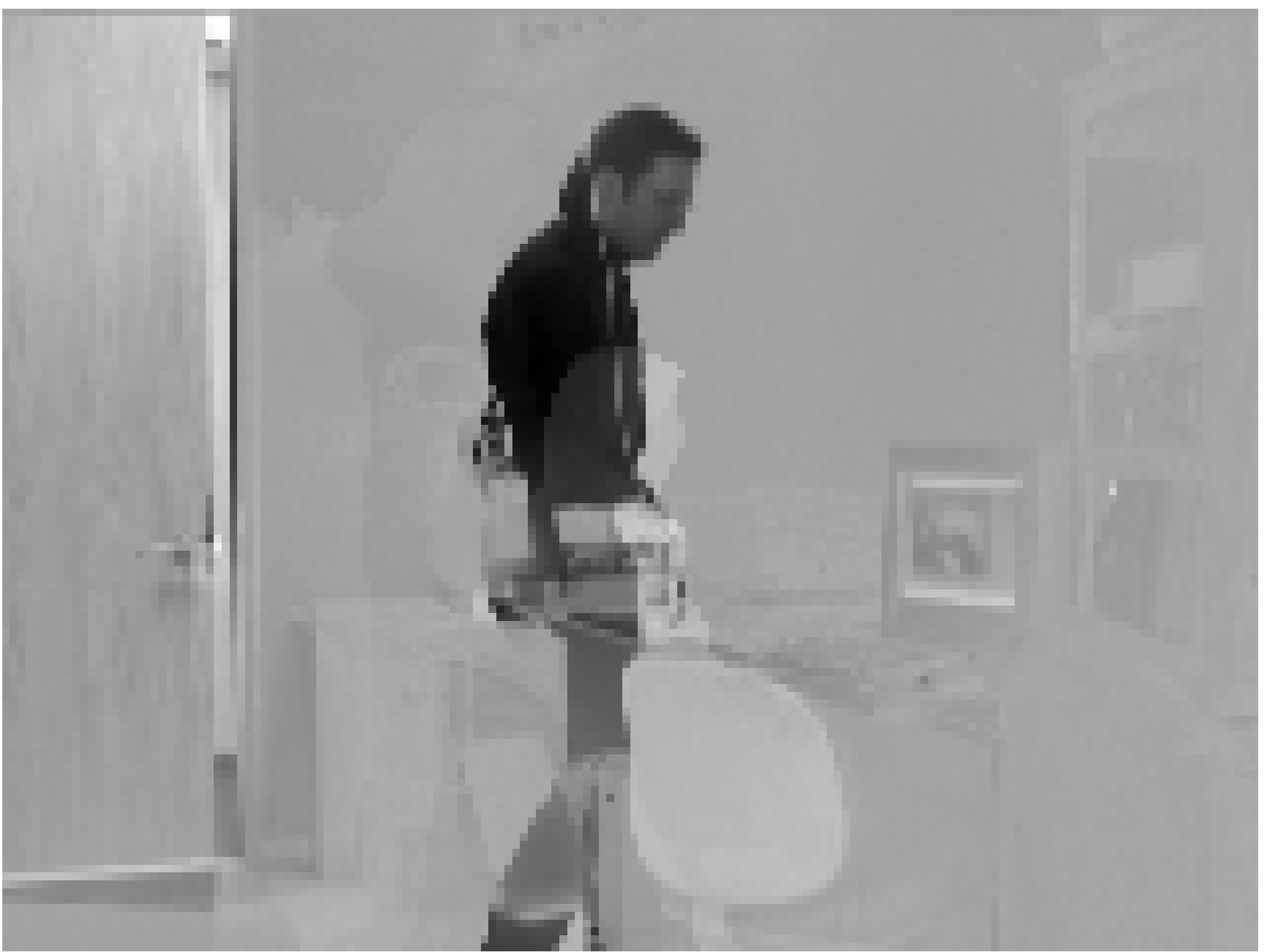} }	\hspace{-1mm}
{\includegraphics[width=0.18\columnwidth]{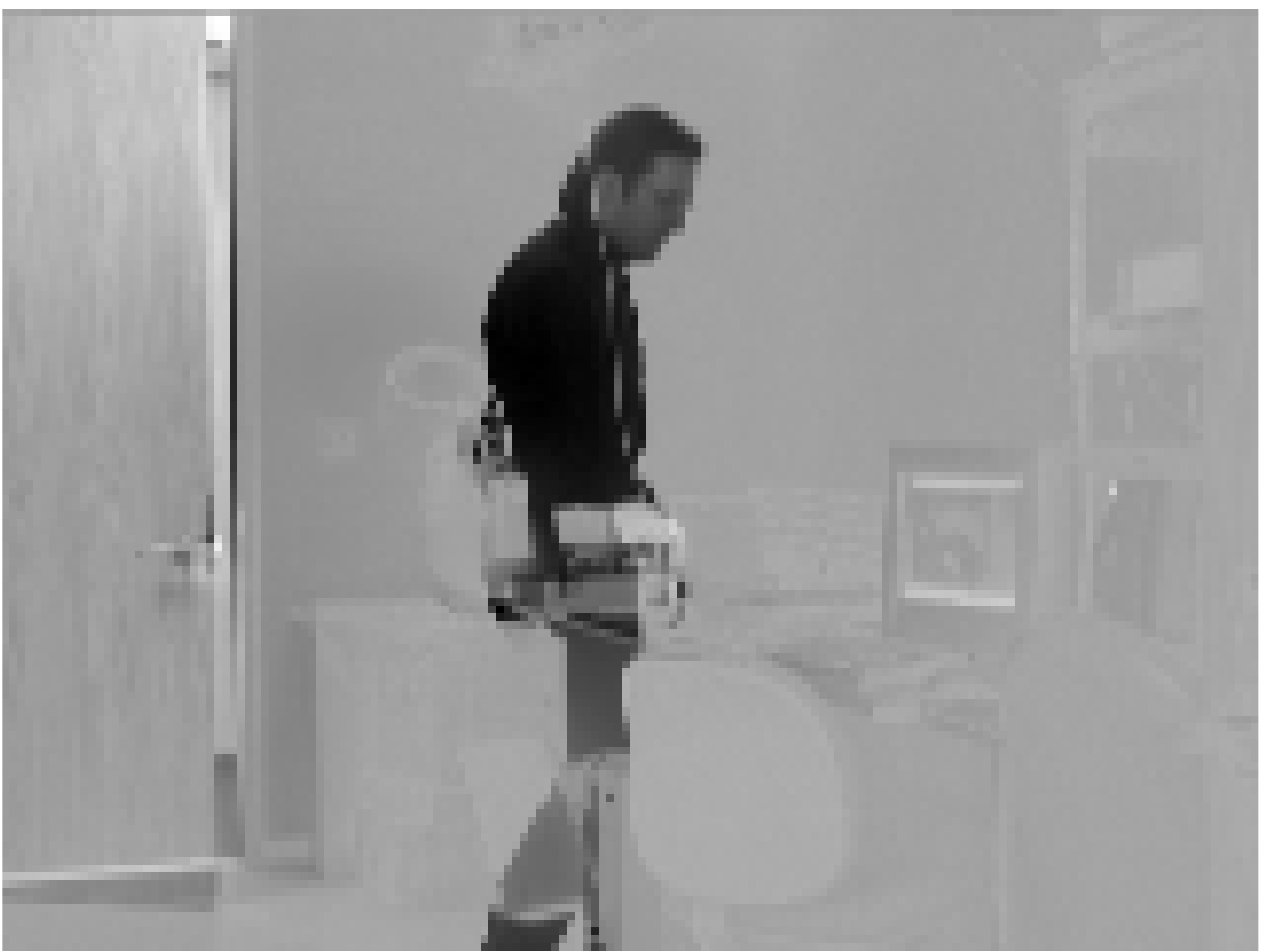} }			\\

\caption{ \footnotesize Foreground-background separation in the Highway and Light Switch-2 videos corresponding to (1)-(3) in Figure \ref{fig_sep_frame}, respectively. From left to right are results of IALM, AltProj (with $k$), F-FFP, AltProj (with $r=5$), and U-FFP, respectively. For every two panels that are grouped together, the top and bottom are recovered background and foreground, respectively. }
\label{fig_sep}
\end{figure}

We also show some video frames in Figure \ref{fig_sep_frame} and the visual results in Figure \ref{fig_sep}. It is observed that the backgrounds recovered by IALM still have shadows; AltProj separates foreground and background well with $k$ known, but the backgrounds are not clean when $k$ is unknown; both F-FFP and U-FFP can successfully separate foreground and background from the video.

\begin{table}[h]
\huge
\centering
\caption{ Results on Datasets with $r$ Unknown }
\resizebox{1.0\columnwidth}{!}{
\begin{tabular}{|c||c|| c |c |c |c | c| c| }
\hline		
Data 	
& Method	& Rank($L$) & ${\|S\|_0}/{(d n)}$ 	& $\frac{\|X-L-S\|_F}{\|X\|_F}$	& \# of Iter. 	& \# of SVDs	& Time	\\ \hline	

\multirow{2}{3cm}{ Highway } 	
& AltProj	& 5			& 0.9007		& 3.66e-4		& 43			& 48			& 75.60		\\	\cline{2-8}
& U-FFP		& 1			& 0.8854 		& 5.75e-4		& 24			& 24+24+24		& 18.02		\\	\hline\hline

\multirow{2}{3cm}{ Office } 	
& AltProj	& 5			& 0.7159		& 8.61e-4		& 47			& 52			& 98.54		\\	\cline{2-8}
& U-FFP		& 1			& 0.8761 		& 5.40e-4		& 24			& 24+24+24		& 24.40		\\	\hline\hline

\multirow{2}{3cm}{ PETS2006 } 	
& AltProj	& 5			& 0.8543		& 6.15e-4		& 39			& 43			& 63.33		\\	\cline{2-8}
& U-FFP		& 1			& 0.8675		& 5.61e-4		& 24			& 24+24+24		& 17.29		\\	\hline\hline

\multirow{2}{3cm}{ Shopping Mall } 	
& AltProj	& 5			& 0.9611		& 9.82e-5		& 41			& 46			& 63.34		\\	\cline{2-8}
& U-FFP		& 1			& 0.9122		& 7.70e-4		& 23			& 23+23+23		& 14.37		\\	\hline\hline

\multirow{2}{3cm}{ Pedestrian } 	
& AltProj	& 5			& 0.6202		& 6.37e-4		& 44			& 49			& 58.10		\\	\cline{2-8}
& U-FFP		& 1			& 0.6714 		& 5.65e-4		& 23			& 23+23+23		& 12.40		\\	\hline\hline

\multirow{2}{3cm}{ Bootstrap } 	
& AltProj	& 5			& 0.9875		& 3.02e-4		& 47			& 52			& 169.06	\\	\cline{2-8}
& U-FFP		& 1			& 0.9288		& 7.70e-4		& 23			& 23+23+23		& 31.03		\\	\hline\hline

\multirow{2}{3cm}{Water Surface} 	
& AltProj	& 5			& 0.9090		& 2.38e-4		& 46			& 50			& 33.78		\\	\cline{2-8}
& U-FFP		& 1			& 0.8355		& 9.77e-4		& 23			& 23+23+23		& 7.27		\\	\hline\hline

\multirow{2}{3cm}{ Campus } 	
& AltProj	& 5			& 0.9482		& 3.18e-5		& 46			& 51			& 92.90		\\	\cline{2-8}
& U-FFP		& 1			& 0.9377 		& 6.26e-4		& 23			& 23+23+24		& 16.29		\\	\hline\hline

\multirow{2}{3cm}{ Curtain } 	
& AltProj	& 5			& 0.8079		& 8.82e-4		& 36			& 39			& 101.79	\\	\cline{2-8}
& U-FFP		& 1			& 0.8680 		& 6.28e-4		& 24			& 24+24			& 34.33 	\\	\hline\hline

\multirow{2}{3cm}{ Fountain } 	
& AltProj	& 5			& 0.7435		& 7.55e-4		& 48			& 52			& 32.24		\\	\cline{2-8}
& U-FFP		& 1			& 0.8852		& 4.91e-4		& 24			& 24+24+24		& 6.26		\\	\hline\hline

\multirow{2}{3cm}{ Escalator Airport } 	
& AltProj	& 5			& 0.8474		& 8.43e-4		& 43			& 48			& 162.49	\\	\cline{2-8}
& U-FFP		& 1			& 0.8877		& 5.45e-4		& 23			& 23+23+23		& 39.70		\\	\hline	\hline

\multirow{2}{3cm}{Lobby} 	
& AltProj	& 5			& 0.9176		& 1.71e-4		& 40			& 44			& 61.50		\\	\cline{2-8}
& U-FFP		& 2			& 0.8523		& 6.42e-4		& 24			& 24+24+24		& 17.91		\\	\hline\hline

\multirow{2}{3cm}{Light Switch-2} 	
& AltProj	& 5			& 0.8507		& 4.29e-4		& 37			& 41			& 80.37	\\	\cline{2-8}
& U-FFP		& 2			& 0.8329		& 7.57e-4		& 24			& 24+24+24		& 28.53	\\	\hline\hline

\multirow{2}{3cm}{Camera Parameter} 	
& AltProj	& 5			& 0.7311		& 8.34e-4		& 50			& 55			& 147.28	\\	\cline{2-8}
& U-FFP		& 2			& 0.8689		& 6.26e-4		& 24			& 24+24+24		& 26.35		\\	\hline\hline

\multirow{2}{3cm}{Time Of Day} 	
& AltProj	& 5			& 0.8651		& 4.61e-4		& 46			& 51			& 73.35		\\	\cline{2-8}
& U-FFP		& 2			& 0.8425		& 7.20e-4		& 25			& 25+25+25		& 21.83		\\	\hline
%---------------------
\end{tabular}
}
\scriptsize 
\begin{flushleft}
For AltProj, (partial) SVDs are performed on $d\times n$ matrices. For U-FFP, SVDs are for $d\times k$, $n\times k$, and $k\times k$ matrices, which are computationally far less expensive than those required by AltProj.
\end{flushleft}
\label{tab_unknown}
\end{table}

\subsection{Shadow removal from face images }
Learning face is an important topic in pattern recognition; however, shadows make it more challenging. There are often shadows on face images due to varying lighting conditions \cite{basri2003lambertian}. Therefore, it is crucial to handle shadows, peculiarities and saturations on face images, so as to improve the learning capability on face image data. This can be handled with RPCA, because clean images reside in a low-rank subspace while the shadows correspond to sparse components. Following \cite{kang2015robustpca,candes2011robust}, we use Extended Yale B (EYaleB) dataset \cite{georghiades2001few} for this test. Out of 38 persons in this dataset, we choose the first two subjects. For each subject, there are 64 heavily corrupted images of size $192\times 168$ taken under varying lighting conditions. Each image is vectorized into a column of a $32,256\times 64$ data matrix. It is natural to assume that the face images of the same person reside in a rank one space, hence we have $r=1$.

\begin{figure}[!tb]
\centering
\resizebox{1.0\columnwidth}{!}{
\begin{tabular}{c c c c c || c c c c c  }
\includegraphics[width=0.1\textwidth]{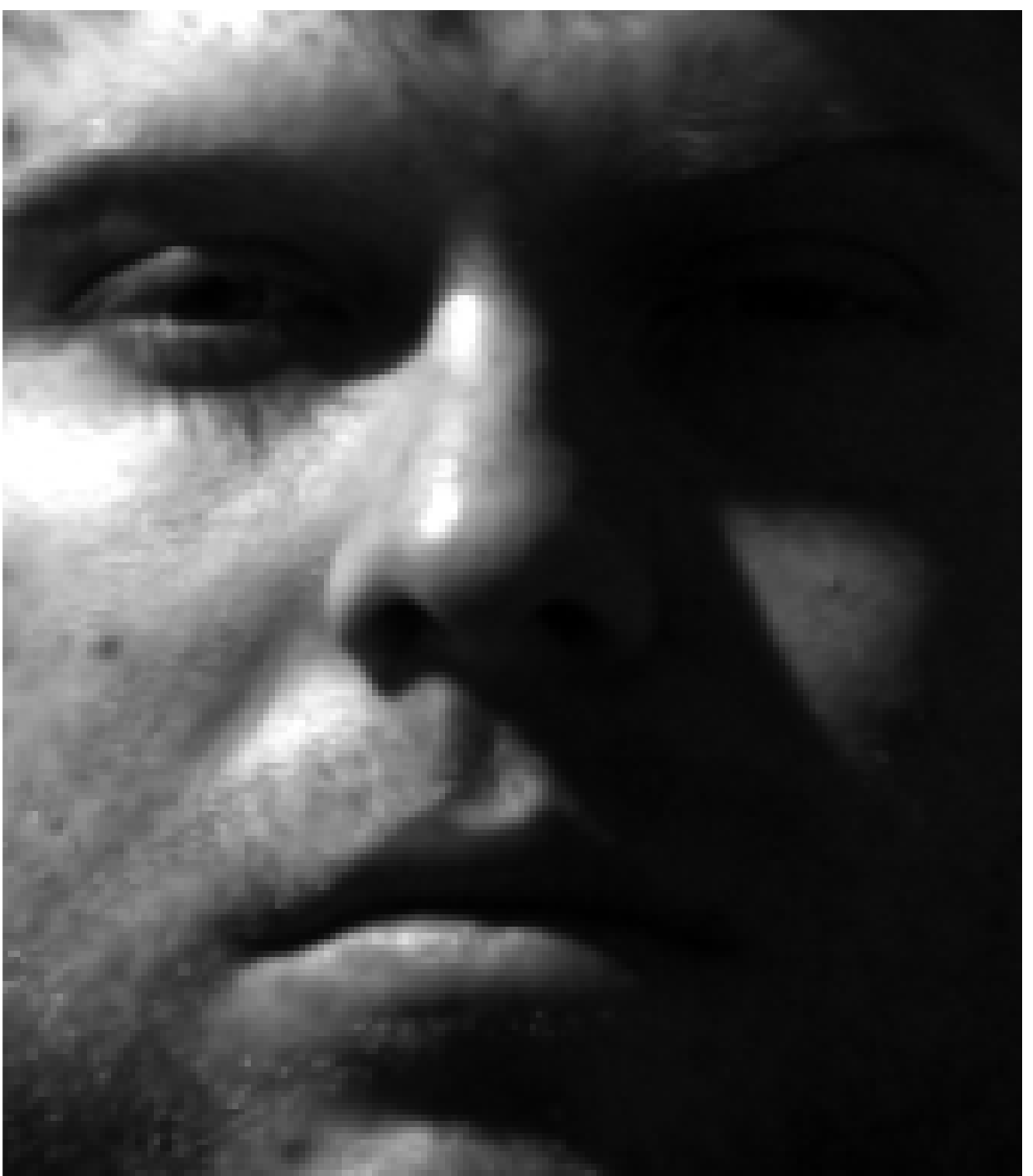}			&
\includegraphics[width=0.1\textwidth]{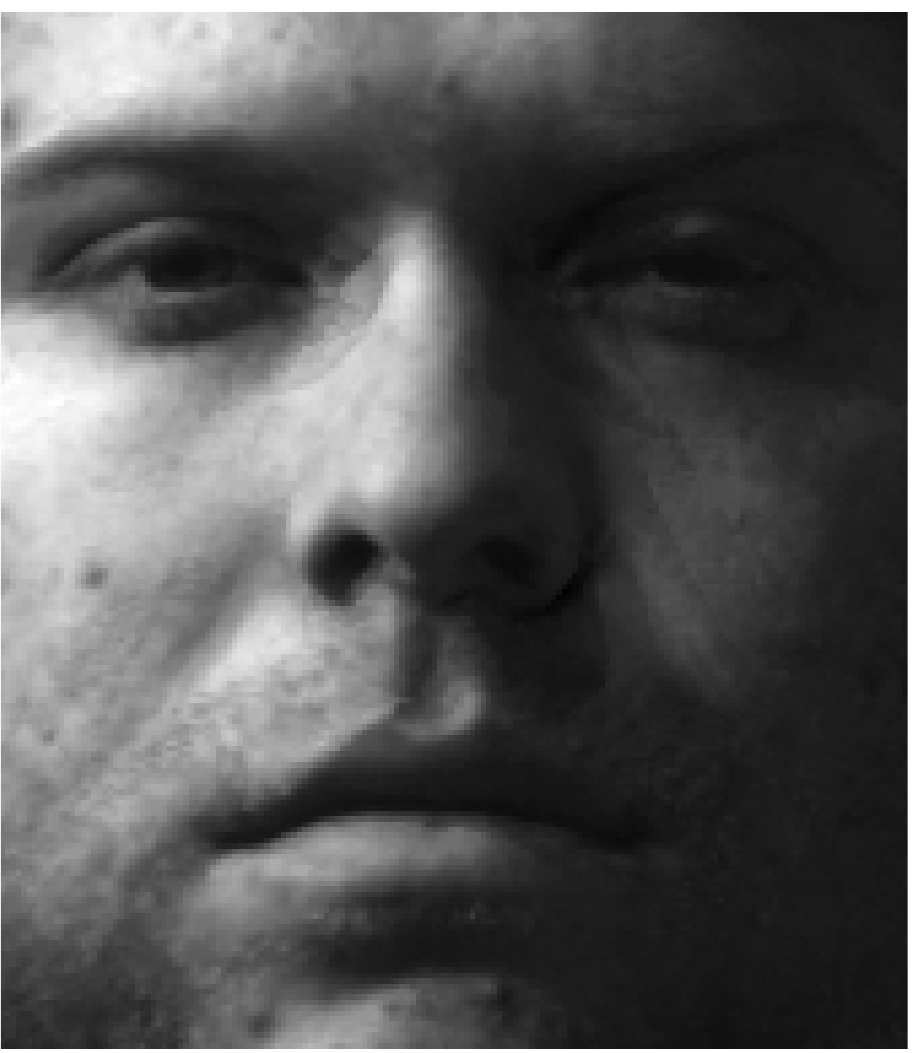}		&
\includegraphics[width=0.1\textwidth]{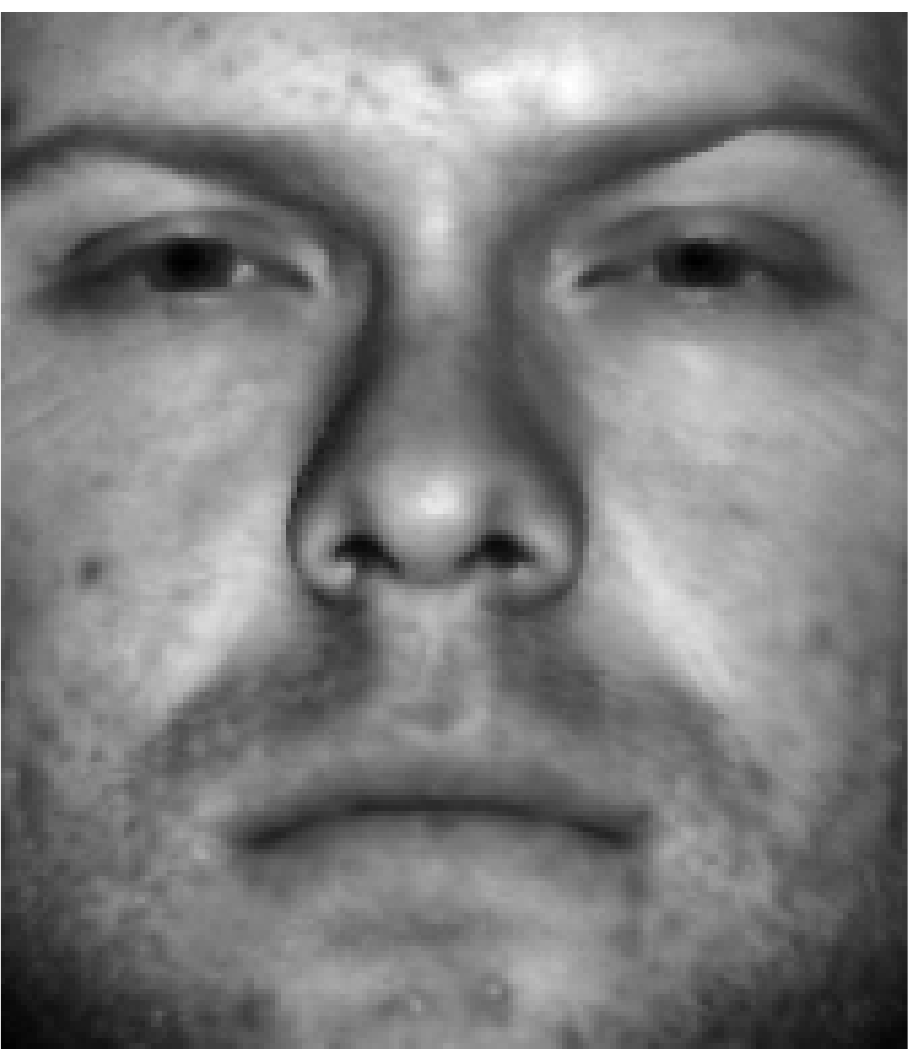}	&
\includegraphics[width=0.1\textwidth]{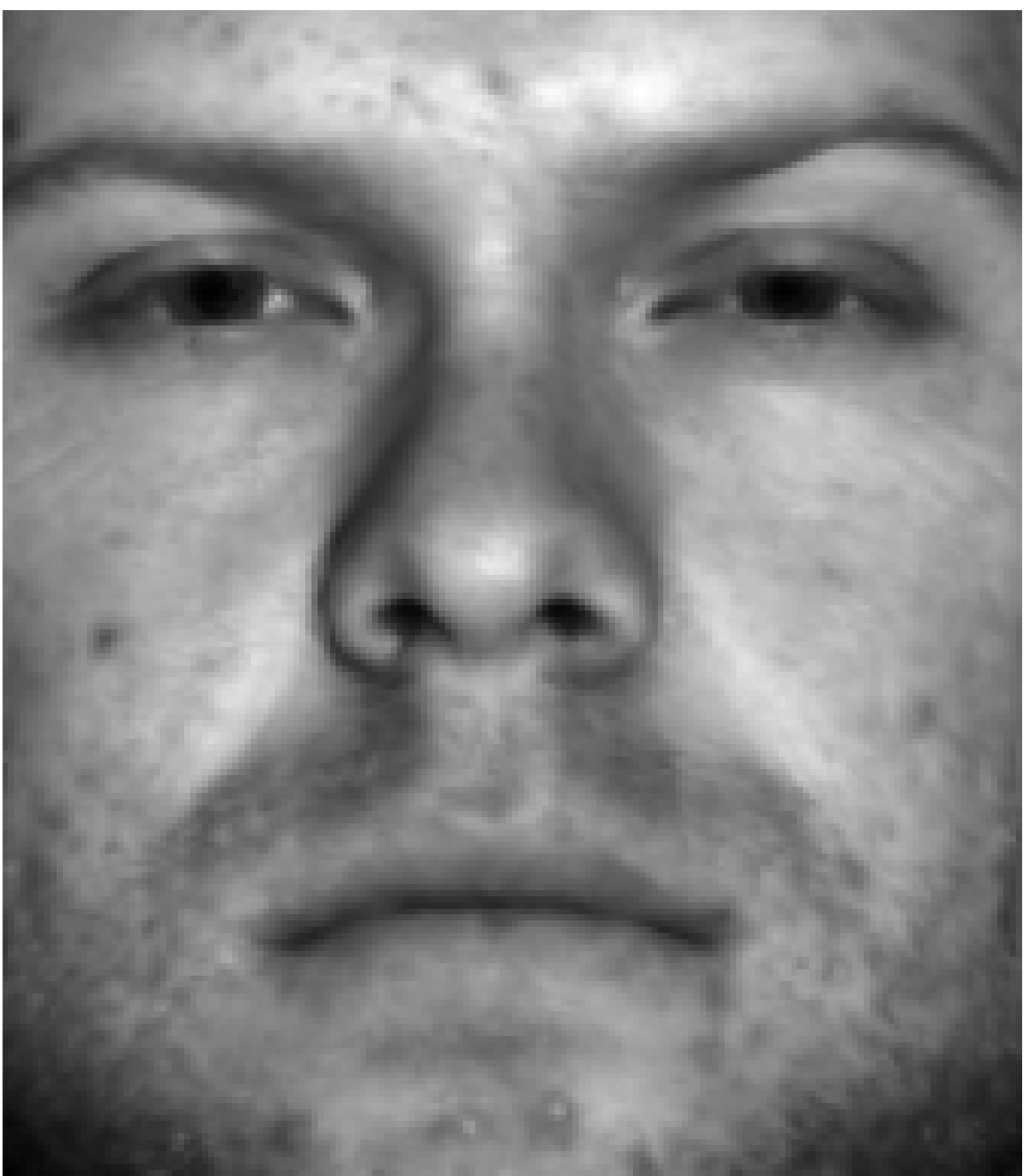}		&
&
& 
\includegraphics[width=0.1\textwidth]{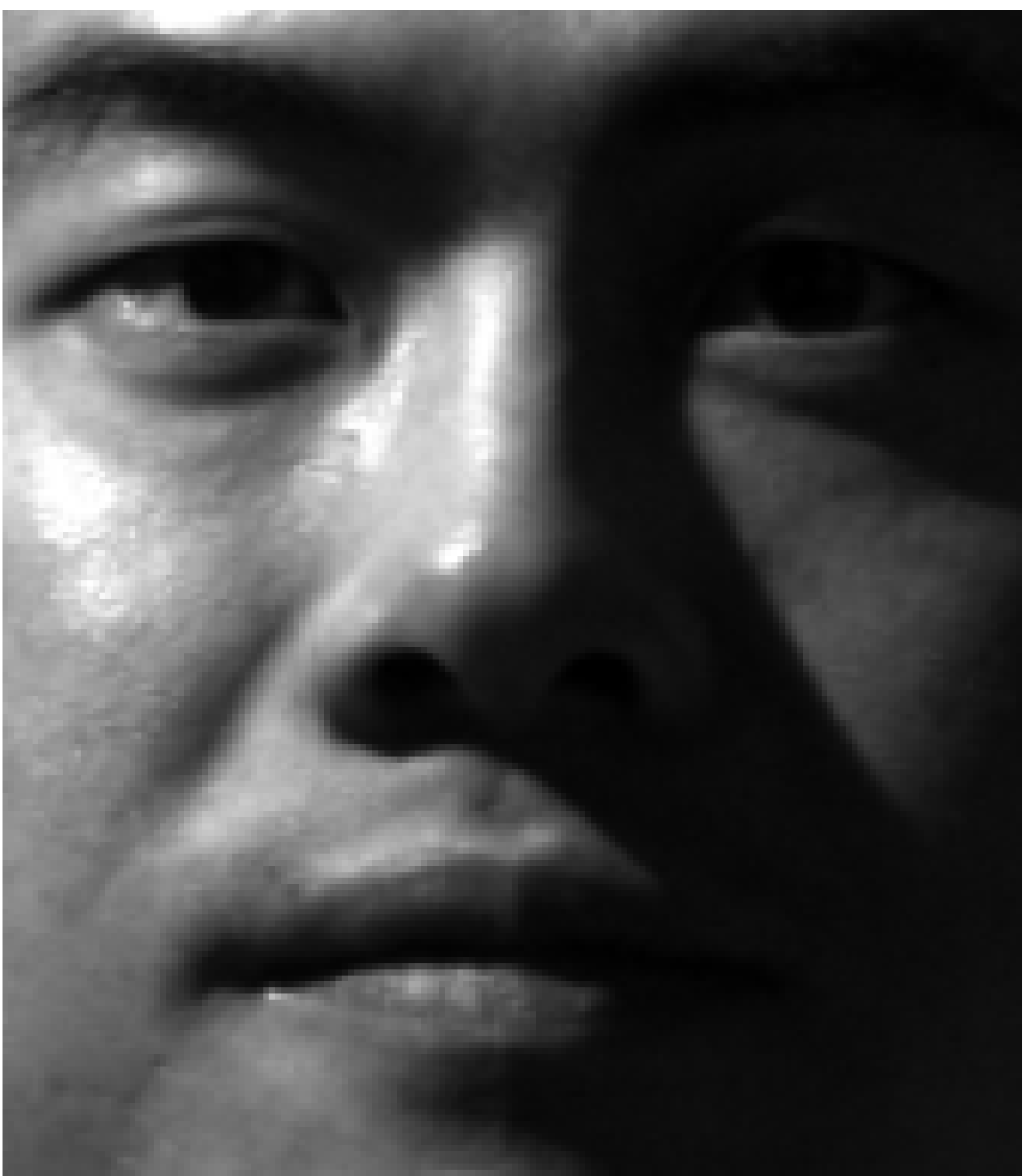}			&
\includegraphics[width=0.1\textwidth]{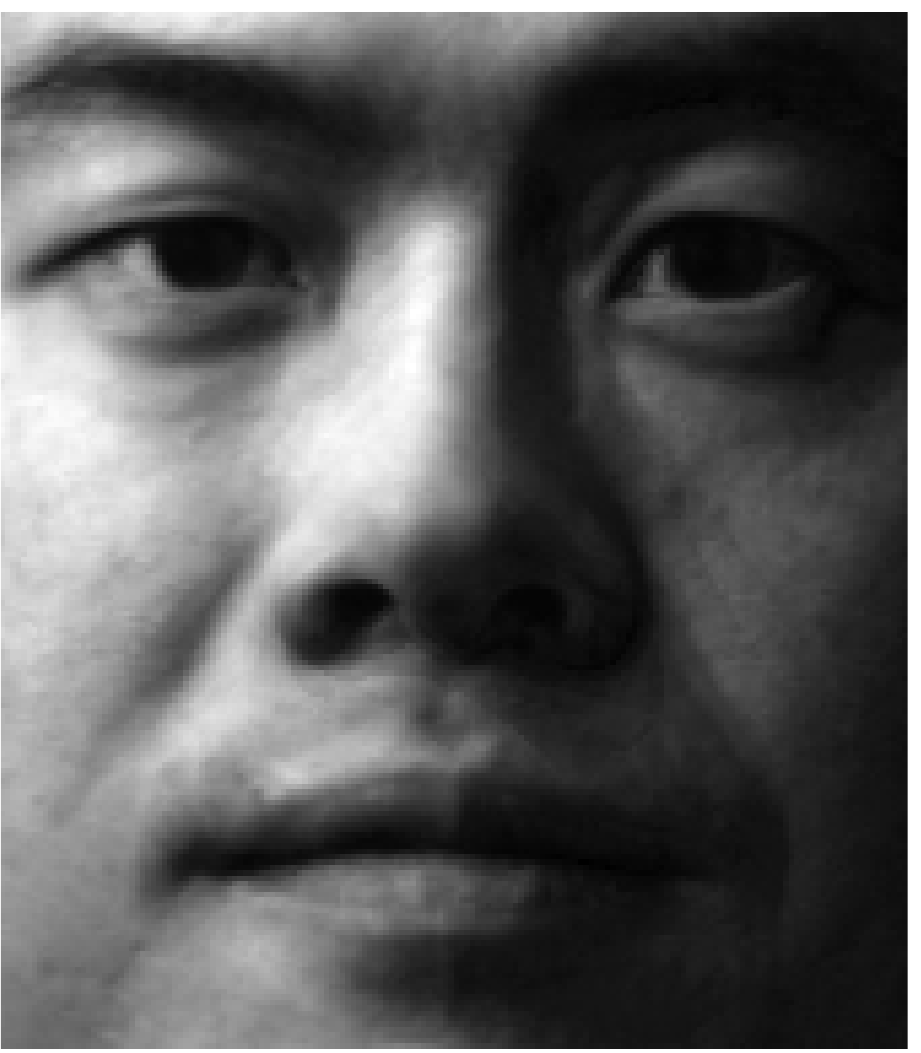}		&
\includegraphics[width=0.1\textwidth]{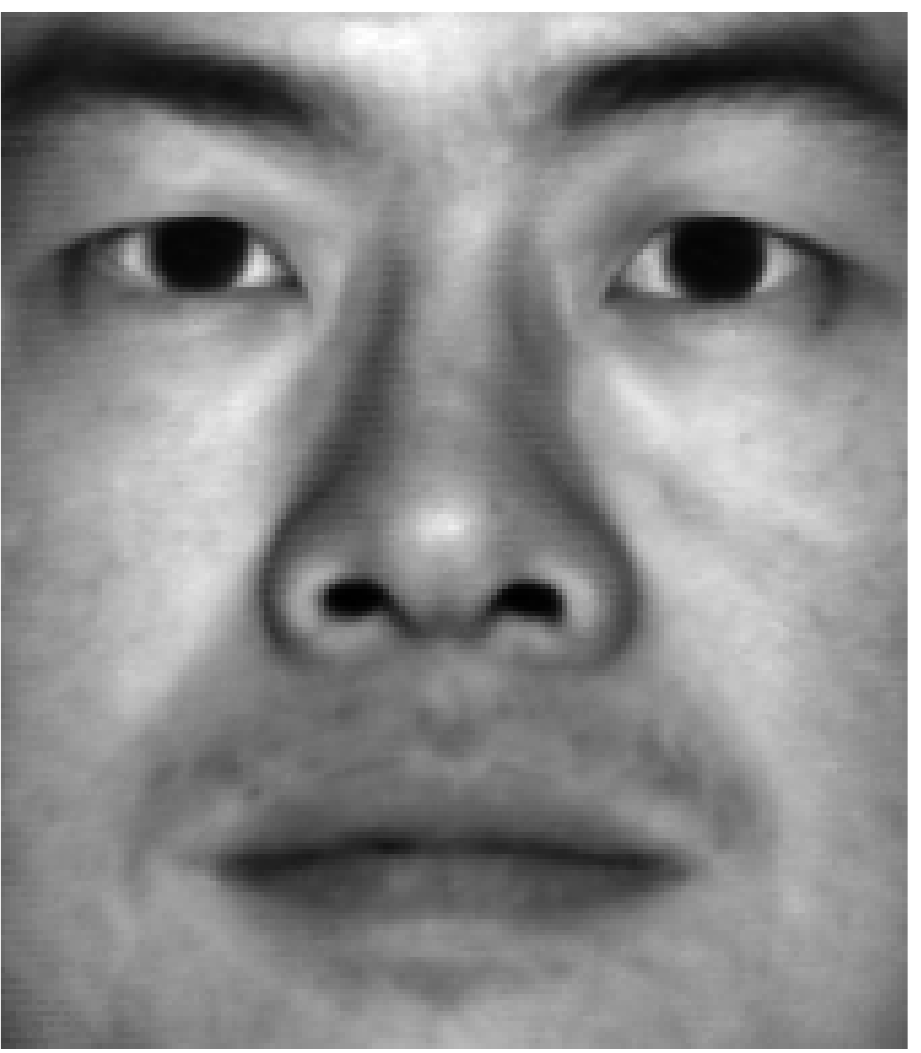}	&
\includegraphics[width=0.1\textwidth]{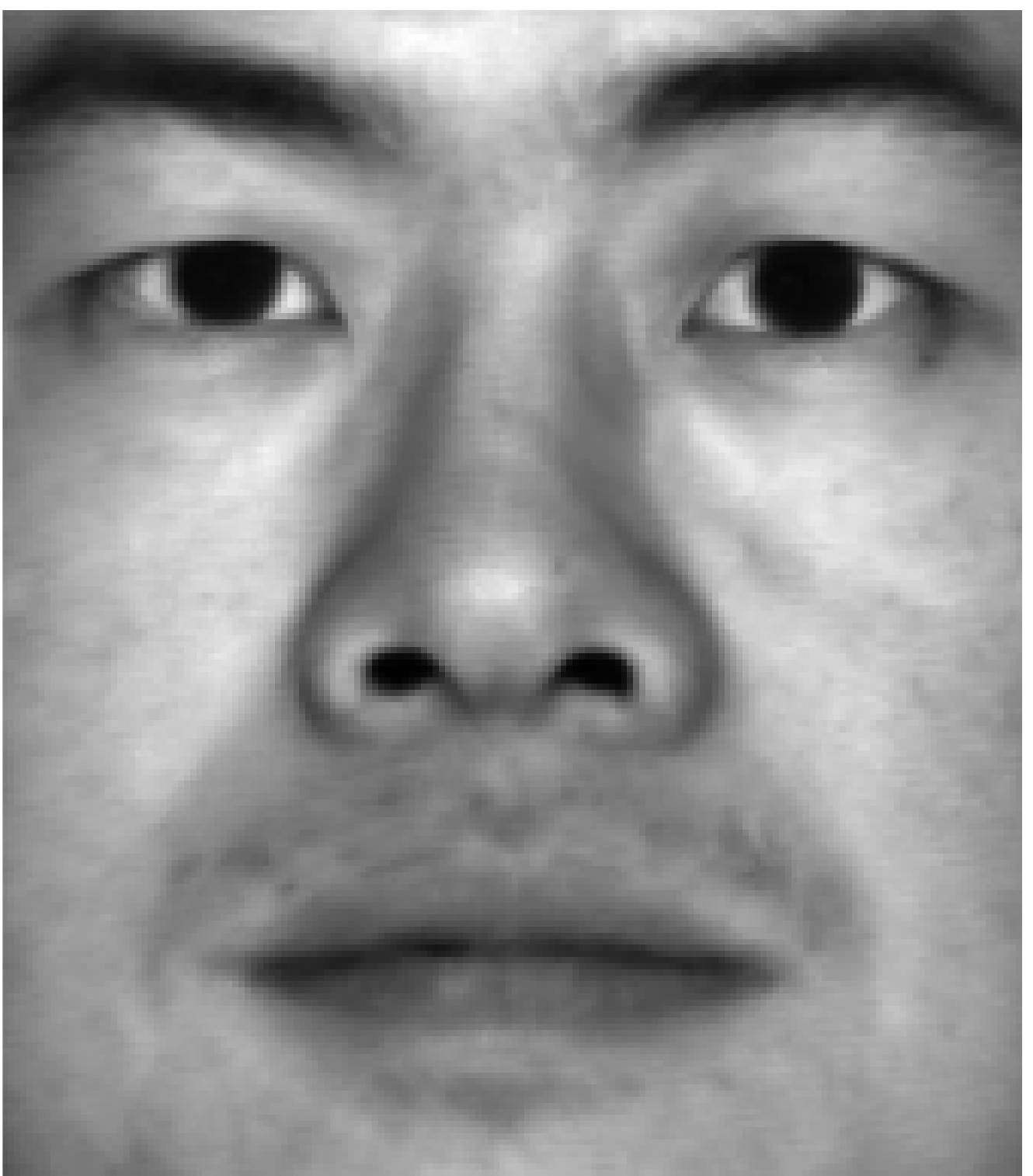}			\\

&
\includegraphics[width=0.1\textwidth]{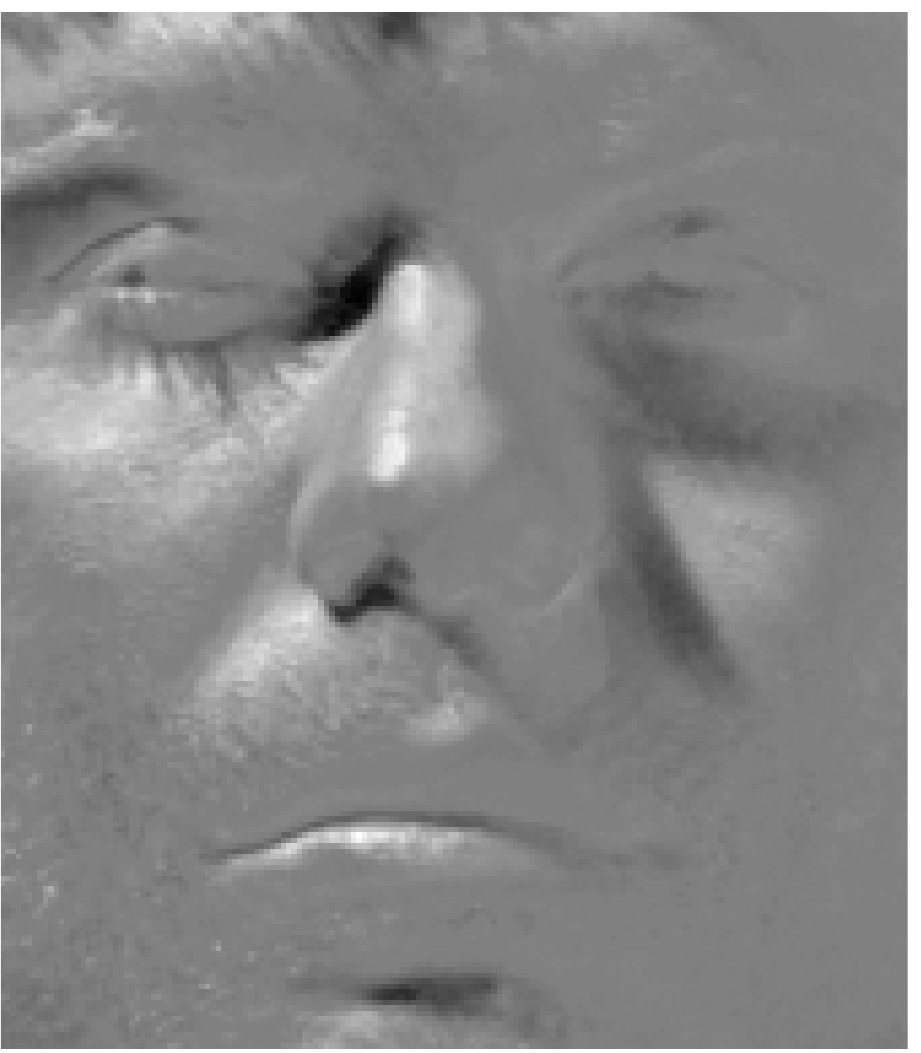}		&
\includegraphics[width=0.1\textwidth]{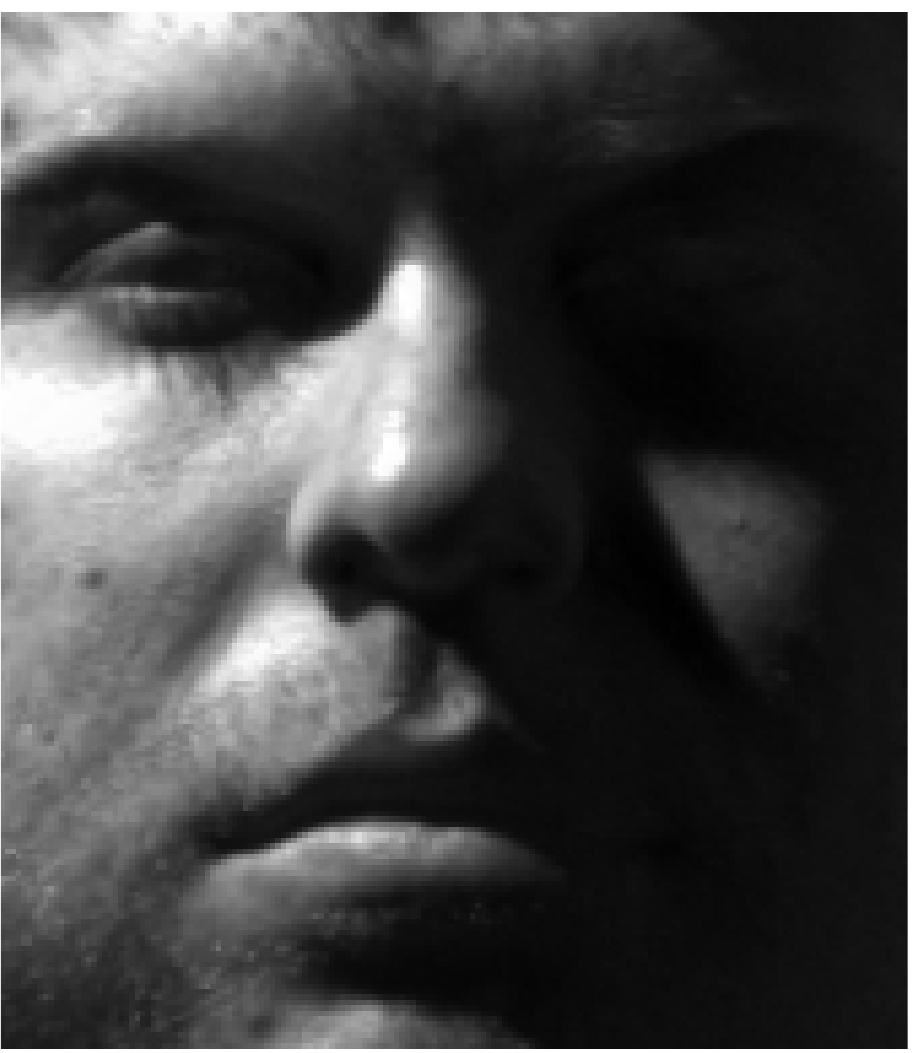}	&
\includegraphics[width=0.1\textwidth]{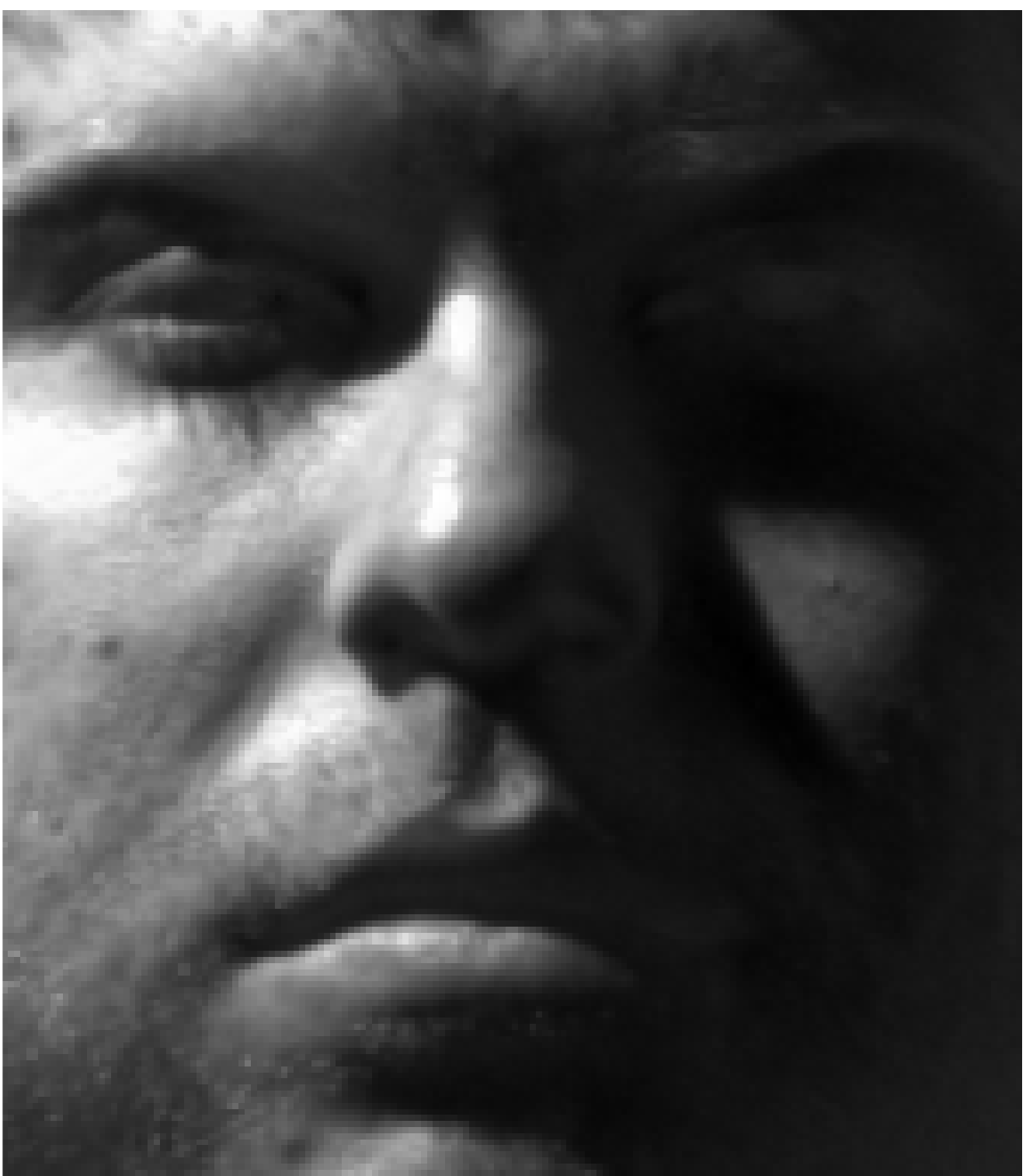}		&

&
&
&
\includegraphics[width=0.1\textwidth]{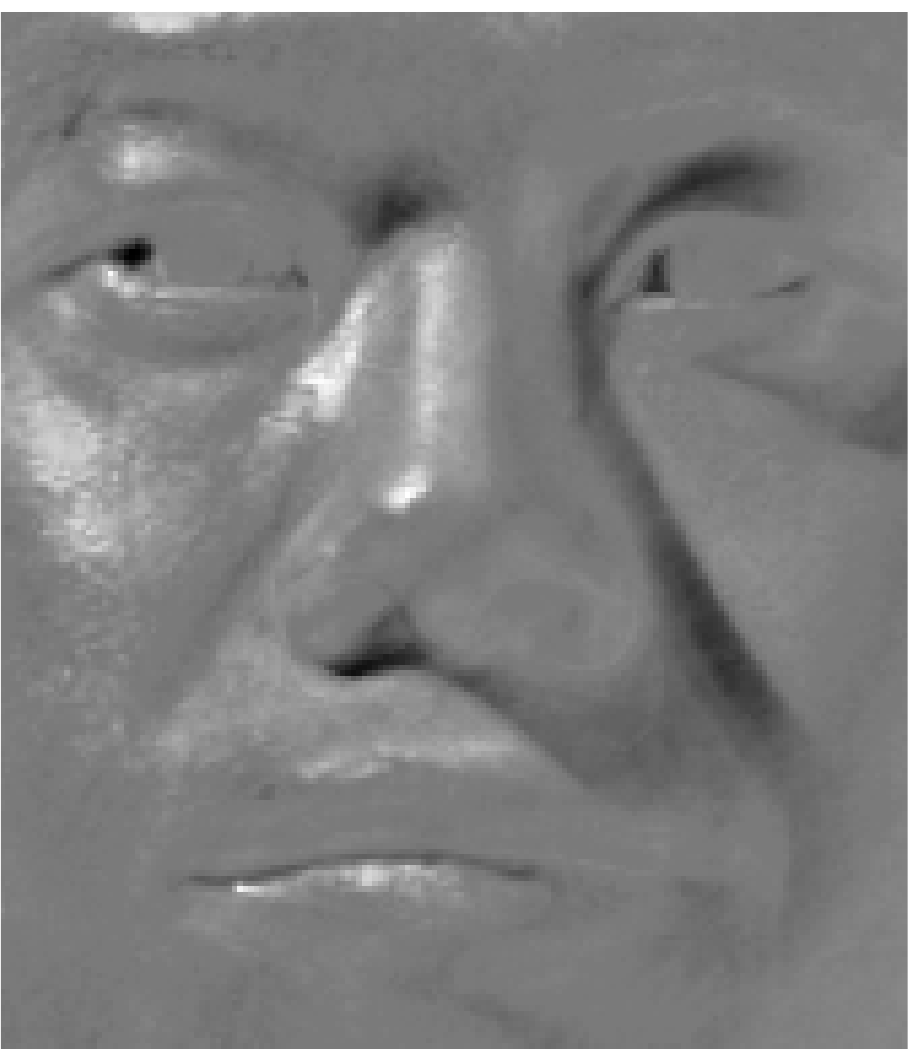}		&
\includegraphics[width=0.1\textwidth]{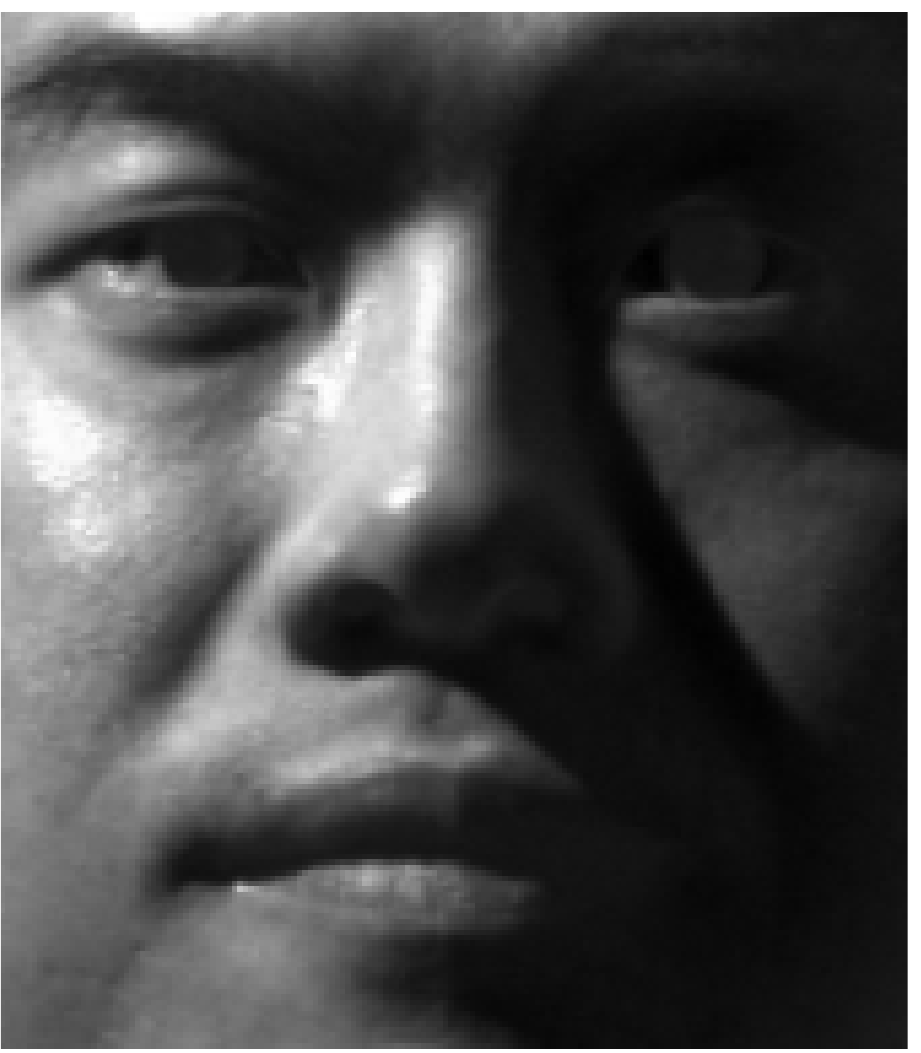}	&
\includegraphics[width=0.1\textwidth]{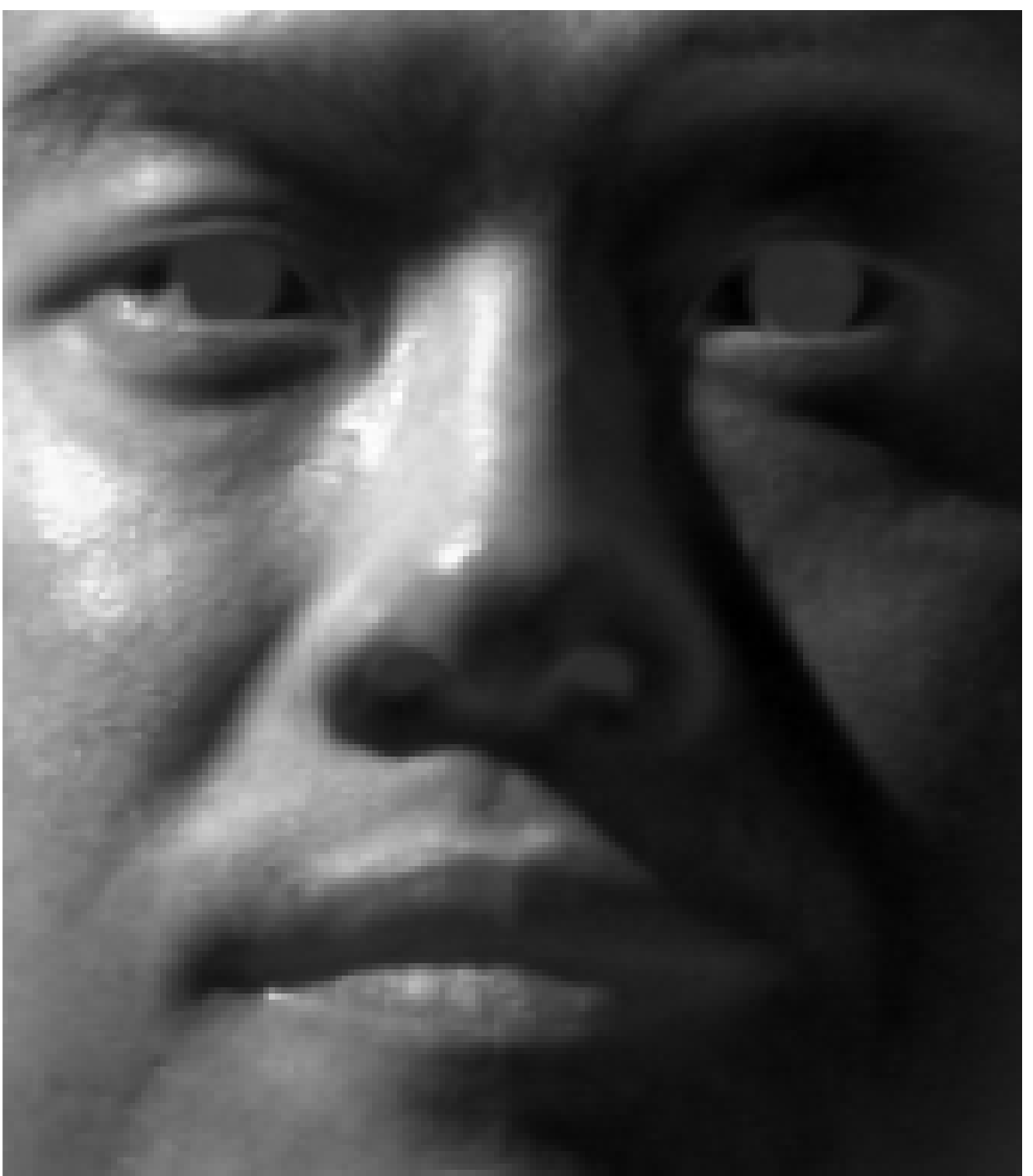}			\\	

 	& (1)  	& (2)  	& (3)  	& \multicolumn{3}{c}{}  	& (1)  	& (2) 	& (3)  
\end{tabular}
}
\caption{ \footnotesize Shadow removal results for subjects 1 and 2 from EYaleB data. For each of the two parts, the top left is the original image and the rest are recovered clean images (top) and shadows (bottom) by (1) IALM, (2) AltProj, and (3) F-FFP, respectively. } 
\label{fig_face}
\end{figure}

\begin{figure}[!tb]
\centering
\resizebox{0.7\columnwidth}{!}{
\begin{tabular}{c c c || c c c  }
\includegraphics[width=0.1\textwidth]{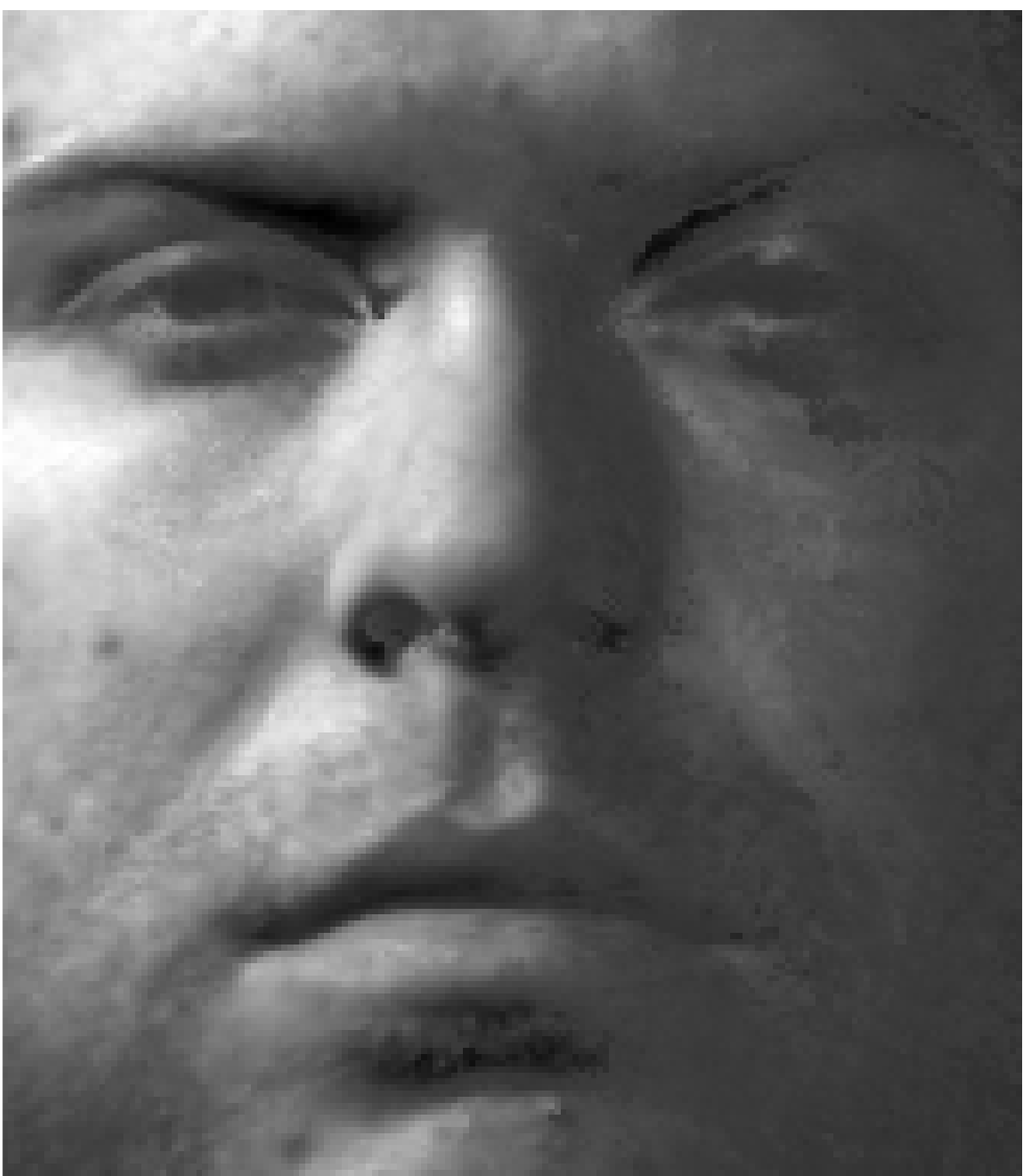}	&
\includegraphics[width=0.1\textwidth]{L_subject1_25_ffp.pdf}		&
&
&
\includegraphics[width=0.1\textwidth]{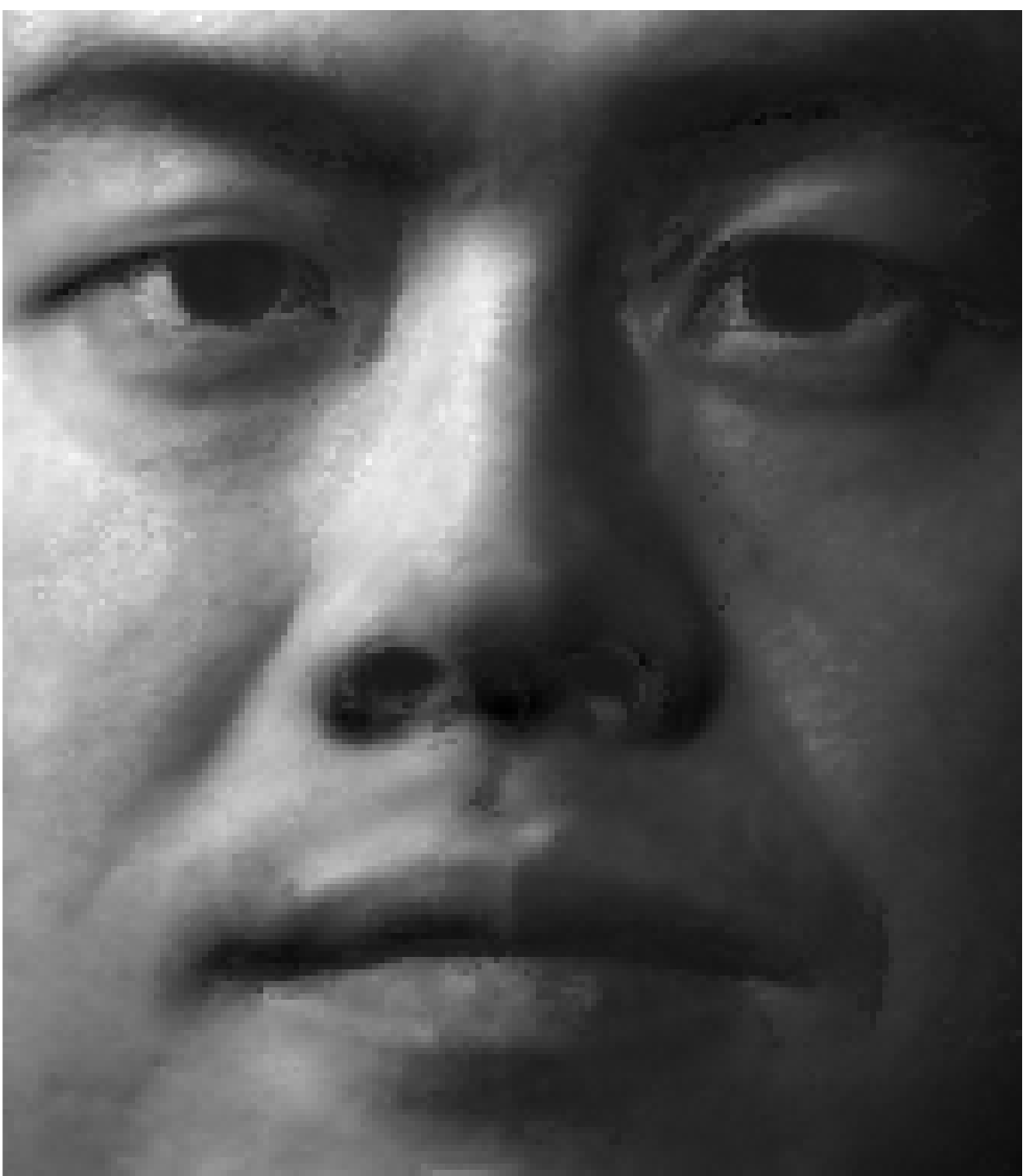}	&
\includegraphics[width=0.1\textwidth]{L_subject2_25_ffp.pdf}			\\

\includegraphics[width=0.1\textwidth]{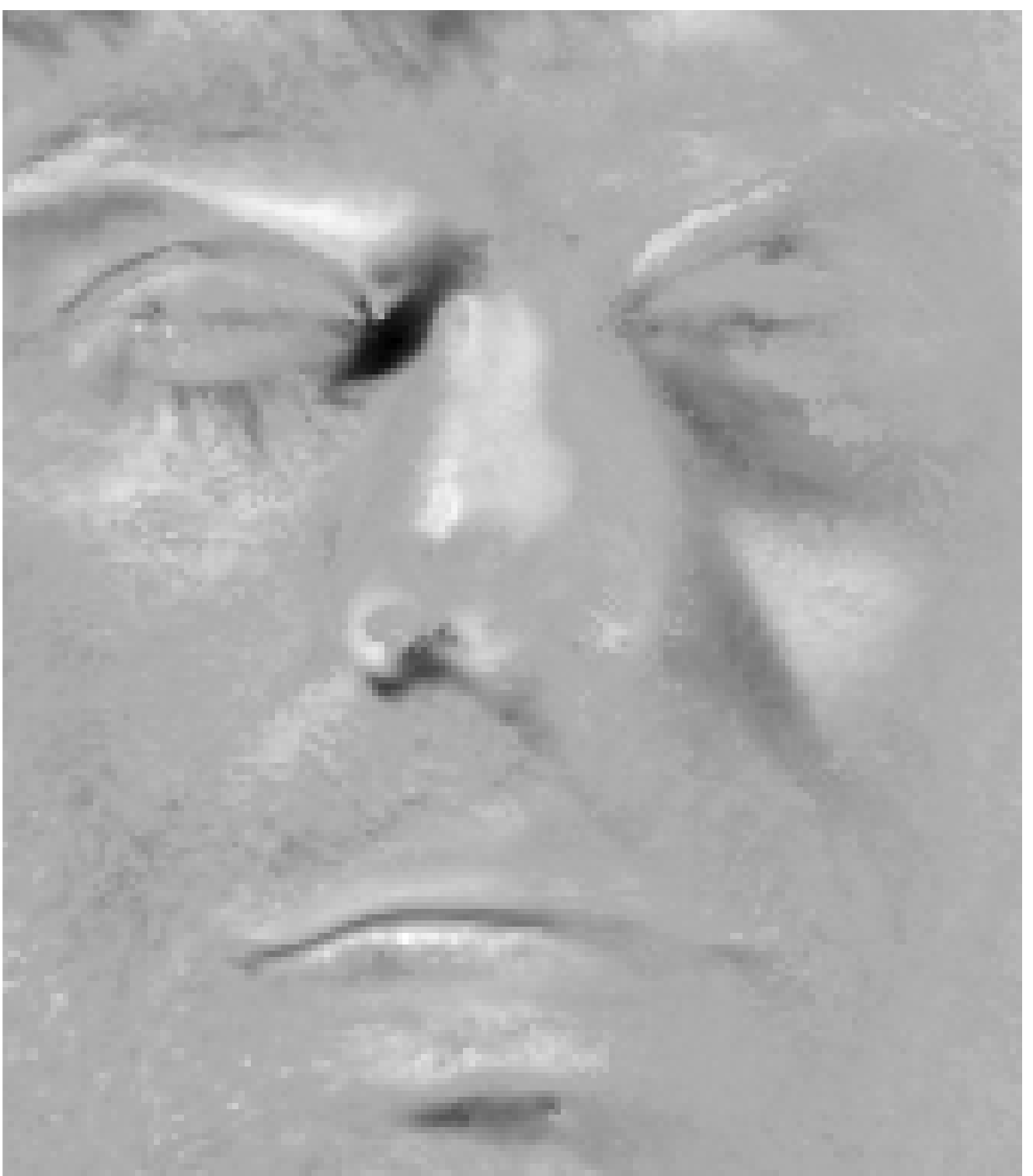}	&
\includegraphics[width=0.1\textwidth]{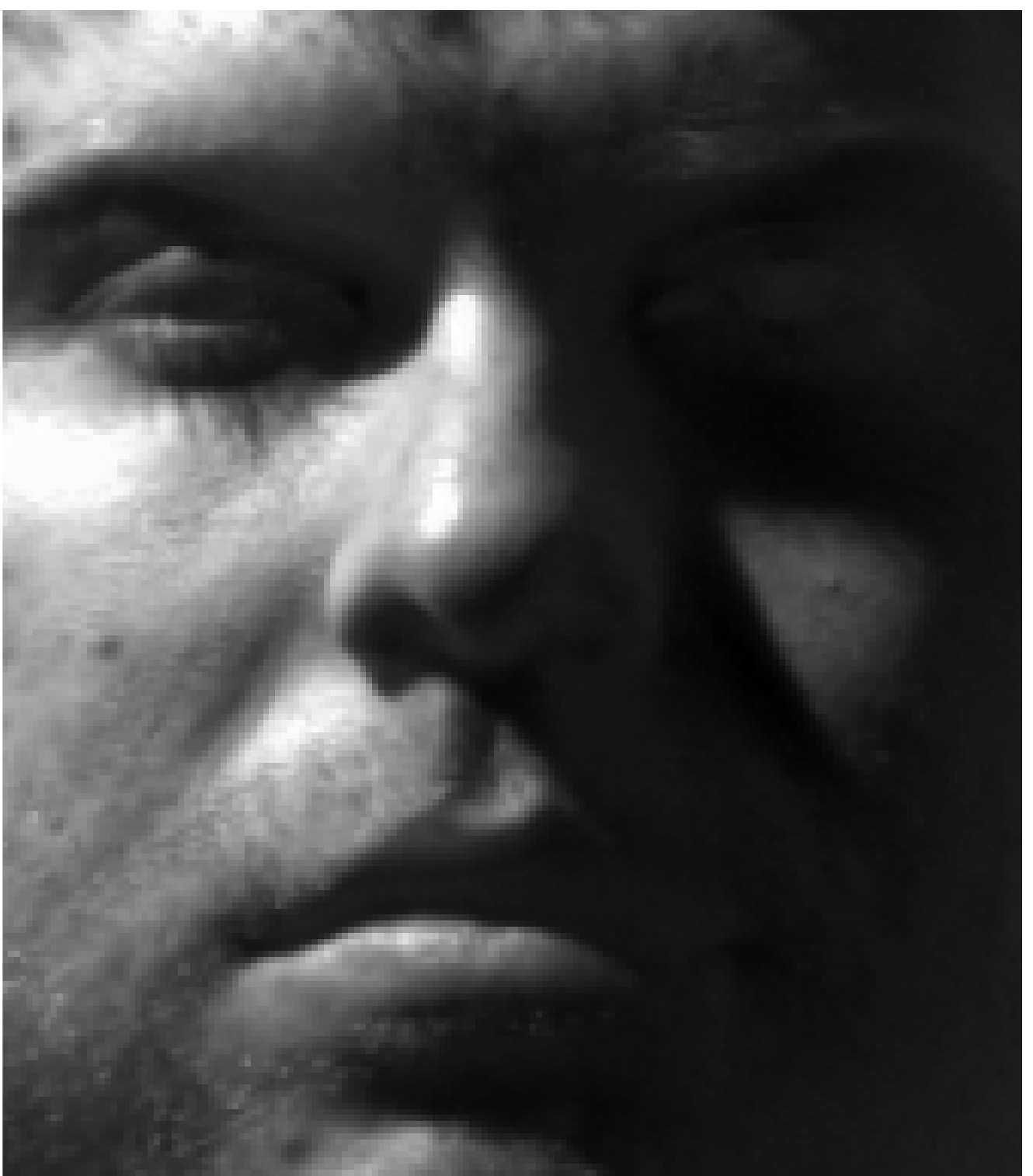}		&
&
&
\includegraphics[width=0.1\textwidth]{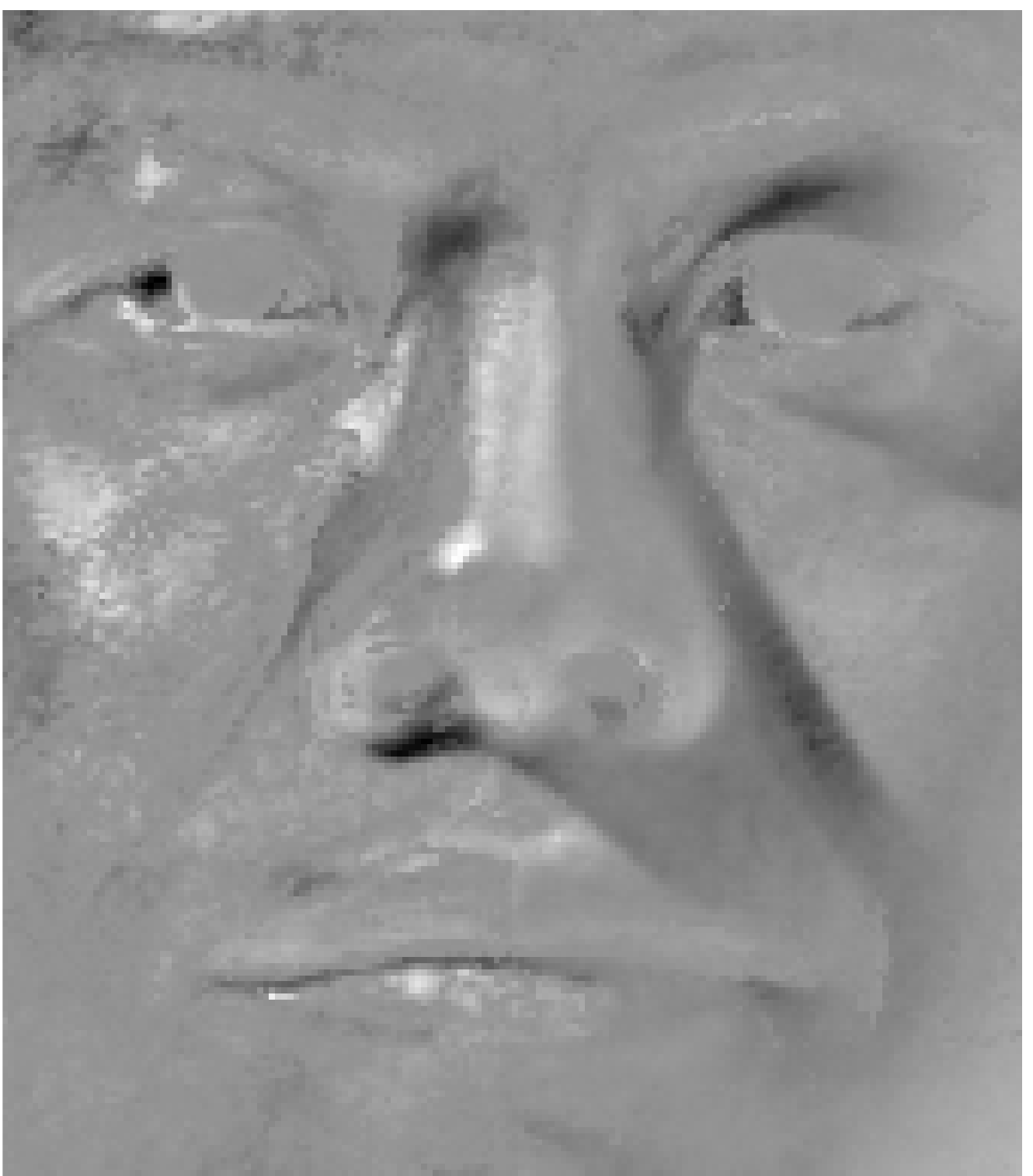}	&
\includegraphics[width=0.1\textwidth]{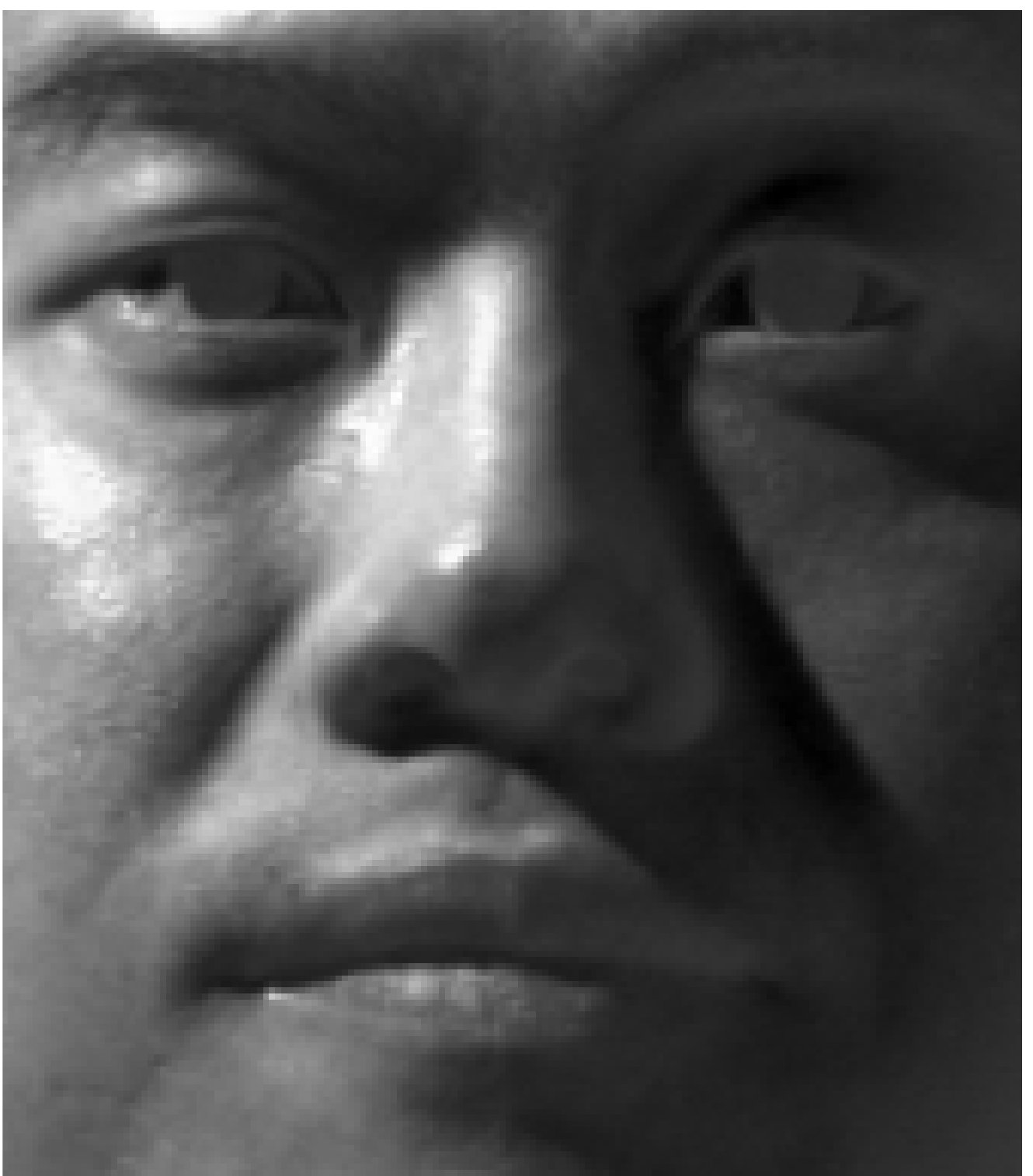}			\\	

 (1)  	& (2)  	  & \multicolumn{2}{c}{}  	& (1)  	& (2) 
\end{tabular}
}
\caption{\footnotesize Shadow removal results for subjects 1 and 2 from EYaleB data. The top panel are the recovered clean images and the bottom panel are the shadows by (1) AltProj and (2) U-FFP, respectively. $\lambda$ = 2e4 for U-FFP.} 
\label{fig_face_k5}
\end{figure}

First, we consider the case where $r$ is known. Following \cite{kang2015robustpca,candes2011robust}, IALM, AltProj, and F-FFP are applied to each subject and the quantitative and visual results are shown in Table \ref{tab_face} and Figure \ref{fig_face}, respectively. It is observed from Figure \ref{fig_face} that AltProj and F-FFP can successfully remove shadows from face images. The majority of shadows can be removed by IALM, but some still remain. From Table \ref{tab_face}, we can see that AltProj and F-FFP can recover the low-rank component with exactly rank 1, while IALM recovers $L$ which has a higher rank. Then, we consider the case where $r$ is unknown. Following the setting in \ref{sec_exp_background}, we set $k=5$ and apply AltProj and U-FFP to each subject. The quantitative and visual results are shown in Table \ref{tab_face_k5} and Figure \ref{fig_face_k5}, respectively. It is observed from Table \ref{tab_face_k5} that U-FFP has similar visual results as F-FFP while AltProj appears unable to remove shadows completely. Quantitatively as shown in Table \ref{tab_face_k5}, AltProj gives an $L$ that has a higher rank while U-FFP gives an $L$ that has the true rank. Besides, the proposed U-FFP is the fastest among all these methods.
\begin{table}[!tb]
\huge

\centering
\caption{Recovery Results of Face Data with $k=1$\label{tab_face}}
\resizebox{1.0\columnwidth}{!}{
\begin{tabular}{|c||c|| c |c |c |c | c| c| }
\hline		
Data 	
& Method	& Rank($Z$) & ${\|S\|_0}/{(d n)}$ & $\frac{\|X-Z-S\|_F}{\|X\|_F}$	& \# of Iter. 	& \# of SVDs	& Time	\\ \hline	

\multirow{4}{3cm}{ Subject 1 } 	
& AltProj	& 1			& 0.9553		& 8.18e-4		& 50			& 51			& 4.62		\\	\cline{2-8}
& IALM		& 32		& 0.7745		& 6.28e-4		& 25			& 26			& 2.43		\\	\cline{2-8}
& F-FFP		& 1			& 0.9655		& 8.86e-4		& 36			& 36			& 1.37		\\	\hline\hline

\multirow{4}{3cm}{ Subject 2 } 	
& AltProj	& 1			& 0.9755		& 2.34e-4		& 49			& 50			& 5.00		\\	\cline{2-8}
& IALM		& 31		& 0.7656 		& 6.47e-4		& 25			& 26			& 2.66		\\	\cline{2-8}
& F-FFP		& 1			& 0.9492		& 9.48e-4		& 36			& 36			& 1.37		\\	\hline
%---------------------
\end{tabular}
}

\end{table}
\begin{table}[!tb]
\huge
\centering
\caption{Recovery Results of Face Data with $k=5$}
\resizebox{1.0\columnwidth}{!}{
\begin{tabular}{|c||c|| c |c |c |c | c| c| }
\hline		
Data 	
& Method	& Rank($Z$) & ${\|S\|_0}/{(d n)}$ & $\frac{\|X-Z-S\|_F}{\|X\|_F}$	& \# of Iter. 	& \# of SVDs	& Time	\\ \hline	

\multirow{3}{3cm}{ Subject 1 } 	
& AltProj	& 5			& 0.9309		& 3.93e-4		& 51			& 55			& 6.08		\\	\cline{2-8}
& U-FFP		& 5			& 0.9647		& 8.40e-4		& 36			& 36+36+36		& 1.69		\\	\hline\hline

\multirow{3}{3cm}{ Subject 2 } 	
& AltProj	& 5			& 0.8903		& 6.40e-4		& 54			& 58			& 7.92		\\	\cline{2-8}
& U-FFP		& 1			& 0.9651		& 5.72e-4		& 37			& 37+37+37		& 1.74		\\	\hline
%---------------------
\end{tabular}
}
\label{tab_face_k5}
\end{table}
\subsection{Anomaly Detection}
\begin{figure}[!tb]
\centering
\includegraphics[width=1.0\columnwidth]{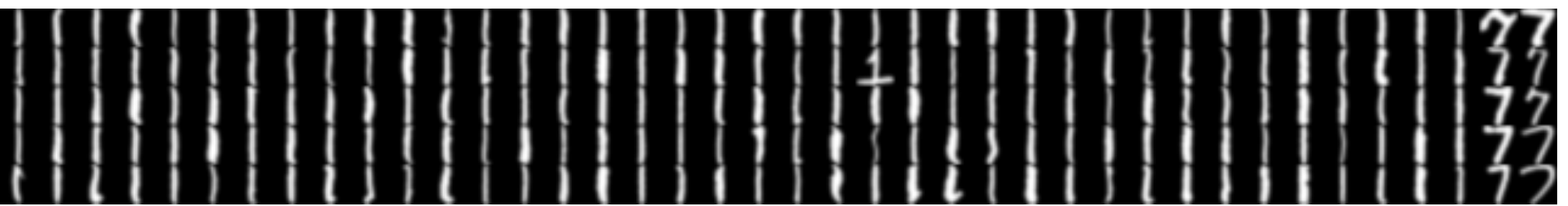} 
\caption{\footnotesize Selected `1's and `7's from USPS dataset. }
\label{fig_usps_example}
\end{figure}

\begin{figure}[!tb]
\centering
\includegraphics[width=0.5\columnwidth]{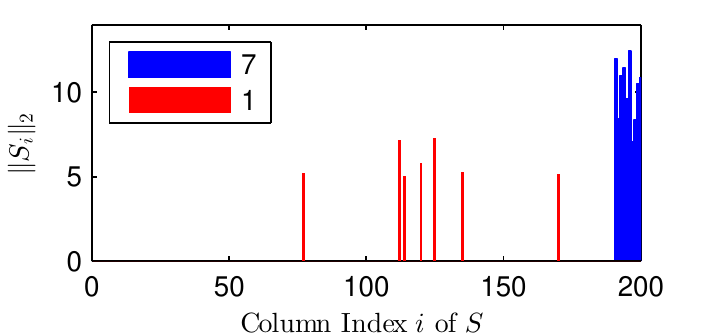}
\caption{\footnotesize $\ell_2$-norms of the rows of $S$. }
\label{fig_usps_bar}
\end{figure}

\begin{figure}[!tb]
\centering
\subfigure{\includegraphics[width=1.0\columnwidth]{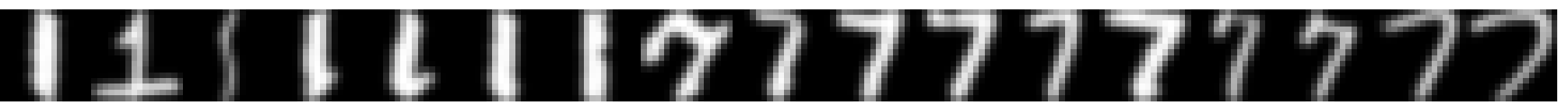} }
\caption{\footnotesize Outliers including some unusual `1's and all `7's identified by F-FFP. }
\label{fig_usps_anomaly}
\end{figure}

Given a number of images from one subject, they form a low-dimensional subspace. Any image significantly different from the majority of the images can be regarded as an outlier; besides, fewer images from another subject can be regarded as outliers. Anomaly detection is to identify such kinds of outliers. USPS dataset contains 9,298 images of hand-written digits of size $16\times 16$. Following \cite{kang2015robustpca}, among these images, we select the first 190 images of `1's and the last 10 of `7's and construct a data matrix of size $256\times 200$ by regarding each vectorized image as a column. Since the number of `1's is far greater than `7's, the former is the dominant digit, while the images of the latter are outliers. The true rank of $L$ should be 1. Some examples of these selected images are shown in Figure \ref{fig_usps_example}. It is observed that besides `7's, some `1's are quite different from the majority. Therefore, anomaly detection, in this case, is to detect not only the `7's, but also the anomaly of `1's. After applying F-FFP, the columns in $S$ that correspond to anomalies contain relatively larger values. We use $\ell_2$ norm to measure every column of $S$ and show the values in Figure \ref{fig_usps_bar}. These outliers can be identified by finding the columns with the highest bars. For ease of visualization, we vanish all those values that are smaller than 5 in Figure \ref{fig_usps_bar}. The corresponding digits are shown in Figure \ref{fig_usps_anomaly} as outliers, which include all `7's and several unusual `1's.
\subsection{Scalability }
\label{sec_scalability}
\begin{figure}[!tb]
\centering
\subfigure[ F-FFP (time v.s. $n$)]{\includegraphics[width=0.46\columnwidth]{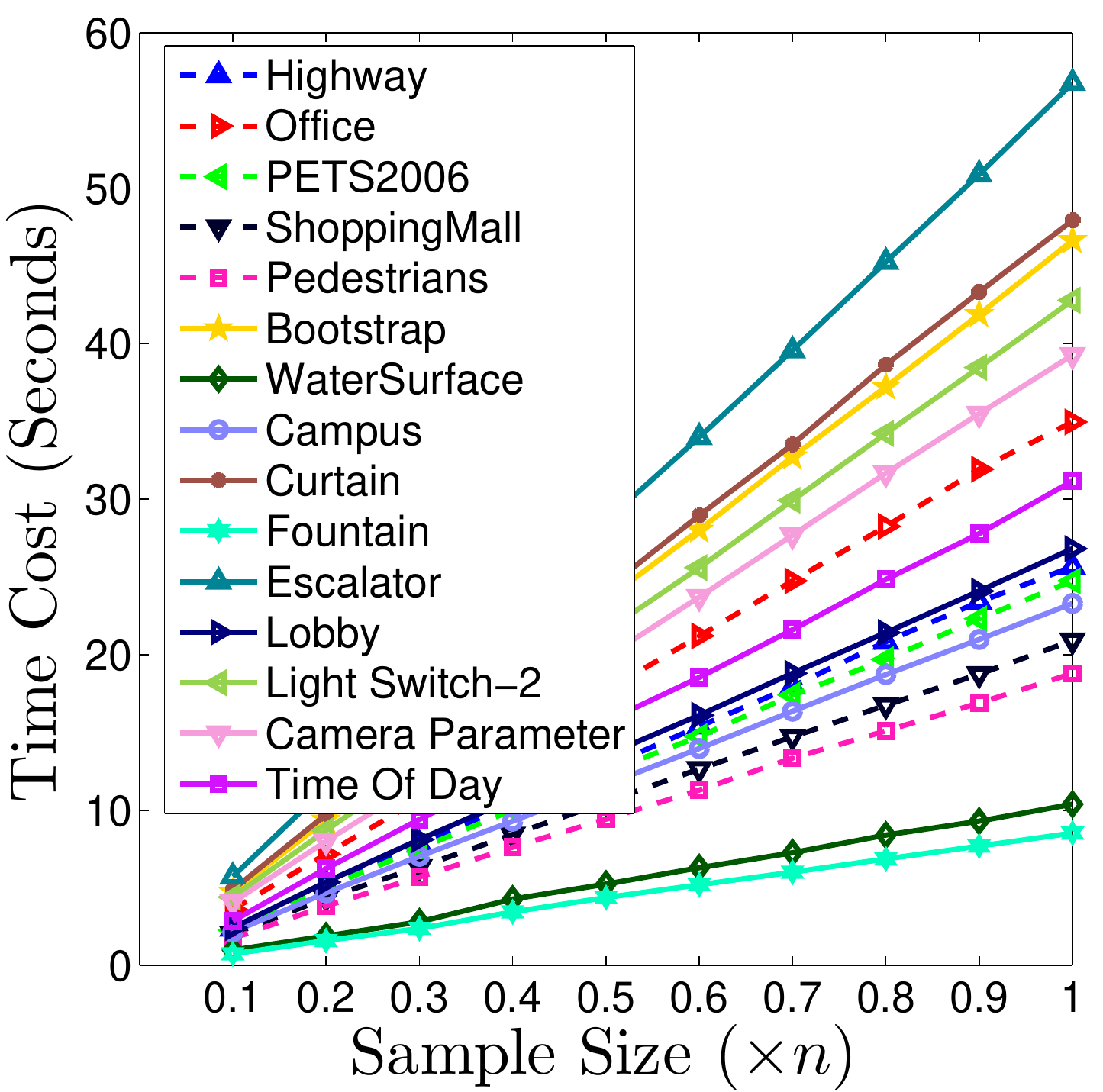} }
\subfigure[ F-FFP (time v.s. $d$)]{\includegraphics[width=0.46\columnwidth]{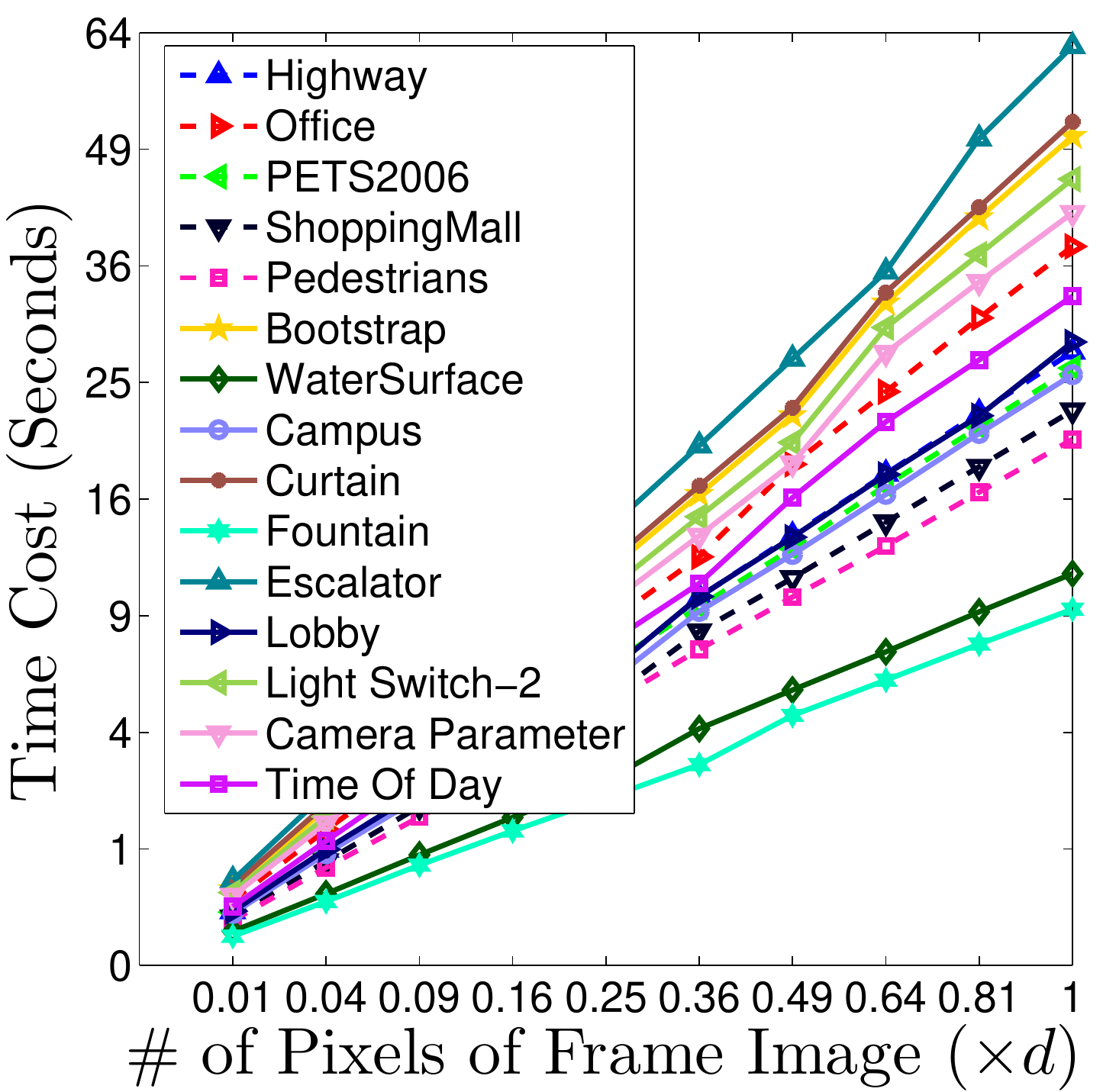} }

\subfigure[ U-FFP (time v.s. $n$)]{\includegraphics[width=0.46\columnwidth]{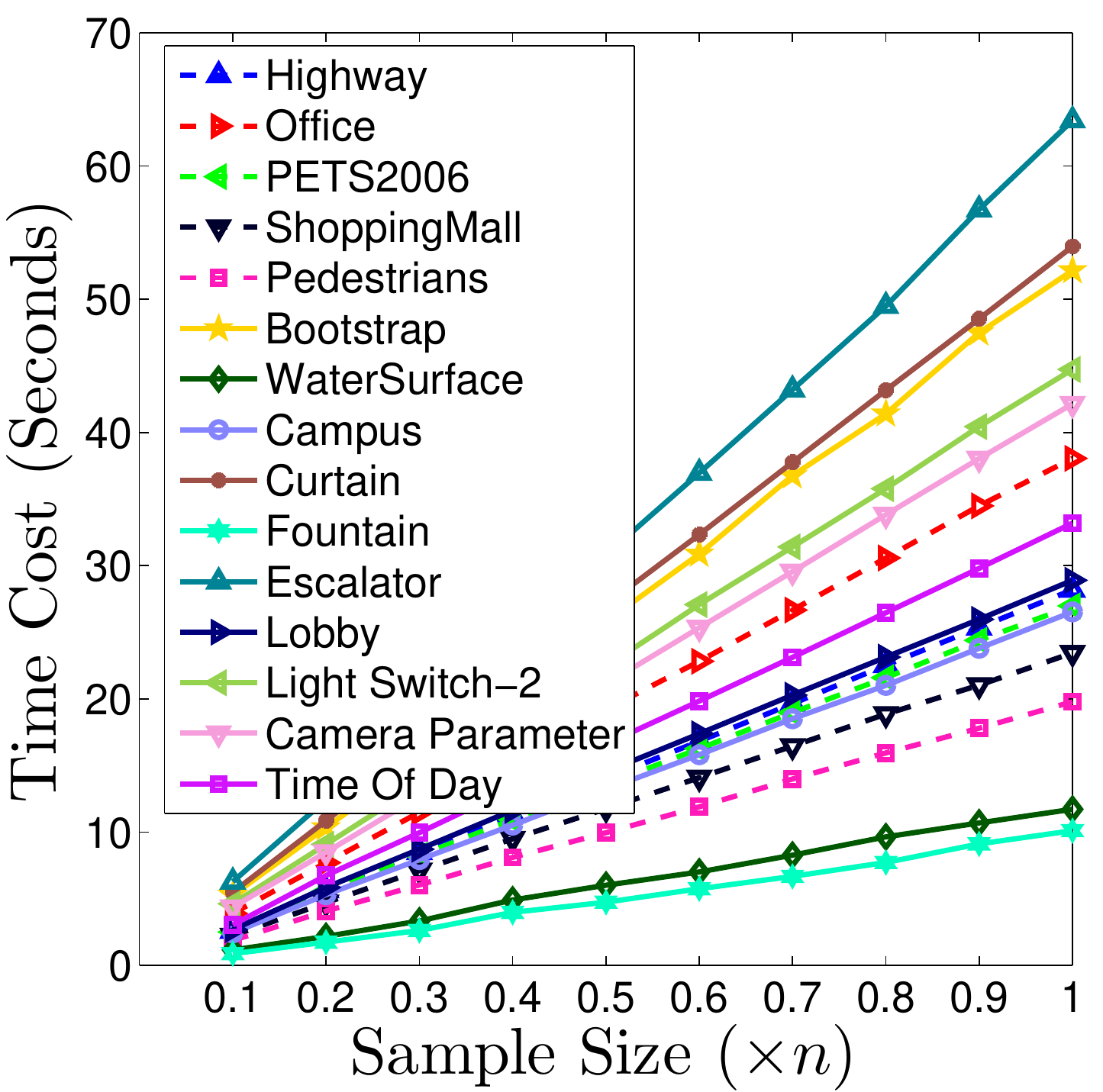} }
\subfigure[ U-FFP (time v.s. $d$)]{\includegraphics[width=0.46\columnwidth]{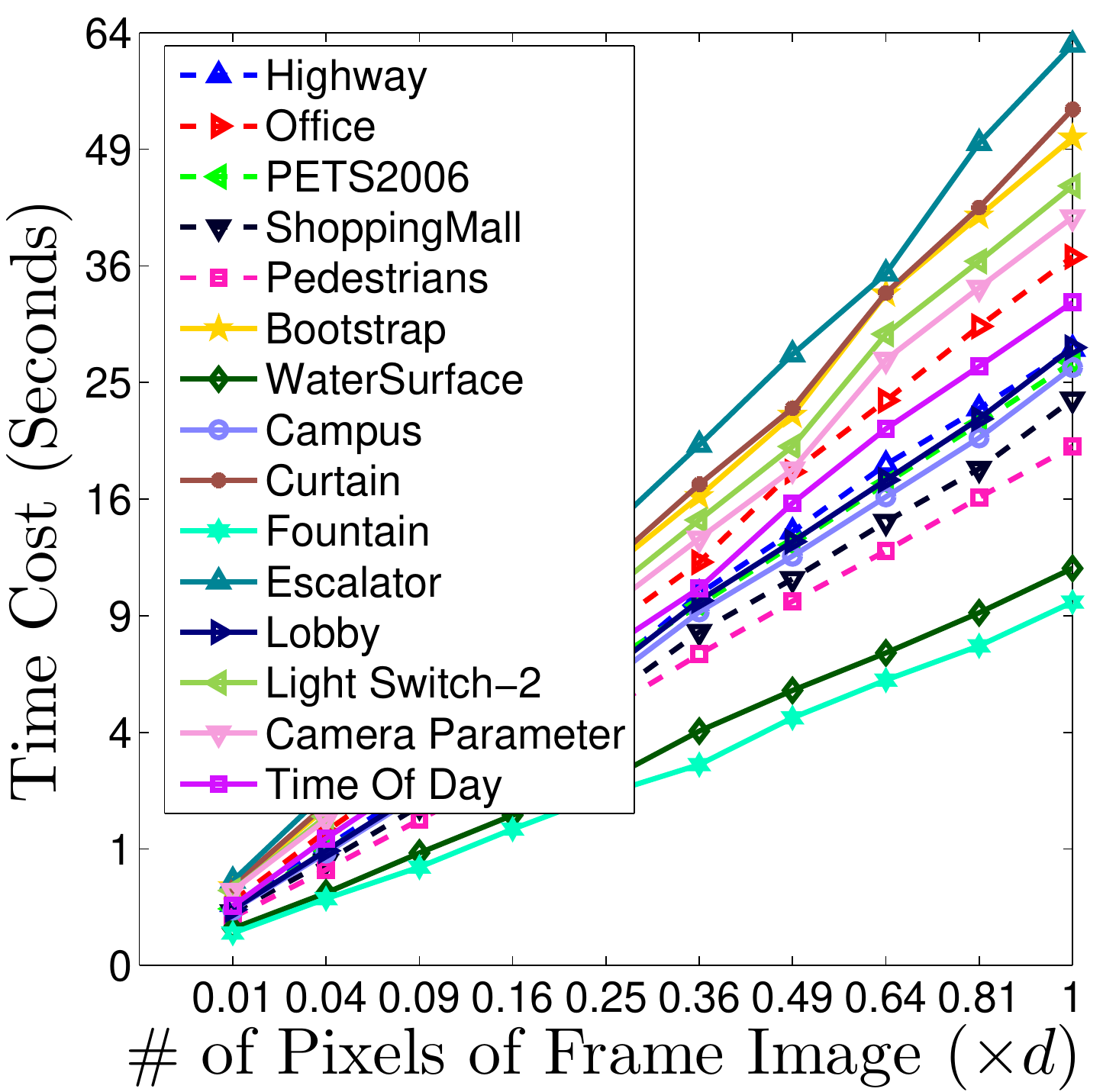} }
\caption{\footnotesize Plots of time cost of F-FFP and U-FFP as functions of data size and dimension. 
}
\label{fig_scalability}
\end{figure}

To numerically illustrate the scalability of F-FFP and U-FFP, we test the time cost versus the values of $n$ and $d$, respectively. To test how the computation time increases with $n$, we uniformly sample a partition of the frames with the sampling rate in $\{0.1,0.2,\cdots,1.0\}$ and keep all pixels for each frame. To test the relationship of time with respect to $d$, we use all frames and down-sample each frame with different rates varying over $\{0.1^2,0.2^2,\cdots,1.0^2\}$, to keep the spatial information of each frame. We run F-FFP and U-FFP with 50 iterations in 10 repeated runs and we report the average time in Figure \ref{fig_scalability}. It is observed that the time cost increases essentially linearly with both $n$ and $d$ for both F-FFP and U-FFP. 

\section{Conclusion}
\label{sec_conclusion}
In this paper, we propose a new, factorization-based RPCA model. Non-convex rank approximation is used to minimize the rank when the ground truth is unknown. ALM-type optimization is developed for solving the model. Our model and algorithms admit scalability in both data dimension and sample size, suggesting a potential for real world applications. Extensive experiments testify to the effectiveness and scalability of the proposed model and algorithms both quantitatively and qualitatively.

%% use section* for acknowledgement
\section*{Acknowledgment}
Qiang Cheng is the corresponding author. This work is supported by National Science Foundation under grant IIS-1218712, National Natural Science Foundation of China, under grant 11241005, and Shanxi Scholarship Council of China 2015-093. %, Fund Program for the Scientific Activities of Selected Returned Overseas Professionals in Shanxi Province.

%\newpage
%\footnotesize
\bibliographystyle{IEEEtran}
\footnotesize \bibliography{frpca_ref}

% that's all folks
\end{document}